\documentclass{article}

\usepackage{iclr2027_conference,times}

\usepackage[utf8]{inputenc}
\usepackage[T1]{fontenc}
\usepackage{hyperref}
\usepackage{url}
\usepackage{booktabs}
\usepackage{amsfonts}
\usepackage{nicefrac}
\usepackage{microtype}
\usepackage{xcolor}
\usepackage{graphicx}
\usepackage{wrapfig}
\usepackage{caption}
\usepackage{multirow}
\usepackage{subcaption}
\usepackage{amsmath}
\usepackage{algorithm}
\usepackage{algpseudocode}
\usepackage{colortbl}
\usepackage{siunitx}
\usepackage{float}
\usepackage{enumitem}
\usepackage{pifont}
\usepackage{soul}
\usepackage[most]{tcolorbox}

\definecolor{rowhl}{gray}{0.92}
\definecolor{softblue}{RGB}{230, 240, 250}
\definecolor{lavender}{RGB}{229, 221, 241}
\definecolor{takeawayblue}{RGB}{40, 68, 130}

\newtcolorbox{findingbox}[1][]{
    enhanced,
    colback=gray!5,
    colframe=black,
    fonttitle=\bfseries,
    boxed title style={colback=black, sharp corners},
    attach boxed title to top left={xshift=3mm, yshift=-2.5mm},
    title={#1}
}

\newtcolorbox{takeawaybox}[1][]{
    enhanced,
    colback=gray!5,
    colframe=takeawayblue,
    fonttitle=\bfseries,
    boxed title style={colback=takeawayblue, sharp corners},
    attach boxed title to top left={xshift=3mm, yshift=-2.5mm},
    title={#1}
}

\newtcolorbox{remarkbox}[1][]{
    enhanced,
    colback=yellow!5,
    colframe=black,
    fonttitle=\bfseries,
    boxed title style={colback=black, sharp corners},
    attach boxed title to top left={xshift=3mm, yshift=-2.5mm},
    title={#1}
}

\tcbuselibrary{skins, breakable}

\definecolor{prompttitle}{RGB}{228, 138, 140}
\definecolor{promptbody}{RGB}{253, 227, 225}
\definecolor{unstableinterval_one}{RGB}{198, 81, 159}
\definecolor{unstableinterval_two}{RGB}{113, 109, 178}

\newtcolorbox{promptbox}[1][]{
    enhanced,
    breakable,
    colback=promptbody,
    colbacktitle=prompttitle,
    colframe=prompttitle,
    boxrule=1pt,
    arc=4pt,
    outer arc=4pt,
    fontupper=\small,
    left=8pt,
    right=8pt,
    top=6pt,
    bottom=6pt,
    toptitle=4pt,
    bottomtitle=4pt,
    lefttitle=8pt,
    righttitle=8pt,
    coltitle=white,
    fonttitle=\small\bfseries,
    halign title=left,
    title={#1},
}

\title{To Think or Not to Think: Allocating Reasoning Where It Helps}

\author{
Zhengdong He\textsuperscript{1,2},
Yunfan Zhou\textsuperscript{1,2},
Jianguo Yao\textsuperscript{1,2},
Haibing Guan\textsuperscript{1,2},
Xijun Li\textsuperscript{1,2}\thanks{Corresponding author.}\\[2mm]
\textsuperscript{1}Shanghai Key Laboratory of Scalable Computing and Systems\\
\textsuperscript{2}School of Computer Science, Shanghai Jiao Tong University
}

\iclrfinalcopy

\begin{document}

\maketitle

\begin{abstract}
Reinforcement learning (RL) has proven effective in enhancing the reasoning performance of large language models (LLMs), particularly in complex mathematical and programming tasks. However, this capability comes with systematic \textit{length misallocation}, in which models devote excessive reasoning to simple questions while terminating prematurely on harder ones, degrading inference efficiency with negligible accuracy improvement. Many length-adaptive methods mitigate this issue by allocating token budgets according to question difficulty, under the implicit assumption that harder questions benefit monotonically from extended reasoning. In contrast, we find that the effect of reasoning length on accuracy is concentrated on \textit{partially solvable} questions. Our further analysis reveals that explicit length rewards can produce unintended training dynamics. Motivated by these findings, we propose \textbf{CARE}---\textbf{C}ontrastive \textbf{A}ccuracy \textbf{R}eward \textbf{E}stimation---which compares the beneficial length adjustment per question from online sampled responses and applies adaptive length rewards within Group Relative Policy Optimization, with no extra hyperparameters or additional inference cost. Experiments across multiple reasoning benchmarks demonstrate that our method improves Pass@1 by up to \(4\%\) while simultaneously reducing reasoning length by \(37\%\), achieving higher token efficiency. Code will be available upon the acceptance of this paper.

\end{abstract}

\section{Introduction}

Reinforcement learning (RL) has achieved remarkable success in enhancing the reasoning capabilities of large language models (LLMs), as exemplified by OpenAI-o1~\citep{openai2024o1}, DeepSeek-R1~\citep{guo2025deepseekr1}, and Kimi-1.5~\citep{team2025kimi}. Building on advances in chain-of-thought prompting~\citep{wei2022chain, feng2024chainthought, merrill2023expressive} and self-improvement via verifiable rewards~\citep{zelikman2022star, ouyang2022training, lightman2023verify}, RL-trained LLMs exhibit sophisticated reasoning behaviors, including self-reflection, verification, and exploration of alternative paths. However, a persistent side-effect has emerged alongside these improvements: \textit{length misallocation}. Specifically, such models systematically overthink simple questions by generating unnecessarily long reasoning traces~\citep{chen2024overthinking}, while underthinking harder ones through premature abandonment of promising solution paths~\citep{wang2025underthinking} and reluctance to revise their initial answers~\citep{kumar2024score}. These behaviors inflate inference costs without improving accuracy, motivating a growing line of work on reasoning length control.

To mitigate length misallocation, recent studies incorporate explicit length signals into RL training objectives. One line of work targets \textit{static length compression}:~\citet{arora2025efficient} adjust rewards as a function of reasoning length; ~\citet{hou2025thinkprune} imposes progressively tightened token limits; ~\citet{aggarwal2025l1} optimizes policies under user-specified length budgets; ~\citet{team2025kimi} constrains length via a penalty term. A complementary direction pursues \textit{adaptive length allocation}: ~\citet{shen2025dast} and ~\citet{dai2025stablerl} adapt length penalties using online difficulty signals; ~\citet{xiang2025alp} adjusts length penalties inversely with question difficulty, and other methods~\citep{fang2025thinkless, zhang2025adaptthink} select reasoning modes based on task complexity. Despite their differences, these approaches share an implicit assumption: \textit{the optimal reasoning length scales monotonically with question difficulty, meaning that allocating more tokens to harder questions improves accuracy, whereas compressing easier ones may sacrifice little accuracy.} Notably, ~\citet{wu2025whenmore} demonstrate that the relationship between reasoning length and accuracy can be non-monotonic, calling this assumption into question. Nevertheless, it remains unclear how explicit length rewards affect questions of varying difficulty and whether the resulting training dynamics remain stable over the course of training. We investigate both questions in Section~\ref{sec:reasoning_length_accuracy}.

Our experiments on the Qwen3 family models reveal that the effect of reasoning length on accuracy is \textit{not} universally present across difficulty levels but is concentrated on partially solvable questions—those whose accuracy falls in \((0.25, 0.75]\), where the model generates correct answers for a proportion of sampled responses. For these questions, accuracy peaks near the mean length of correct responses and degrades when reasoning length is substantially shorter or longer. This selective pattern challenges the implicit assumption in several prior methods. Moreover, we find undesirable behaviors: \textit{short-reward training can unintentionally induce longer outputs, while long-reward training increases length through cyclic repetition rather than deeper reasoning in later training stages.}

Motivated by these findings, we propose \textbf{C}ontrastive
\textbf{A}ccuracy \textbf{R}eward \textbf{E}stimation (\textbf{CARE}). For each question during training, we estimate whether longer or shorter reasoning is currently more beneficial and use this signal to determine the direction of the length term in the reward. Since this estimate is derived directly from the sampled response group, it incurs no additional inference cost and adapts online as the policy shifts. This design avoids counterproductive length pressure by encouraging extended reasoning where it improves accuracy and promoting conciseness where brevity suffices.

Our main contributions are as follows:
\begin{itemize}[leftmargin=1.75em]
    \item We demonstrate that reasoning length primarily affects accuracy on partially solvable questions, and that directly applying length rewards can lead to unanticipated training effects, including length inflation and cyclic repetition.

    \item We propose \textbf{CARE}, a lightweight design, which estimates the beneficial length adjustment per question from online sampled responses at no additional inference cost and integrates directly into the GRPO~\citep{shao2024deepseekmath} framework without modifying the underlying training pipeline.

    \item We evaluate on Qwen3-1.7B, 4B and 8B across multiple public reasoning benchmarks, and show that our method improves Pass@1 by up to 4 percentage points over GRPO on HMMT 2026, and up to 37\% fewer tokens than the base model on average for Qwen3-4B, achieving better token efficiency without introducing additional hyperparameters or inference cost.
\end{itemize}

\section{Related Work}
\label{sec:related_work}
\paragraph{Reinforcement Learning for LLMs Reasoning.}

RL has demonstrated considerable success in advancing LLM reasoning capability, with notable examples including OpenAI-o1~\citep{openai2024o1} and DeepSeek-R1~\citep{guo2025deepseekr1}, which show that RL can elicit self-reflection and verification without relying on human-annotated trajectories.~\citet{team2025kimi} further scales reinforcement learning with verifiable rewards (RLVR) to 128K-token contexts and employs a length penalty to curb verbosity during training.~\citet{yu2025dapo} addresses exploration and training-stability challenges that arise in long-chain RL through Clip-Higher, Dynamic Sampling, and Token-Level policy gradient loss. However,~\cite{yue2025does} show that current RLVR primarily improves sampling efficiency over reasoning paths already present in the base model, rather than eliciting new reasoning patterns or capabilities. 

\paragraph{Overthinking and Underthinking in RL-Trained Models.}

Although RL-based training has markedly improved LLMs' mathematical reasoning, it also introduces length misallocation.~\citet{chen2024overthinking} show that models frequently overthink simple questions, consuming extra tokens with negligible accuracy improvement.~\citet{wang2025underthinking} identify the opposite issue, underthinking, where models abandon promising paths prematurely.~\citet{su2025between} demonstrate that both behaviors coexist, with overthinking on easy questions and underthinking on hard ones.~\citet{yeo2025demystifying} further find that, without reward shaping, reasoning length can grow uncontrollably during training and eventually degrade performance.~\citet{shojaee2025illusion} report that standard models can surpass reasoning models on easy questions, while both fail on highly complex ones.~\citet{an2025dontthinker} improve efficiency by suppressing harmful reasoning patterns while reinforcing useful ones, and~\citet{sui2025stopoverthinking} argue that existing approaches still do not fully resolve the overthinking and underthinking trade-off.

\paragraph{Efficient and Adaptive Reasoning via RL.}

To address length misallocation, RL-based approaches have developed along two main directions. The first focuses on efficient reasoning:~\cite{arora2025efficient} train models to reduce reasoning length under a single scalar control; ~\cite{hou2025thinkprune} imposes progressively stricter token budgets under GRPO~\citep{shao2024deepseekmath} with zero reward for over-budget outputs, and~\cite{aggarwal2025l1} introduces Length-Controlled Policy Optimization to support smooth accuracy--compute trade-offs under user-specified budgets. The second focuses on question-adaptive reasoning:~\cite{shen2025dast} estimates question difficulty via a token-length budget metric and adjusts penalties accordingly;~\cite{xiang2025alp} reallocates compute by scaling length penalties with question solve rates, and~\cite{fang2025thinkless} adaptively selects short-form or long-form reasoning based on question complexity. These methods share the assumption that harder questions benefit monotonically from longer reasoning. In contrast, our method makes no prior assumption about the preferred length adjustment based on question difficulty; instead, it estimates the beneficial length adjustment per question from online rollouts and applies length control only when the sampled responses indicate a clear accuracy gap.

\section{Preliminary: Group Relative Policy Optimization}
\label{sec:grpo}
GRPO~\citep{shao2024deepseekmath} is a PPO-based~\citep{schulman2017proximal} RL algorithm 
tailored for LLM reasoning that replaces the critic network in standard PPO with group-relative
reward normalization. For each question \(q\), GRPO samples a group of \(G\) responses \(\{o_1, o_2, \ldots, o_G\}\) from the old policy \(\pi_{\theta_{\mathrm{old}}}\), and computes a group-normalized advantage for each response:
\begin{equation}
\label{eq:grpo_advantage}
    \hat{A}_i = \frac{r(q, o_i) - \mu_{r}}{\sigma_{r}},
    \quad \mu_{r} = \frac{1}{G}\sum_{j=1}^{G} r(q, o_j),
    \quad \sigma_{r} = \sqrt{\frac{1}{G}\sum_{j=1}^{G}
    \left(r(q, o_j) - \mu_{r}\right)^2}
\end{equation}
where \(\mu_{r}\) and \(\sigma_{r}\) are the group-level mean and standard
deviation of rewards, serving as a self-contained baseline in place of the value network
required by PPO. The policy is then updated by maximizing the clipped surrogate objective:
\begin{equation}
\label{eq:grpo_objective}
\begin{aligned}
    &\mathcal{L}_{\mathrm{GRPO}}(\theta) = \mathbb{E}_{q \sim P(\mathcal{Q}),\, \{o_i\}_{i=1}^{G} \sim \pi_{\theta_{\mathrm{old}}}(\cdot \mid q)} \left[
    \frac{1}{G} \sum_{i=1}^{G} \frac{1}{|o_i|} \sum_{t=1}^{|o_i|}
    \left\{ \min \left[
        \frac{\pi_\theta(o_{i,t} \mid q, o_{i,<t})}{\pi_{\theta_{\mathrm{old}}}(o_{i,t} \mid q, o_{i,<t})} \hat{A}_{i},
    \right.\right.\right.\\
    &\left.\left.\left.
        \qquad\qquad\qquad\qquad
        \mathrm{clip}\!\left(
            \frac{\pi_\theta(o_{i,t} \mid q, o_{i,<t})}{\pi_{\theta_{\mathrm{old}}}(o_{i,t} \mid q, o_{i,<t})},\, 1-\varepsilon,\, 1+\varepsilon
        \right) \hat{A}_{i}
    \right] \right\}
    - \beta\, \mathbb{D}_{\mathrm{KL}}\!\left[\pi_\theta \,\|\, \pi_{\mathrm{ref}}\right]
    \right]
\end{aligned}
\end{equation}
where \(\varepsilon\) is the clipping threshold, \(\beta\) controls the strength of 
the KL penalty, \(\mathbb{D}_{\mathrm{KL}}[\pi_\theta \| \pi_{\mathrm{ref}}]\) denotes 
the KL divergence between the current policy and the frozen reference policy 
\(\pi_{\mathrm{ref}}\), and \(P(\mathcal{Q})\) denotes the distribution over the question 
set \(\mathcal{Q}\).

\section{The Relationship between Reasoning Length and Accuracy}
\label{sec:reasoning_length_accuracy}

Before presenting our method, we conduct \emph{motivating experiments} to investigate how reasoning length interacts with accuracy. We first show that reasoning length selectively influences accuracy depending on question difficulty (Section~\ref{sec:length_sensitivity}). We then demonstrate that training with length rewards produces effects misaligned with the conventional difficulty--token allocation assumption (Section~\ref{sec:accuracy_misalignment}). Finally, we show that length rewards may induce unexpected behaviors in later training stages (Section~\ref{sec:unexpected_behaviors}).

\textbf{Setup.}\quad All experiments in this section are conducted on the Qwen3 family model using mathematical reasoning benchmarks, including GSM8K~\citep{cobbe2021training}, MATH500~\citep{lightman2023verify}\footnote{MATH500 is a 500-question subset selected from the full MATH~\citep{hendrycks2021math} dataset.}, and MATH~\citep{hendrycks2021math}. For difficulty-dependent analyses, we estimate the difficulty of each question as its average accuracy over sampled responses and assign it to a difficulty subset accordingly, partitioning into \emph{hard} \([0, 0.25]\), \emph{partially solvable} \((0.25, 0.75]\), and \emph{easy} \((0.75, 1.0]\). For the training-based studies, we isolate the effect of length control from any particular reward design by adopting two controlled proxies: a \emph{short reward} \(r = y - L/L_{\max}\) that penalizes reasoning length to encourage concise outputs, and a \emph{long reward} \(r = y + L/L_{\max}\) that incentivizes longer outputs, where \(y \in \{0,1\}\) indicates response correctness and \(L\) denotes the reasoning length. Subsection-specific experimental protocols and full implementation details are provided in 
Appendix~\ref{app:experimental_details}; additional results are in Appendices~\ref{app:accuracy_misalignment_results} and~\ref{app:unexpected_behaviors_results}.

\subsection{Length Sensitivity Across Difficulty Subsets}
\label{sec:length_sensitivity}
We first examine how reasoning length influences accuracy on the base models without any training intervention. For each question, reasoning lengths are z-score normalized using the mean and standard deviation of all the correct samples within that question, and the average accuracy is then computed for each length bin across all questions in each difficulty subset.

\begin{figure}[t]
  \centering
  \includegraphics[width=\textwidth]{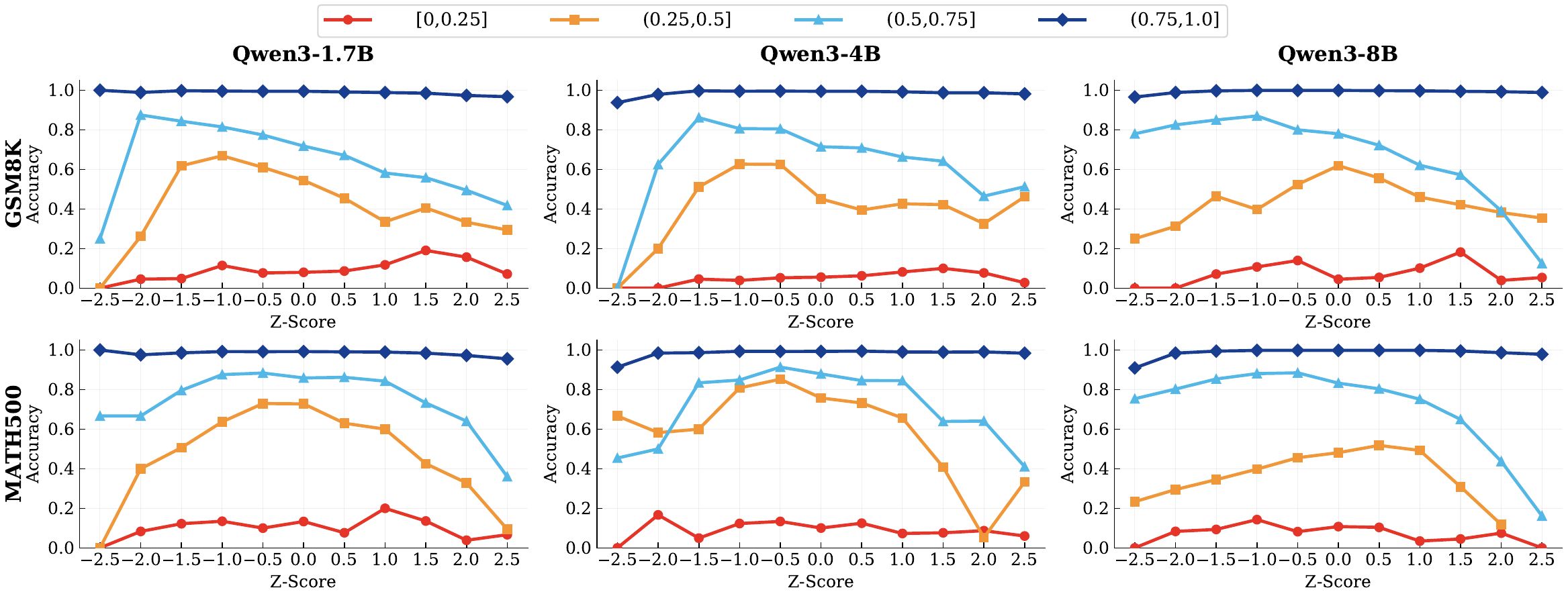}
  \caption{Relationship between reasoning length and accuracy across difficulty groups on GSM8K (top) and MATH500 (bottom) for Qwen3-1.7B, 4B, and 8B. Reasoning lengths are z-score normalized per question using correct-response statistics. Partially solvable questions (Blue/Orange curves) show a clear peak near zero z-score, whereas easy and hard questions show little sensitivity to length.}
  \label{fig:response_length_acc}
  \vspace{-1.25em}
\end{figure}

As shown in Figure~\ref{fig:response_length_acc}, the effect of reasoning length on accuracy is concentrated on partially solvable questions. For easy questions, accuracy remains high across nearly all length z-score intervals, with only a slight drop at large positive z-scores, indicating that performance is largely insensitive to length in this subset. In contrast, for partially solvable questions, accuracy peaks when reasoning lengths are near or slightly below the mean length of correct samples, and declines when responses are substantially shorter or longer. For the hardest questions, accuracy remains uniformly low and shows little sensitivity to reasoning length, implying that allocating additional tokens yields little benefit when the model is already unable to solve them. This pattern suggests that length control strategies should primarily target partially solvable questions, as they represent the subset where reasoning length has the greatest impact on outcomes.
\begin{findingbox}[Finding 1]
Partially solvable questions are \textit{length-sensitive}: Accuracy peaks around a moderate normalized reasoning length, and both insufficient and excessive length degrade accuracy.
\end{findingbox}

\subsection{Accuracy Misalignment of Conventional Token Allocation}
\label{sec:accuracy_misalignment}
Prior work~\citep{shen2025dast, xiang2025alp, fang2025thinkless} implicitly assumes that harder questions generally require longer reasoning length to be solved. While Section~\ref{sec:length_sensitivity} reveals that reasoning length selectively influences accuracy most on partially solvable questions, it remains unclear whether training with length rewards produces analogous effects. To probe this, we train short-reward and long-reward Qwen3-1.7B variants and evaluate the base, short-reward, and long-reward models on the MATH benchmarks. Questions are grouped into four difficulty subsets by base model accuracy, and each subset is visualized as a density heatmap over the \(\Delta\text{Length}\)--\(\Delta\text{Accuracy}\) grid, where \(\Delta\text{Length}\) and \(\Delta\text{Accuracy}\) denote the per-question change in reasoning length and accuracy relative to the base model, respectively.

\begin{figure}[t]
  \vspace{3.5em}
  \centering
  \includegraphics[width=\textwidth]{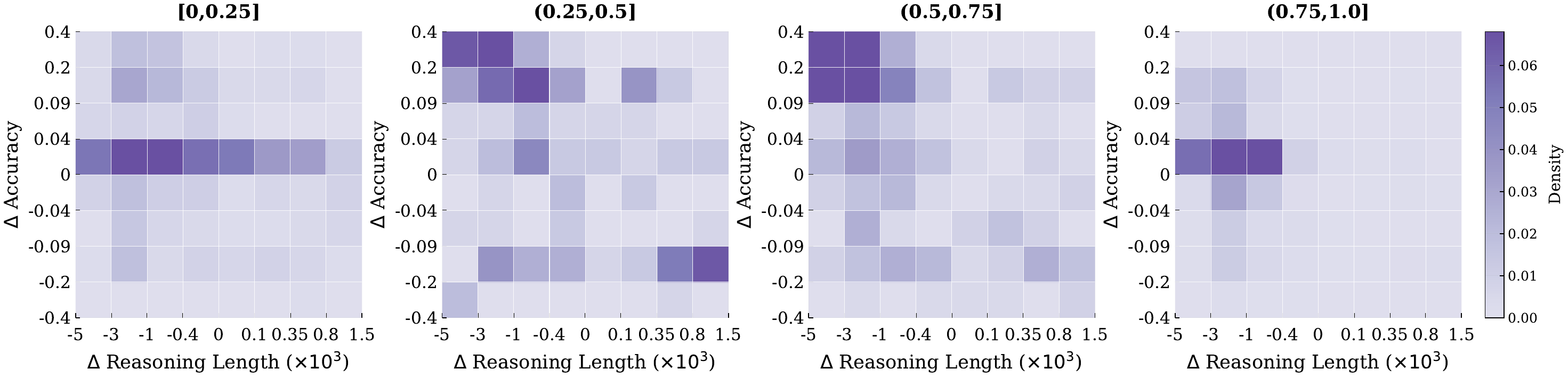}
  \caption{Density heatmaps of \(\Delta\)length vs.\ \(\Delta\)accuracy for short-reward Qwen3-1.7B across four difficulty subsets. A notable fraction of
questions in the \((0.25,0.5]\) and \((0.5,0.75]\) subsets show improved accuracy, contradicting the assumption that harder questions benefit from additional tokens.}
  \label{fig:1.7b_short_reward_delta_heatmap}
  \vspace{-1.5em}
\end{figure}

Figure~\ref{fig:1.7b_short_reward_delta_heatmap} shows that short-reward training systematically reduces reasoning length across all four difficulty subsets, yet its effect on accuracy is inconsistent. For the hardest subset, most questions tend to cluster around near-zero or slightly positive \(\Delta\)Accuracy, suggesting that reducing reasoning length has limited impact on questions that the model already fails to solve. For the easiest subset, the \(\Delta\)Accuracy remains concentrated near zero with only a slight drop. However, in the (0.25, 0.5] subset questions that the model partially solves, accuracy improves under length reductions. For these borderline-hard questions, the model has sufficient capability to reach the correct answer, whereas longer responses often introduce unnecessary reasoning steps that override the correct trajectory and thereby lower observed accuracy. A similar pattern holds across different model–reward combinations.
\begin{findingbox}[Finding 2]
The conventional difficulty-token allocation strategy, guided purely by
empirical accuracy, may not fully align with where adjusting reasoning
length in practice improves accuracy. 
\end{findingbox}

\subsection{Unexpected Training Effects under Length Rewards}
\label{sec:unexpected_behaviors}

Section~\ref{sec:accuracy_misalignment} reveals that the conventional difficulty--token allocation strategy does not adequately match the actual accuracy effects of reasoning length across questions. Beyond this misalignment, we also observe that both long-reward and short-reward training strategies exhibit distinct and unexpected behaviors across different models during the training process.

\begin{wrapfigure}{r}{0.53\linewidth}
  \centering
  \includegraphics[width=\linewidth]{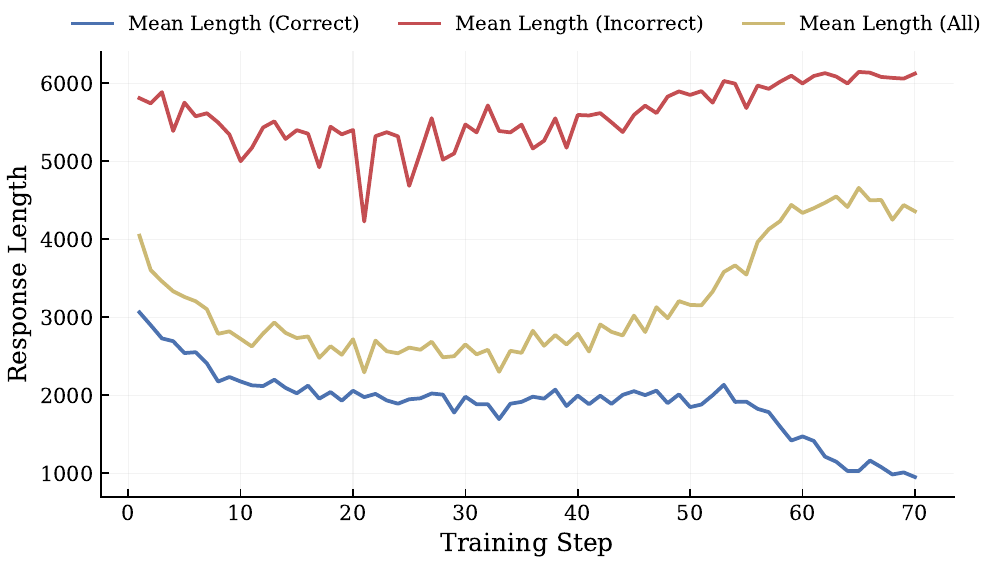}
  \captionof{figure}{Training trajectory of Qwen3-4B under the short reward.}
  \label{fig:traj_4b_short_inline}
  \vspace{-1.5em}
\end{wrapfigure}

We train Qwen3-1.7B, Qwen3-4B, and Qwen3-8B, tracking five training dynamics throughout the training process. Taking Qwen3-4B under short-reward training as a representative example, we find that training initially succeeds in reducing reasoning length while maintaining correctness. However, beyond this initial point the training dynamics drift noticeably. Mean reasoning length begins to increase counterintuitively, driven primarily by the growing length of incorrect responses. Specifically, under short-reward training, the model is encouraged to produce shorter outputs for both correct and incorrect answers. This compression signal provides no explicit incentive to preserve progress-making action toward valid solutions. Consequently, we conjecture that continued optimization impairs the model's ability to complete valid solution trajectories, leading to invalid, incoherent, or degenerate responses on an increasing number of questions. This contradiction between the intended objective and the actual observed behavior on questions is what drives the observed degradation. By contrast, long-reward training successfully increases reasoning length. However, comparing model outputs for the same questions across different training stages, we observe that the later increase in reasoning length tends to coincide with a higher frequency of cyclic actions, suggesting that these additional tokens may reflect repetitive and redundant reasoning patterns rather than deeper or more productive exploration.

\section{Methodology}
\label{sec:method}

Section~\ref{sec:reasoning_length_accuracy} exposes limitations of existing length-reward approaches in mitigating length misallocation. To this end, we estimate whether longer or shorter reasoning is currently more beneficial for each question at each training step. This estimate is then converted into a signed length term within the reward. The resulting reward integrates directly into GRPO without modifying the training pipeline.

\subsection{Contrastive Accuracy--Length Estimation}
\label{sec:length_direction}

Given a question \(q\), we compare whether longer or shorter responses are currently more beneficial by comparing accuracy between the shorter and longer halves of the sampled responses. We sample a group of \(G\) responses \(\{o_i\}_{i=1}^{G}\) from the old policy \(\pi_{\theta_{\mathrm{old}}}\). Let \(y_i \in \{0,1\}\) denote the correctness of response \(o_i\), and let \(L_i\) denote its length. We sort the group by length in ascending order:
\begin{equation}
\label{eq:length_sort}
    L_{(1)} \leq L_{(2)} \leq \cdots \leq L_{(G)}
\end{equation}
where \((i)\) denotes the index after sorting. Assuming \(G\) is even, we split the sorted responses into a shorter half and a longer half:
\begin{equation}
\label{eq:length_partition}
\mathcal{S}(q) = \{(1), \ldots, (G/2)\}, \qquad
\mathcal{L}(q) = \{(G/2+1), \ldots, (G)\}
\end{equation}
We then compute the average accuracy of each half:
\begin{equation}
\label{eq:short_long_acc}
    \bar{y}_{\mathcal{S}}(q)
    = \frac{1}{|\mathcal{S}(q)|}\sum_{i \in \mathcal{S}(q)} y_i,
    \qquad
    \bar{y}_{\mathcal{L}}(q)
    = \frac{1}{|\mathcal{L}(q)|}\sum_{i \in \mathcal{L}(q)} y_i
\end{equation}
Based on this comparison, we define a question-level length adjustment direction:
\begin{equation}
\label{eq:length_direction}
    d(q) =
    \begin{cases}
        -1, & \bar{y}_{\mathcal{S}}(q) > \bar{y}_{\mathcal{L}}(q) \\[4pt]
        \;\;0, & \bar{y}_{\mathcal{S}}(q) = \bar{y}_{\mathcal{L}}(q) \\[4pt]
        +1, & \bar{y}_{\mathcal{S}}(q) < \bar{y}_{\mathcal{L}}(q)
    \end{cases}
\end{equation}

The adjustment direction signal \(d(q)\) encodes whether extended reasoning is currently beneficial \((+1)\), redundant \((-1)\), or uninformative \((0)\) for a given question \(q\). A value of \(+1\) indicates that longer responses are empirically more accurate; \(-1\) indicates that concise responses outperform and that length compression is appropriate; \(0\) indicates no clear contrastive signal, in which case no length adjustment is applied. Since \(d(q)\) is estimated directly from the sampled response group, it incurs no additional inference cost and naturally adapts as the policy updates during training.

\subsection{Direction-Conditioned Reward}
\label{sec:reward}
We convert the question-level directional signal into a reward by combining answer correctness with a signed length term. For each response \(o_i\), we define
\begin{equation}
\label{eq:direction_conditioned_reward}
    r(q, o_i) = y_i + d(q) \cdot \frac{L_i}{L_{\max}}
\end{equation}
Equivalently,
\begin{equation}
\label{eq:direction_conditioned_reward_piecewise}
    r(q, o_i) =
    \begin{cases}
        y_i - \dfrac{L_i}{L_{\max}},
            & \bar{y}_{\mathcal{S}}(q) > \bar{y}_{\mathcal{L}}(q) \\[6pt]
        y_i,
            & \bar{y}_{\mathcal{S}}(q) = \bar{y}_{\mathcal{L}}(q) \\[6pt]
        y_i + \dfrac{L_i}{L_{\max}},
            & \bar{y}_{\mathcal{S}}(q) < \bar{y}_{\mathcal{L}}(q)
    \end{cases}
\end{equation}
where \(y_i \in \{0,1\}\) denotes the correctness of the generated response \(o_i\), \(L_i\) denotes its corresponding length, and \(L_{\max}\) is the maximum reasoning length used during training.

The signed length term \(d(q)\cdot L_i/L_{\max}\) converts the directional signal into a reward component: it encourages longer outputs when extended reasoning improves accuracy, penalizes length when shorter reasoning suffices, and vanishes when no directional preference is detected. Crucially, the correctness reward \(y_i\) remains the principal term, while the length component serves only as a secondary shaping signal, ensuring that the model does not over-optimize length and thereby impair correctness.

\subsection{Integration with GRPO}
\label{sec:grpooptimization}

We integrate \(r(q, o_i)\) directly into GRPO (Section~\ref{sec:grpo}).
The group-normalized advantage \(\hat{A}_i\) is computed according to Eq.~\eqref{eq:grpo_advantage}, and the policy is updated by maximizing \(\mathcal{L}_{\mathrm{GRPO}}(\theta)\) in Eq.~\eqref{eq:grpo_objective}. No other component of the training pipeline is modified.

In summary, our method assumes no fixed relationship between question difficulty and appropriate reasoning length. Instead, the beneficial length adjustment of a given question \(q\) is estimated online from sampled responses at no additional inference cost and without introducing extra hyperparameters. The reward adaptively encourages longer reasoning when it improves accuracy, penalizes it when concise responses suffice, and remains neutral otherwise, thereby keeping length control aligned with what is beneficial for each question throughout training in a lightweight manner.

\section{Experiments}
\label{sec:experiment}


To validate the effectiveness of CARE in allocating reasoning length, we conduct experiments on Qwen3-1.7B, 4B and 8B across four mathematical reasoning benchmarks under the same training setup, comparing against the base model, GRPO, and two adaptive length-control baselines.

\subsection{Experimental Setup}
\label{sec:experimental_setup}
\paragraph{Training.}
All experiments are conducted using the VERL~\citep{sheng2025hybridflow} framework. We train Qwen3-1.7B, 4B and 8B on MATH~\citep{hendrycks2021math} with the AdamW~\citep{loshchilov2017decoupled} optimizer at a learning rate of \(1 \times 10^{-6}\) and a batch size of 256 questions per step. For each question, 16 responses are sampled at a temperature of \(1.0\) and top-p of \(1.0\), with the maximum reasoning length \(6{,}144\) tokens. The model parameters are then updated every 256 sampled responses. The KL penalty coefficient \(\beta\) and clipping threshold \(\varepsilon\) are set to \(0.001\) and \(0.2\), respectively.

\paragraph{Baselines.} We compare CARE against four methods: 
(1)~the Qwen3 base model without RL training; 
(2)~GRPO~\citep{shao2024deepseekmath} with a pure correctness reward \(r=y\); 
and (3)~two adaptive length-control baselines,~ALP~\citep{xiang2025alp} and~DAST~\citep{shen2025dast}.

\paragraph{Evaluation.}
We evaluate on four mathematical reasoning benchmarks: AIME 2025~\citep{balunovic_srimatharena_2025}, AIME 2026~\citep{balunovic_srimatharena_2025}, HMMT 2026~\citep{balunovic_srimatharena_2025},\footnote{HMMT Feb 2026 is sourced from \url{https://huggingface.co/datasets/MathArena/hmmt_feb_2026}.} and MATH500~\citep{lightman2023verify}. For each question, we sample 32 responses with temperature \(0.6\), top-p \(0.95\), and a maximum reasoning length of \(16{,}384\) tokens. Pass@1 is then adopted to evaluate the model's reasoning capability, computed using the unbiased estimator for Pass\(@k\) proposed in~\citet{chen2021codex}, where \(k=1\). We also report reasoning length in tokens. The two metrics reflect whether length misallocation has been effectively mitigated: a method that allocates reasoning length well should achieve higher accuracy without inflating token consumption.

\subsection{Result}
\label{sec:main_results}
Table~\ref{tab:main} presents the main results across multiple reasoning benchmarks under different reward designs. Experiments show that CARE achieves the best balance between accuracy and reasoning length across multiple benchmarks, attaining higher Pass@1 while avoiding unnecessary token expenditure.

On Qwen3-1.7B, CARE achieves \(43.82\%\) average Pass@1 with \(8{,}154\) tokens, outperforming GRPO by \(2.64\%\) while reducing token usage by \(9.9\%\). Compared with DAST, the improvement reaches \(5.23\%\) with \(31.7\%\) fewer tokens. The improvement is particularly evident on HMMT~2026, where CARE reaches \(21.78\%\) Pass@1, compared with \(15.81\%\) for DAST and \(17.61\%\) for GRPO.

On Qwen3-4B, CARE outperforms GRPO by \(2.62\%\) in Pass@1 with \(11.1\%\) fewer tokens, while achieving a \(4.68\%\) improvement and \(37.7\%\) token reduction over the base model. On AIME~2026, CARE achieves \(55.62\%\) Pass@1, exceeding ALP and DAST by \(5.20\%\) and \(18.85\%\), respectively.

On Qwen3-8B, CARE maintains a strong overall accuracy--efficiency trade-off, achieving \(59.62\%\) average Pass@1 with \(7{,}707\) tokens. Compared with ALP, it improves average Pass@1 by \(0.83\%\) while reducing token usage by \(7.0\%\). Although competing methods occasionally perform better on individual benchmarks, CARE consistently achieves the highest average Pass@1 with the lowest average token usage across all three model sizes.

\begin{table}[h]
\caption{Results on reasoning benchmarks under different methods. For each benchmark, we report Pass@1 (\%) and length (Token).}
\label{tab:main}
\centering
\small
\setlength{\tabcolsep}{3pt}
\renewcommand{\arraystretch}{1.15}
\resizebox{\textwidth}{!}{%
\begin{tabular}{@{}l cc cc cc cc cc@{}}
\toprule
\multirow{2}{*}{Method} &
\multicolumn{2}{c}{AIME 2025} &
\multicolumn{2}{c}{AIME 2026} &
\multicolumn{2}{c}{HMMT 2026} &
\multicolumn{2}{c}{MATH500} &
\multicolumn{2}{c}{Average} \\
\cmidrule(lr){2-3}
\cmidrule(lr){4-5}
\cmidrule(lr){6-7}
\cmidrule(lr){8-9}
\cmidrule(lr){10-11}
& Pass@1 & Token
& Pass@1 & Token
& Pass@1 & Token
& Pass@1 & Token
& Pass@1 & Token \\
\midrule
\multicolumn{11}{@{}c@{}}{\textbf{\textsl{Qwen3-1.7B}}}\\
\addlinespace[2pt]
\midrule
Qwen3-1.7B
& 26.88 & 13320.41
& 30.21 & 13716.90
& 16.67 & 14062.21
& 86.47 & 5246.36
& 40.06 & 11586.47 \\
GRPO
& 30.42 & 10826.55
& 28.96 & 10245.45
& 17.61 & 11296.01
& \textbf{87.74} & 3829.58
& 41.18 & 9049.40 \\
ALP
& 31.87 & 11623.52
& 32.08 & 11789.75
& 19.41 & 12170.17
& 85.76 & 3863.09
& 42.28 & 9861.63 \\
DAST
& 26.04 & 14052.29
& 29.48 & 14312.01
& 15.81 & 14639.91
& 83.03 & 4746.27
& 38.59 & 11937.62 \\
\rowcolor{rowhl}
\textbf{CARE}
& \textbf{33.96} & \textbf{9352.41}
& \textbf{32.29} & \textbf{10073.94}
& \textbf{21.78} & \textbf{10176.61}
& 87.25 & \textbf{3012.70}
& \textbf{43.82} & \textbf{8153.92} \\
\midrule
\multicolumn{11}{@{}c@{}}{\textbf{\textsl{Qwen3-4B}}}\\
\addlinespace[2pt]
\midrule
Qwen3-4B
& 46.46 & 12967.93
& 50.83 & 12692.93
& 23.30 & 13816.37
& 90.21 & 4833.99
& 52.70 & 11077.81 \\
GRPO
& 48.85 & 9334.90
& 52.60 & 8657.44
& 26.04 & 10112.92
& 91.54 & 2962.31
& 54.76 & 7766.89 \\
ALP
& 50.83 & 9012.51
& 50.42 & 8231.33
& 27.84 & 9765.89
& 90.61 & \textbf{2547.05}
& 54.93 & 7389.20 \\
DAST
& 32.50 & 13973.33
& 36.77 & 14032.09
& 20.36 & 14566.25
& 86.74 & 4670.34
& 44.09 & 11810.50 \\
\rowcolor{rowhl}
\textbf{CARE}
& \textbf{52.92} & \textbf{8455.09}
& \textbf{55.62} & \textbf{7817.14}
& \textbf{28.60} & \textbf{8735.88}
& \textbf{92.39} & 2618.05
& \textbf{57.38} & \textbf{6906.54} \\
\midrule
\multicolumn{11}{@{}c@{}}{\textbf{\textsl{Qwen3-8B}}}\\
\addlinespace[2pt]
\midrule
Qwen3-8B
& 45.73 & 13176.58
& 52.19 & 12623.87
& 25.66 & 13925.07
& 88.07 & 5073.17
& 52.91 & 11199.67 \\
GRPO
& \textbf{55.42} & 10029.52
& 56.46 & 9246.88
& 32.10 & 11185.89
& 90.51 & 3022.00
& 58.62 & 8371.07 \\
ALP
& 54.27 & 9854.36
& 57.92 & 9240.24
& 32.48 & 11088.53
& 90.50 & 2983.70
& 58.79 & 8291.71 \\
DAST
& 51.67 & 11288.34
& 56.56 & 10643.80
& 29.45 & 12502.82
& 90.30 & 2965.53
& 57.00 & 9350.12 \\
\rowcolor{rowhl}
\textbf{CARE}
& 53.85 & \textbf{9166.40}
& \textbf{58.85} & \textbf{8548.70}
& \textbf{32.67} & \textbf{10310.60}
& \textbf{93.10} & \textbf{2803.10}
& \textbf{59.62} & \textbf{7707.20} \\
\bottomrule
\end{tabular}%
}
\end{table}

\subsection{Further Analyses}
\label{sec:analyses}
\paragraph{Stability of the Length Preference Signal.}

To evaluate the stability of CARE's length preference signal under different rollout group sizes, we use \(d_{16}(q)\) computed from the full group \(g=16\) as the reference and randomly subsample \(g\in\{4,8,12\}\) responses without replacement. For each subset, we recompute the short--long split and \(d_g(q)\), repeat the procedure \(20\) times at every training step, and report exact agreement over the first, middle, and final thirds of the training steps, together with non-zero agreement when \(d_{16}(q)\neq0\). As shown in Table~\ref{tab:signal_stability}, both exact agreement and non-zero agreement increase consistently with group size, while the exact agreement remains stable across training stages, indicating that CARE is not sensitive to the rollout group size.

\begin{table}[h]
\caption{Stability of CARE's length preference signal under different rollout group sizes. Exact agreement measures consistency between the subsampled signal \(d_g(q)\) and the full-group reference \(d_{16}(q)\). Non-zero agreement is computed only when \(d_{16}(q)\neq 0\).}
\label{tab:signal_stability}
\centering
\small
\setlength{\tabcolsep}{5pt}
\renewcommand{\arraystretch}{0.85}
\resizebox{\textwidth}{!}{%
\begin{tabular}{@{}l c c c c c c@{}}
\toprule
Model & \(g\) & Early Exact & Middle Exact & Late Exact & Overall Exact & Overall Non-Zero \\
\midrule
\multirow{3}{*}{Qwen3-1.7B} & 4  & 85.39\% & 87.19\% & 87.52\% & \textbf{86.69\%} & 50.90\% \\
                            & 8  & 91.95\% & 92.81\% & 92.81\% & \textbf{92.52\%} & 73.73\% \\
                            & 12 & 96.27\% & 96.23\% & 96.40\% & \textbf{96.30\%} & \textbf{87.91\%} \\
\midrule
\multirow{3}{*}{Qwen3-4B}   & 4  & 89.85\% & 90.67\% & 91.62\% & \textbf{90.69\%} & 50.94\% \\
                            & 8  & 94.63\% & 94.73\% & 94.99\% & \textbf{94.78\%} & 73.82\% \\
                            & 12 & 97.39\% & 97.44\% & 97.40\% & \textbf{97.41\%} & \textbf{87.92\%} \\
\midrule
\multirow{3}{*}{Qwen3-8B}   & 4  & 87.73\% & 91.27\% & 90.89\% & \textbf{89.96\%} & 51.55\% \\
                            & 8  & 93.50\% & 95.52\% & 95.18\% & \textbf{94.73\%} & 75.32\% \\
                            & 12 & 96.94\% & 98.05\% & 97.51\% & \textbf{97.50\%} & \textbf{88.80\%} \\
\bottomrule
\end{tabular}%
}
\end{table}

\paragraph{Effect of Accuracy-Gap Magnitude Weighting.}

We examine whether incorporating the magnitude of the short--long accuracy difference improves CARE by weighting the length term with \(\left|\bar{y}_{\mathcal{L}}-\bar{y}_{\mathcal{S}}\right|\). We compare this CARE-Magnitude variant with the original CARE across Qwen3-1.7B, 4B and 8B on all four benchmarks, reporting Pass@1 and average reasoning length. Table~\ref{tab:magnitude_ablation} shows that the magnitude variant achieves slightly higher Pass@1 on some individual benchmarks, but the original CARE consistently achieves better average Pass@1 with fewer tokens across all three model sizes.

\begin{table}[h]
\caption{Ablation on accuracy-gap magnitude weighting. Each entry reports Pass@1 (\%) / average response length (Token). CARE-Magnitude weights the length term by \(\left|\bar{y}_{\mathcal{L}}-\bar{y}_{\mathcal{S}}\right|\).}
\label{tab:magnitude_ablation}
\centering
\small
\setlength{\tabcolsep}{3pt}
\renewcommand{\arraystretch}{1.1}
\resizebox{\textwidth}{!}{%
\begin{tabular}{@{}l l c c c c c@{}}
\toprule
Model & Method & AIME25 & AIME26 & HMMT26 & MATH500 & Average \\
\midrule
\multirow{2}{*}{Qwen3-1.7B} & CARE-Magnitude & 31.35/10464.91 & 30.73/10496.70 & 20.83/11380.66 & 85.08/3314.91 & 42.00/8914.30 \\
                            & CARE           & \textbf{33.96/9352.41} & \textbf{32.29/10073.94} & \textbf{21.78/10176.61} & \textbf{87.25/3012.70} & \textbf{43.82/8153.92} \\
\midrule
\multirow{2}{*}{Qwen3-4B}   & CARE-Magnitude & \textbf{53.02/8843.30} & 51.98/8117.22 & 26.99/9450.59 & 90.59/2718.12 & 55.65/7282.31 \\
                            & CARE           & 52.92/8455.09 & \textbf{55.62/7817.14} & \textbf{28.60/8735.88} & \textbf{92.39/2618.05} & \textbf{57.38/6906.54} \\
\midrule
\multirow{2}{*}{Qwen3-8B}   & CARE-Magnitude & \textbf{54.17/9583.34} & \textbf{59.06/8874.49} & 32.01/10569.64 & 90.65/2892.74 & 58.97/7980.05 \\
                            & CARE           & 53.85/9166.40 & 58.85/8548.70 & \textbf{32.67/10310.60} & \textbf{93.10/2803.10} & \textbf{59.62/7707.20} \\
\bottomrule
\end{tabular}%
}
\end{table}
\paragraph{Sensitivity to the Length-Term Coefficient.}

Although CARE uses a fixed unit coefficient for the length term by default, we introduce \(\lambda\) here solely for sensitivity analysis, where \(\lambda=1\) corresponds to the original CARE. We evaluate a smaller coefficient, \(\lambda=0.5\), across Qwen3-1.7B, Qwen3-4B and Qwen3-8B using the same four benchmarks and evaluation metrics. The results in Table~\ref{tab:lambda_ablation} indicate that CARE remains effective with \(\lambda=0.5\), while the default \(\lambda=1\) consistently provides a better average accuracy--efficiency trade-off across all model sizes.

\begin{table}[H]
\caption{Ablation on the length-term coefficient. Each entry reports Pass@1 (\%) / average reasoning length (Token). We compare the default CARE setting \(\lambda=1\) with \(\lambda=0.5\).}
\label{tab:lambda_ablation}
\centering
\small
\setlength{\tabcolsep}{3pt}
\renewcommand{\arraystretch}{1.1}
\resizebox{\textwidth}{!}{%
\begin{tabular}{@{}l l c c c c c@{}}
\toprule
Model & Method & AIME25 & AIME26 & HMMT26 & MATH500 & Average \\
\midrule
\multirow{2}{*}{Qwen3-1.7B} & CARE (\(\lambda=0.5\)) & 32.81/10627.63 & 30.31/10700.56 & 20.55/11500.73 & 85.25/3397.02 & 42.23/9056.49 \\
                            & CARE                    & \textbf{33.96/9352.41} & \textbf{32.29/10073.94} & \textbf{21.78/10176.61} & \textbf{87.25/3012.70} & \textbf{43.82/8153.92} \\
\midrule
\multirow{2}{*}{Qwen3-4B}   & CARE (\(\lambda=0.5\)) & \textbf{53.02/8555.19} & 51.25/8218.75 & 27.27/9179.88 & 90.84/2742.92 & 55.60/7174.18 \\
                            & CARE                    & 52.92/8455.09 & \textbf{55.62/7817.14} & \textbf{28.60/8735.88} & \textbf{92.39/2618.05} & \textbf{57.38/6906.54} \\
\midrule
\multirow{2}{*}{Qwen3-8B}   & CARE (\(\lambda=0.5\)) & \textbf{54.06/9674.28} & 56.67/8993.24 & \textbf{32.95/10813.47} & 90.63/3004.62 & 58.58/8121.40 \\
                            & CARE                    & 53.85/9166.40 & \textbf{58.85/8548.70} & 32.67/10310.60 & \textbf{93.10/2803.10} & \textbf{59.62/7707.20} \\
\bottomrule
\end{tabular}%
}
\end{table}

\section{Conclusion}

In this work, we propose CARE, a lightweight method for reasoning length control in LLMs. Empirical analysis reveals that reasoning length predominantly influences accuracy on partially solvable questions, and that naive length rewards induce detrimental training effects. CARE addresses these by comparing the beneficial length adjustment per question from online sampled responses and integrating it into GRPO at no additional overhead. Experiments demonstrate that CARE improves Pass@1 while reducing reasoning length, exhibits stable length-preference signals, and maintains a strong accuracy--efficiency trade-off under different weighting and coefficient settings.

\newpage

\bibliographystyle{iclr2027_conference}
\bibliography{iclr2027_conference}

\newpage

\appendix

\section{Experimental Details for Reasoning Length and Accuracy}
\label{app:experimental_details}
 
The following provides the shared experimental configurations for Section~\ref{sec:reasoning_length_accuracy}.
 
\subsection{Sampling Configuration}
\label{app:sampling_configuration}
The sampling configuration is summarized in Table~\ref{tab:sampling-hparams}. We use
\(64\) samples per question to obtain a stable per-question difficulty estimate and to ensure sufficient coverage across different reasoning lengths. The bin width of \(0.5\) 
is chosen to balance granularity with statistical reliability within each bin. Temperature and top-p are both set to \(1.0\) to ensure diversity in reasoning lengths.

\begin{table}[H]
  \vspace{-1em}
  \caption{Sampling configuration for difficulty estimation on GSM8K and MATH500.}
  \label{tab:sampling-hparams}
  \centering
  \footnotesize
  \setlength{\tabcolsep}{6.0pt}
  \renewcommand{\arraystretch}{1.05}
  \begin{tabular}{@{}cccccc@{}}
  \toprule
  Inference Engine & Samples per Question & Temperature & Top-p & Max Reasoning Length &  Bin Width \\
  \midrule
  vLLM & 64 & 1.0 & 1.0 & 16{,}384 & 0.5 \\
  \bottomrule
  \end{tabular}
\end{table}

\subsection{Filtering Protocol and Difficulty Composition}
\label{app:filtering_protocol_question_retention}

For z-score normalization, we exclude questions with fewer than three correct samples. Table~\ref{tab:zscore_filtering} reports the number and proportion of retained questions for each model and dataset.

\begin{table}[H]
  \vspace{-1em}
  \caption{Number of questions retained after z-score filtering on GSM8K and MATH500.}
  \label{tab:zscore_filtering}
  \centering
  \footnotesize
  \setlength{\tabcolsep}{12pt}
  \renewcommand{\arraystretch}{1.15}
  \begin{tabular}{@{}l ccc@{}}
  \toprule
  & Qwen3-1.7B & Qwen3-4B & Qwen3-8B \\
  \midrule
  GSM8K   & 1286/1319 (97.50\%) & 1277/1319 (96.82\%) & 1292/1319 (97.95\%) \\
  MATH500 & 475/500 (95.00\%) & 474/500 (94.80\%) & 479/500 (95.80\%) \\
  \bottomrule
  \end{tabular}
\end{table}

For completeness, before applying the correct-response filtering
criterion, we additionally report the fine-grained composition of the hard and easy difficulty subsets used in Section~\ref{sec:length_sensitivity}. Let \(p\) denote the per-question accuracy under the same sampling configuration. The hard difficulty subset \([0,0.25]\) is divided into \(p=0\) and \(0<p\leq0.25\), while the easy difficulty subset \((0.75,1]\) is divided into \(0.75<p<1\) and \(p=1\).

\begin{table}[H]
\caption{Fine-grained composition of the hard and easy difficulty subsets. Each entry reports the number of questions (fraction within the corresponding interval).}
\label{tab:difficulty_decomposition}
\centering
\footnotesize
\setlength{\tabcolsep}{5pt}
\renewcommand{\arraystretch}{1.0}
\begin{tabular*}{\textwidth}{@{\extracolsep{\fill}}llcccc@{}}
\toprule
\multirow{2}{*}{Model} &
\multirow{2}{*}{Dataset} &
\multicolumn{2}{c}{Hard \([0,0.25]\)} &
\multicolumn{2}{c}{Easy \((0.75,1]\)} \\
\cmidrule(lr){3-4}\cmidrule(lr){5-6}
& & \(p=0\) & \(0<p\leq0.25\) & \(0.75<p<1\) & \(p=1\) \\
\midrule
\multirow{2}{*}{Qwen3-1.7B}
& GSM8K   & 33 (30.56\%) & 75 (\textbf{69.44\%}) & 179 (\textbf{15.85\%}) & 950 (84.15\%) \\
& MATH500 & 25 (36.76\%) & 43 (\textbf{63.24\%}) & 112 (\textbf{28.28\%}) & 284 (71.72\%) \\
\midrule
\multirow{2}{*}{Qwen3-4B}
& GSM8K   & 42 (41.18\%) & 60 (\textbf{58.82\%}) & 184 (\textbf{15.62\%}) & 994 (84.38\%) \\
& MATH500 & 26 (40.62\%) & 38 (\textbf{59.38\%}) & 107 (\textbf{25.72\%}) & 309 (74.28\%) \\
\midrule
\multirow{2}{*}{Qwen3-8B}
& GSM8K   & 23 (30.26\%) & 53 (\textbf{69.74\%}) & 70 (\textbf{5.75\%}) & 1148 (94.25\%) \\
& MATH500 & 21 (40.38\%) & 31 (\textbf{59.62\%}) & 68 (\textbf{16.00\%}) & 357 (84.00\%) \\
\bottomrule
\end{tabular*}
\end{table}

\subsection{Length--Accuracy Relationships Across Model Families}
\label{app:cross_family_length_accuracy}

We additionally evaluate Gemma-4-E2B-it, Gemma-4-E4B-it, and DeepSeek-R1-Distill-Llama-8B on GSM8K and MATH500, following the same experimental procedure as in Section~\ref{sec:length_sensitivity}. For each model, we sample \(64\) responses per question and normalize response lengths using the mean and standard deviation of its correct responses. We then report the average accuracy within each normalized length interval for every difficulty subset.

\begin{figure}[t]
    \centering
    \includegraphics[width=1.0\textwidth]{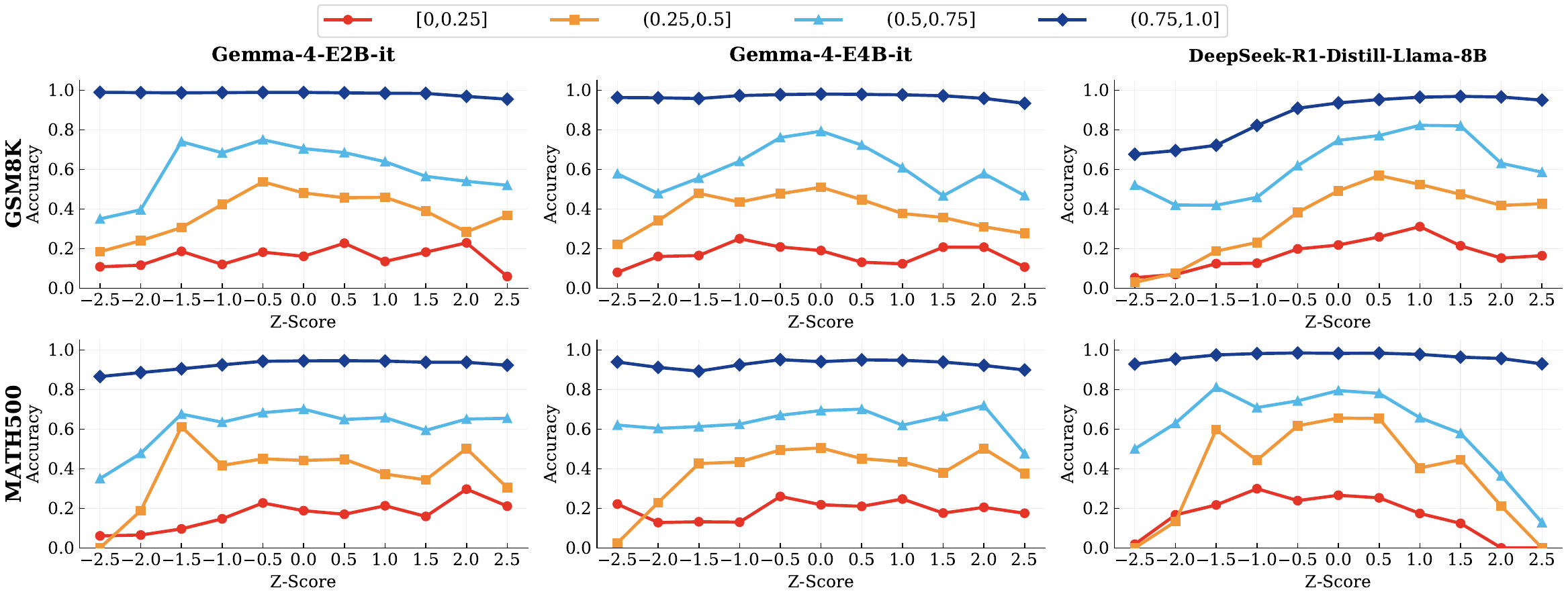}
    \caption{Relationship between reasoning length and accuracy across difficulty groups on GSM8K (top) and MATH500 (bottom) for Gemma-4-E2B-it, Gemma-4-E4B-it, and DeepSeek-R1-Distill-Llama-8B.}
    \label{fig:cross_family_length_accuracy}
    \vspace{-1.5em}
\end{figure}

 Figure~\ref{fig:cross_family_length_accuracy} shows that the length–accuracy relationships for the Gemma-4 and DeepSeek-R1-Distill-Llama models are broadly consistent with those observed in Section~\ref{sec:length_sensitivity}, although these patterns are not equally evident across models and difficulty subsets. 

\subsection{Training Configuration}
\label{app:training_configuration}
The training setup is listed in Table~\ref{tab:training-hparams}. We adopt GRPO as the advantage estimator with a sampling group size of 16 per question. The global batch size is set to \(256\), meaning that \(256\) distinct questions are sampled at each training step, generating a total of \(256 \times 16 = 4{,}096\) responses. The mini batch size is also set to  \(256\), meaning that the model updates its parameters once every  \(256\) collected responses, mitigating the risk of converging to local optima.

\begin{table}[H]
  \vspace{-1em}
  \caption{Hyperparameters for training.}
  \label{tab:training-hparams}
  \centering
  \footnotesize
  \setlength{\tabcolsep}{2.0pt}
  \renewcommand{\arraystretch}{1.05}
  \begin{tabular}{@{}cccccccccc@{}}
  \toprule
  \shortstack[c]{Advantage\\Estimator} &
  \shortstack[c]{Training\\Temperature} &
  \shortstack[c]{Global\\Batch Size} &
  \shortstack[c]{Mini\\Batch Size} &
  \shortstack[c]{Rollout\\Number} &
  \shortstack[c]{Regularization\\~} &
  \shortstack[c]{Actor Clip\\ Ratio} &
  \shortstack[c]{Max Response\\Length} &
  \shortstack[c]{Learning\\Rate} \\
  \midrule
  GRPO & 1.0 & 256 & 256 & 16 & KL/Entropy & 0.2 & 6144 & 1e-6 \\
  \bottomrule
\end{tabular}
\end{table}

\subsection{Evaluation Protocol}
\label{app:evaluation_protocol}

The evaluation settings are reported in Table~\ref{tab:evaluation-hparams}. We generate \(32\) responses per question at temperature \(0.6\) and top-p \(0.95\), with the maximum reasoning length set to \(16{,}384\) tokens. 

\begin{table}[H]
  \vspace{-1em}
  \caption{Evaluation configuration.}
  \label{tab:evaluation-hparams}
  \centering
  \footnotesize
  \setlength{\tabcolsep}{6.0pt}
  \renewcommand{\arraystretch}{1.05}
  \begin{tabular}{@{}ccccc@{}}
  \toprule
  Samples per Question & Temperature & Top-p & Max Reasoning Length & Aggregation \\
  \midrule
  32 & 0.6 & 0.95 & 16{,}384 & Mean over samples \\
  \bottomrule
  \end{tabular}
\end{table}

\section{Accuracy Misalignment Analysis}
\label{app:accuracy_misalignment_results}

Here we describe the visualization protocol and present full results for all model–reward combinations.

\subsection{Heatmap Visualization Protocol}
\label{app:heatmap_visualization}
Each difficulty subset, split by base-model accuracy, is visualized as a density heatmap on a shared 8\(\times\)8 grid. The horizontal axis (\(\Delta\)Reasoning length) measures the change in token relative to the base model for the same question, and the vertical axis (\(\Delta\)Accuracy) measures the corresponding change in per-question accuracy. Each cell reports the within-bin fraction of questions. Darker cells indicate a higher concentration of questions in that region. All heatmaps share the following density scale:
\begin{figure}[H]
    \centering
    \includegraphics[width=1.0\textwidth]{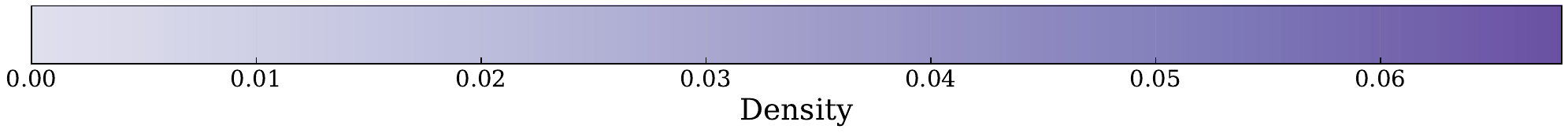}
    \vspace{-1.5em}
\end{figure}
 
\subsection{Qwen3-1.7B}
\label{app:heatmap_results}

\begin{figure}[H]
    \centering
    \begin{subfigure}[t]{0.49\linewidth}
        \centering
        \includegraphics[width=\linewidth]{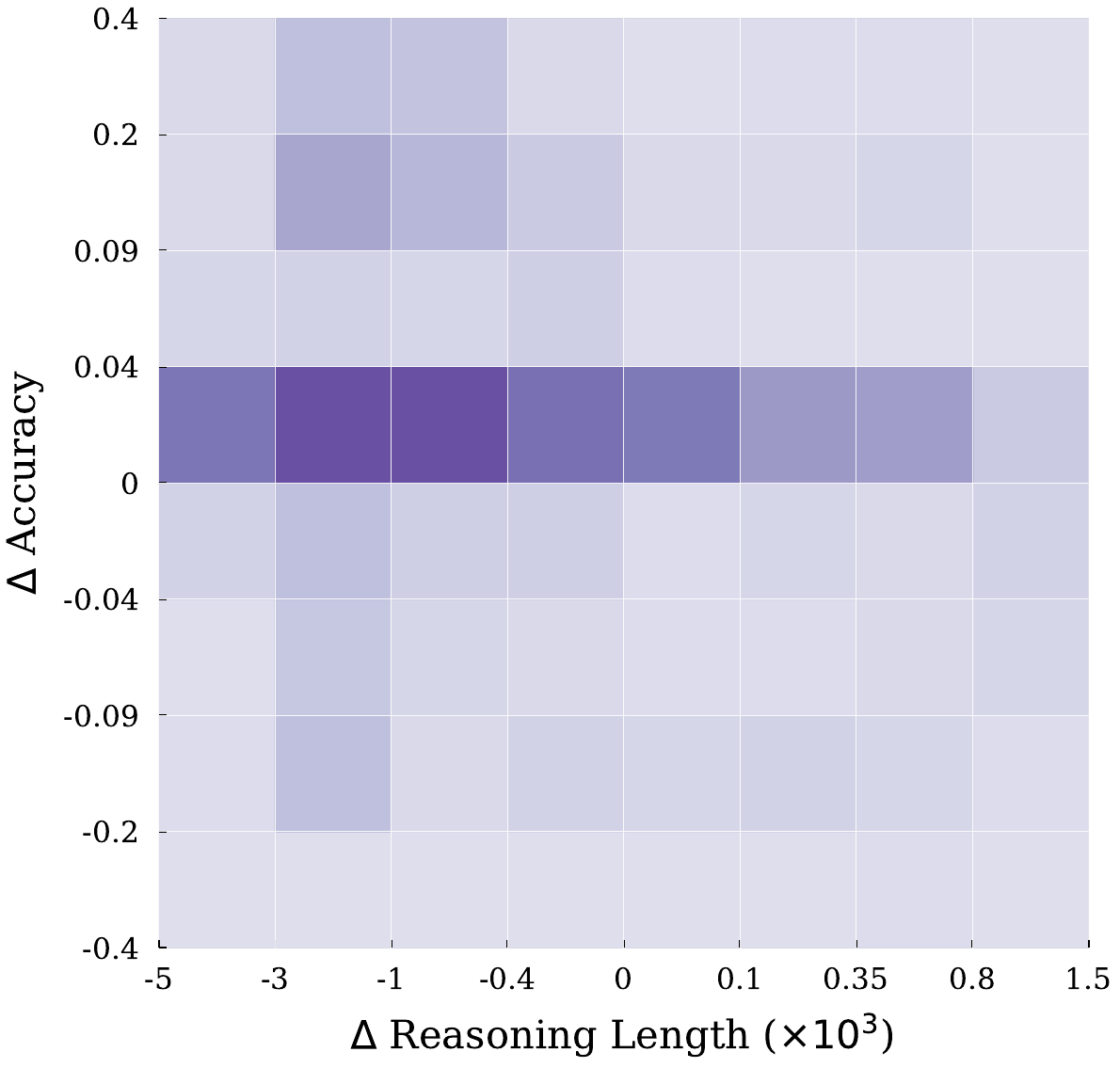}
        \caption{Difficulty Subset [0, 0.25]}
        \label{fig:heatmap_1.7b_short_1}
    \end{subfigure}
    \hfill
    \begin{subfigure}[t]{0.49\linewidth}
        \centering
        \includegraphics[width=\linewidth]{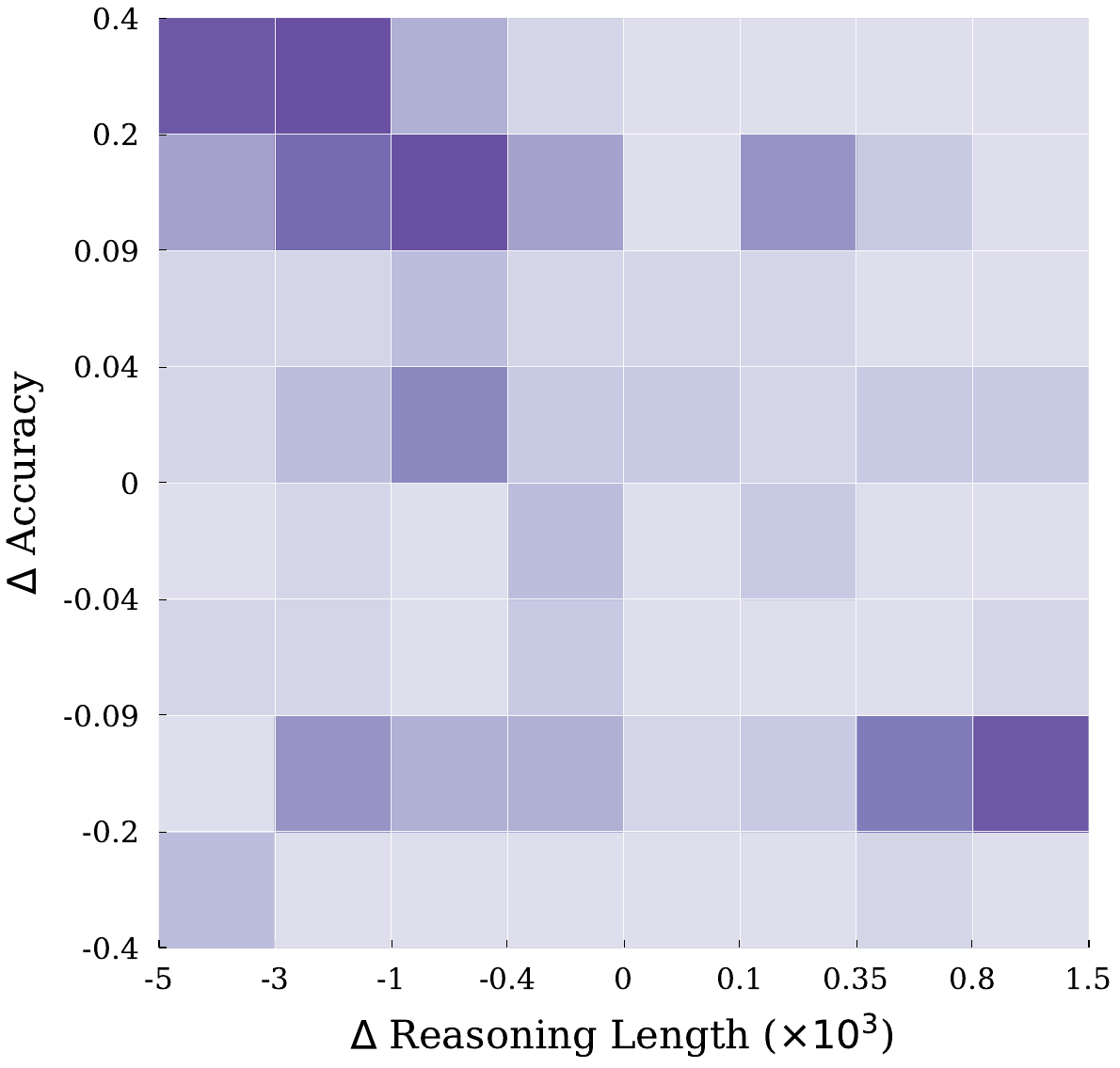}
        \caption{Difficulty Subset (0.25, 0.5]}
        \label{fig:heatmap_1.7b_short_2}
    \end{subfigure}
 
    \vspace{1em}
 
    \begin{subfigure}[t]{0.49\linewidth}
        \centering
        \includegraphics[width=\linewidth]{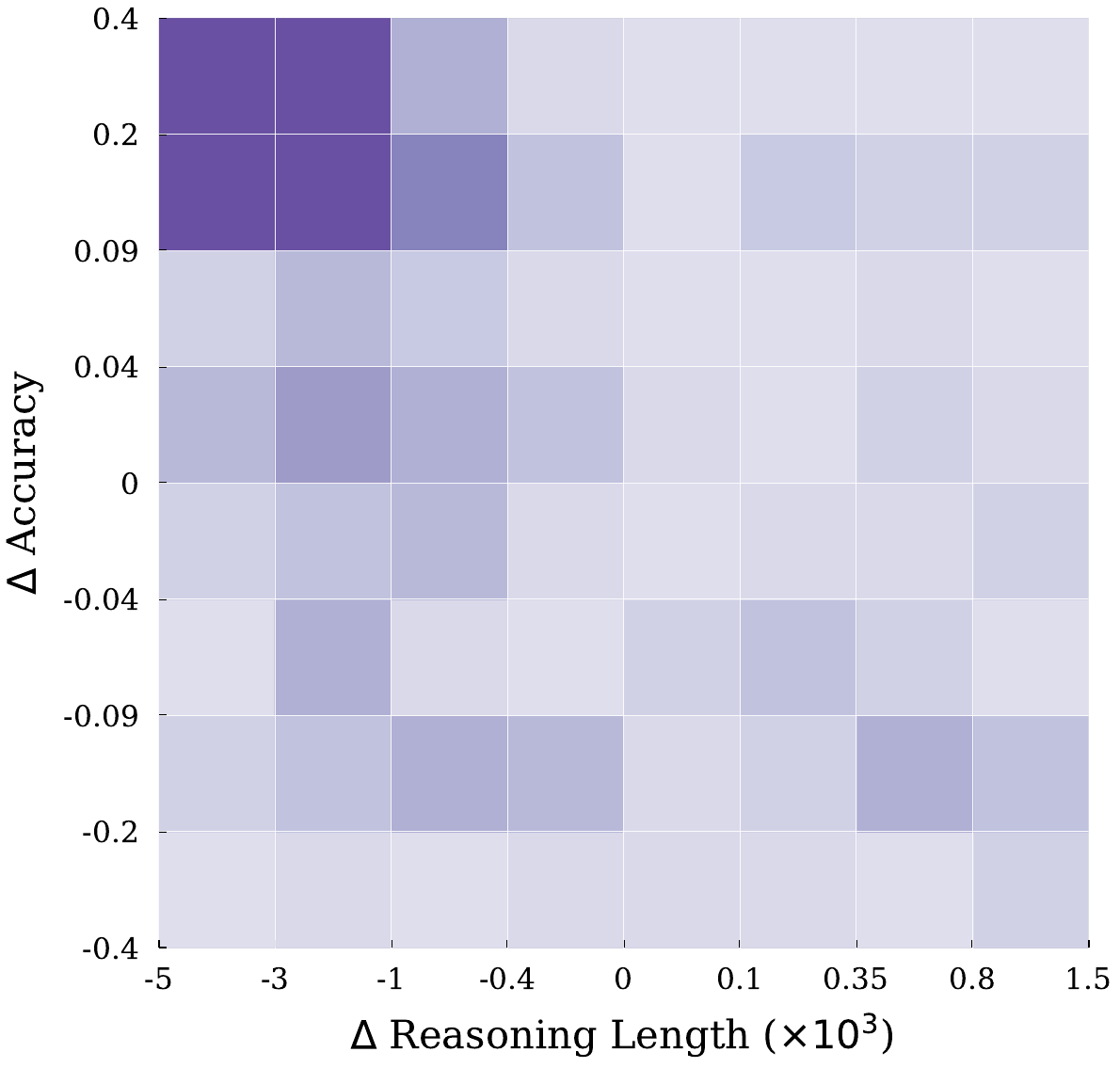}
        \caption{Difficulty Subset (0.5, 0.75]}
        \label{fig:heatmap_1.7b_short_3}
    \end{subfigure}
    \hfill
    \begin{subfigure}[t]{0.49\linewidth}
        \centering
        \includegraphics[width=\linewidth]{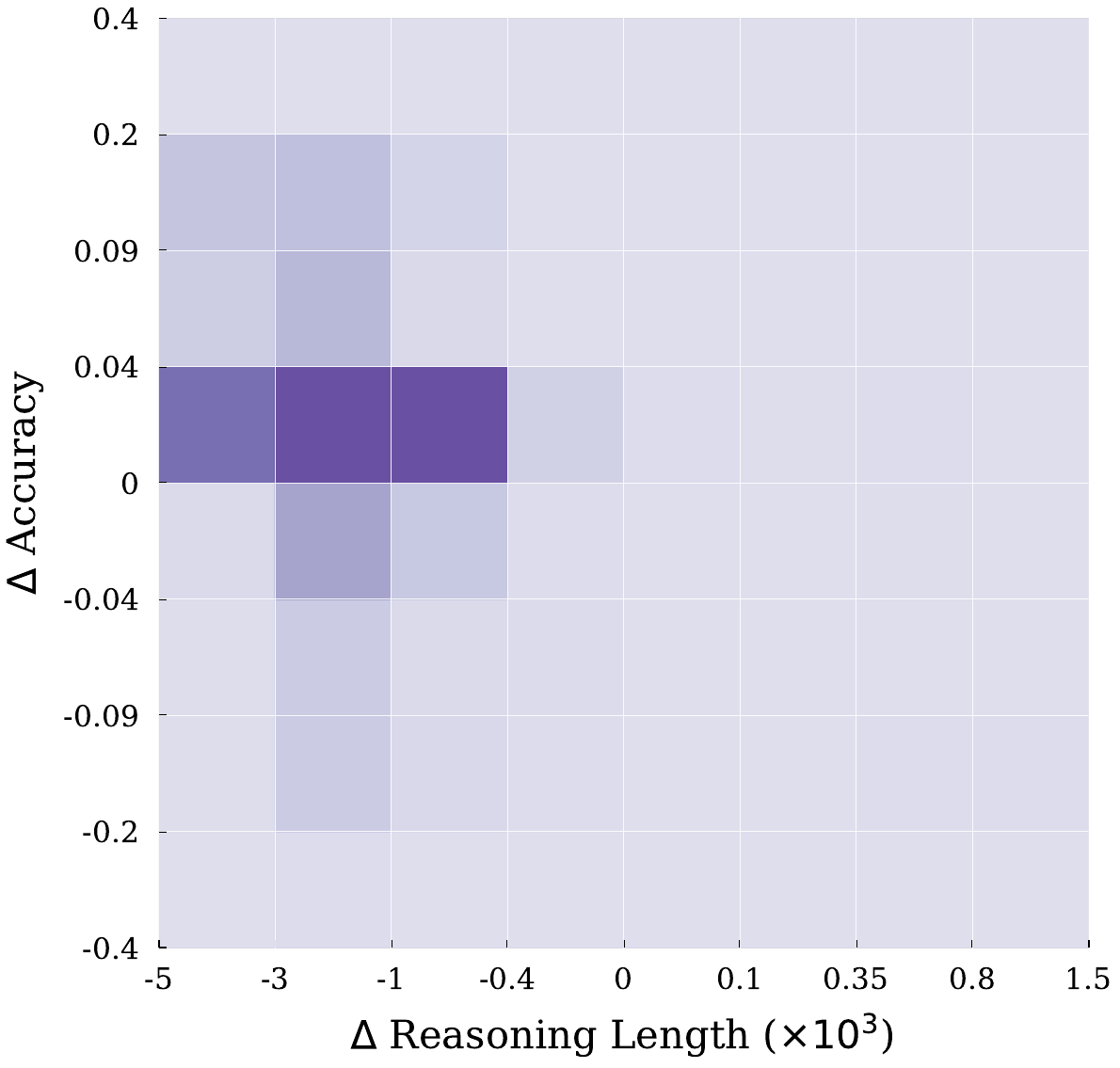}
        \caption{Difficulty Subset (0.75, 1.0]}
        \label{fig:heatmap_1.7b_short_4}
    \end{subfigure}
    \caption{Density heatmaps of \(\Delta\)length vs.\ \(\Delta\)accuracy for \textbf{Qwen3-1.7B} under Short-Reward.}
    \label{fig:heatmap_1.7b_short}
\end{figure}

\paragraph{Short-Reward.}
As shown in Figure~\ref{fig:heatmap_1.7b_short}, under short-reward training, Qwen3-1.7B exhibits a strong leftward shift across all difficulty subsets, confirming that the short reward successfully compresses reasoning length. Notably, the \((0.25, 0.5]\) shows a concentration of density in the upper-left region, indicating that removing redundant reasoning steps improves accuracy for partially solvable questions. The hardest subset \([0, 0.25]\) remains clustered near zero \(\Delta\) Accuracy, 
while the easiest subset \((0.75, 1.0]\) exhibits only a slight accuracy decline under length reduction.

\begin{figure}[H]
    \centering
    \begin{subfigure}[t]{0.49\linewidth}
        \centering
        \includegraphics[width=\linewidth]{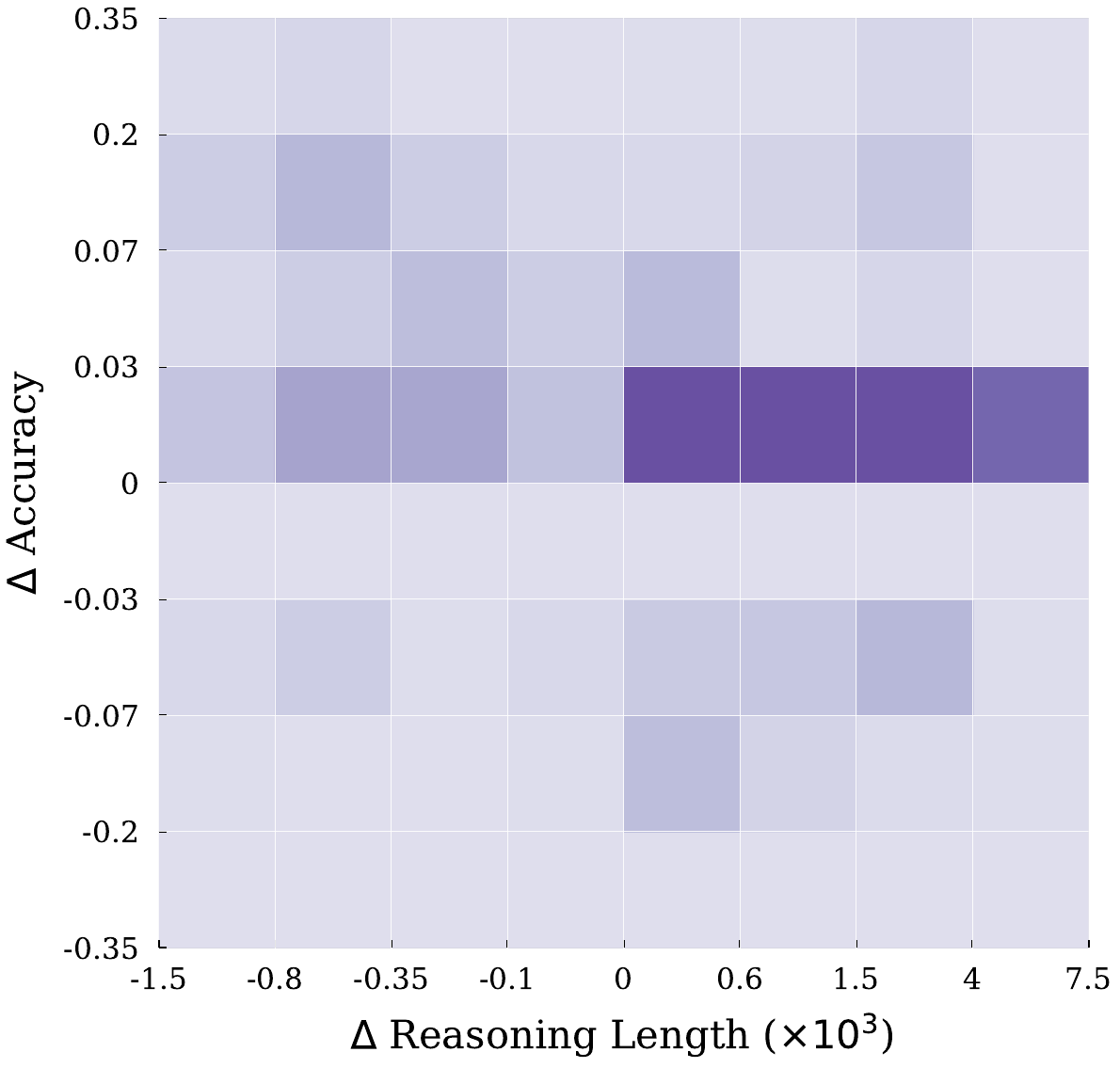}
        \caption{Difficulty Subset [0, 0.25]}
        \label{fig:heatmap_1.7b_long_1}
    \end{subfigure}
    \hfill
    \begin{subfigure}[t]{0.49\linewidth}
        \centering
        \includegraphics[width=\linewidth]{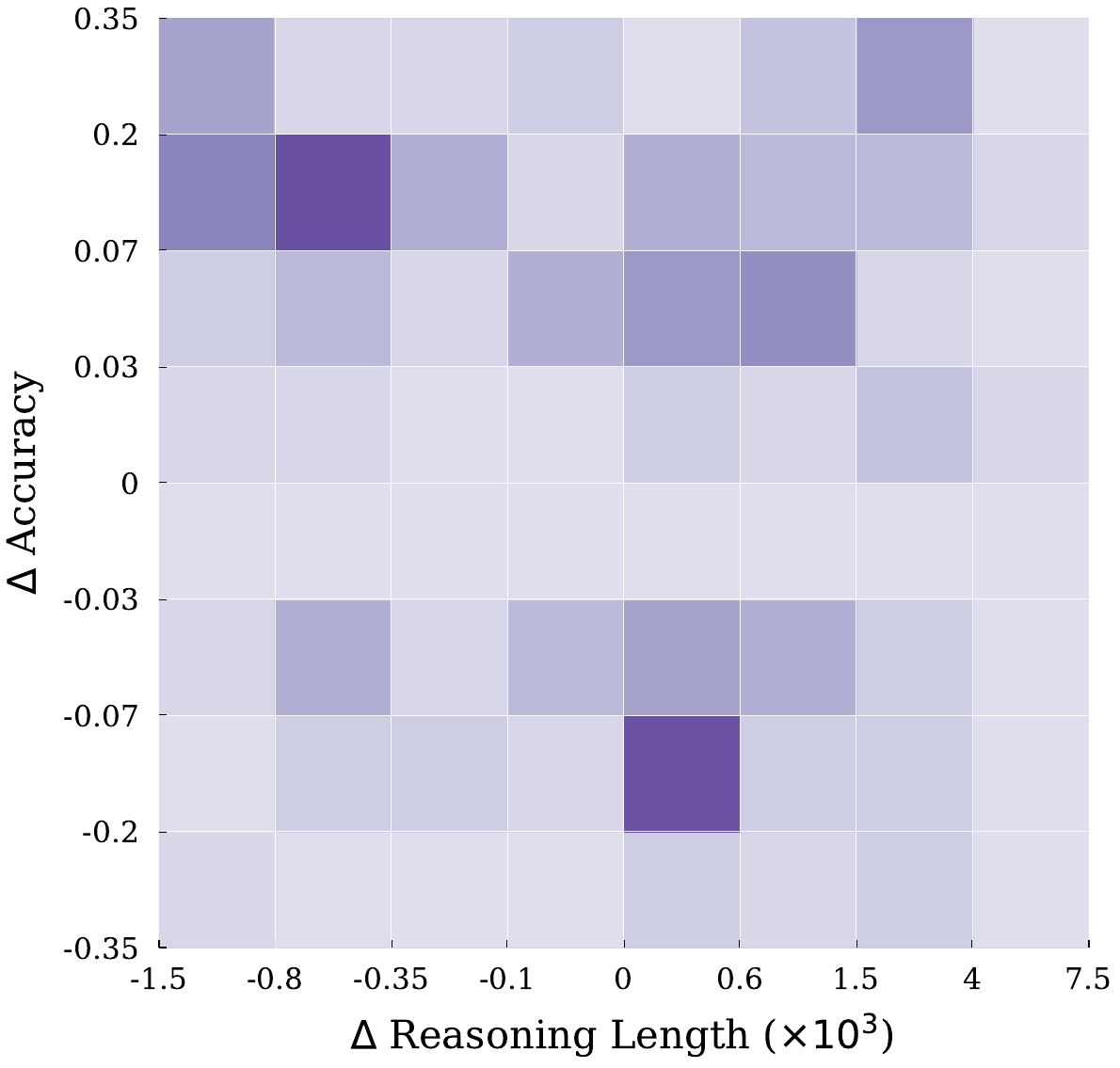}
        \caption{Difficulty Subset (0.25, 0.5]}
        \label{fig:heatmap_1.7b_long_2}
    \end{subfigure}
 
    \vspace{1em}
 
    \begin{subfigure}[t]{0.49\linewidth}
        \centering
        \includegraphics[width=\linewidth]{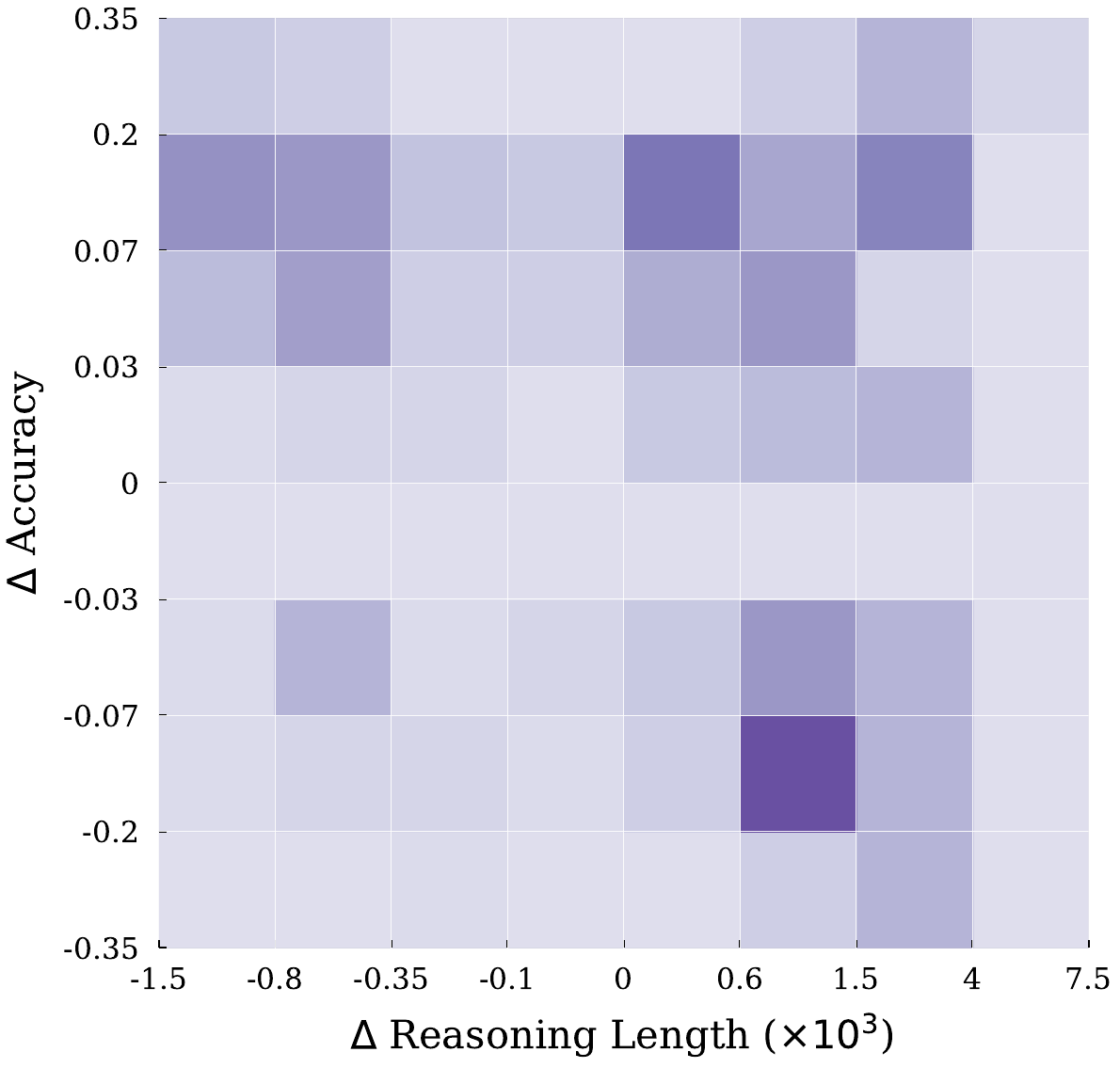}
        \caption{Difficulty Subset (0.5, 0.75]}
        \label{fig:heatmap_1.7b_long_3}
    \end{subfigure}
    \hfill
    \begin{subfigure}[t]{0.49\linewidth}
        \centering
        \includegraphics[width=\linewidth]{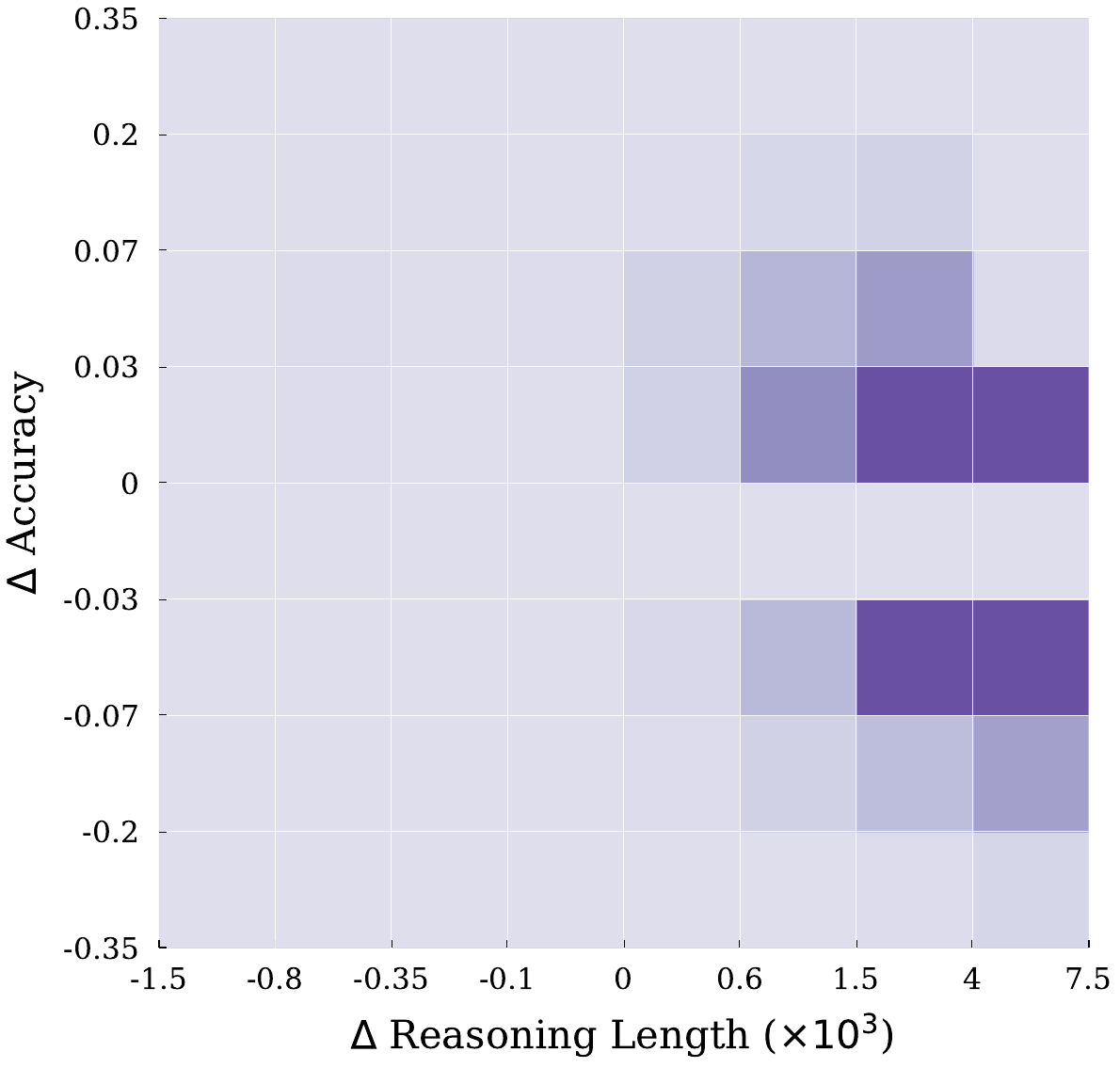}
        \caption{Difficulty Subset (0.75, 1.0]}
        \label{fig:heatmap_1.7b_long_4}
    \end{subfigure}
    \caption{Density heatmaps of \(\Delta\)length vs.\ \(\Delta\)accuracy for \textbf{Qwen3-1.7B} under Long-Reward.}
    \label{fig:heatmap_1.7b_long}
\end{figure}

\paragraph{Long-Reward.}
Figure~\ref{fig:heatmap_1.7b_long} presents the long-reward results for Qwen3-1.7B. Unlike the short-reward setting, all subsets exhibit a rightward shift, reflecting longer reasoning traces. However, this length increase does not consistently lead to higher accuracy. For the partially solvable subsets \((0.25, 0.5]\) and \((0.5, 0.75]\), density spreads across both positive and negative \(\Delta\)Accuracy regions despite reasoning length increases, suggesting that additional tokens contribute unreliably to improving reasoning quality. Both the hardest subset \([0, 0.25]\) and easiest subset \((0.75, 1.0]\) show rightward concentration with \(\Delta\)Accuracy fluctuating around zero, confirming that extended reasoning on questions the model either reliably solves or consistently fails to solve provides negligible benefit.

\paragraph{Key Observations for Qwen3-1.7B.}
\begin{itemize}[nosep,leftmargin=1.5em]
    \item Both the hardest subset \([0, 0.25]\) and the easiest subset \((0.75, 1.0]\) show \(\Delta\)Accuracy concentrated near zero under short-reward and long-reward training, indicating that questions at these two subsets are largely insensitive to reasoning length changes.
    \item The partially solvable subsets \((0.25, 0.5]\) and \((0.5, 0.75]\) are sensitive to reasoning length changes. Short-reward training produces relatively more concentrated accuracy improvement than long-reward training, which distributes \(\Delta\)Accuracy across both positive and negative regions.
\end{itemize}

\subsection{Qwen3-4B}
\begin{figure}[H]
    \centering
    \begin{subfigure}[t]{0.49\linewidth}
        \centering
        \includegraphics[width=\linewidth]{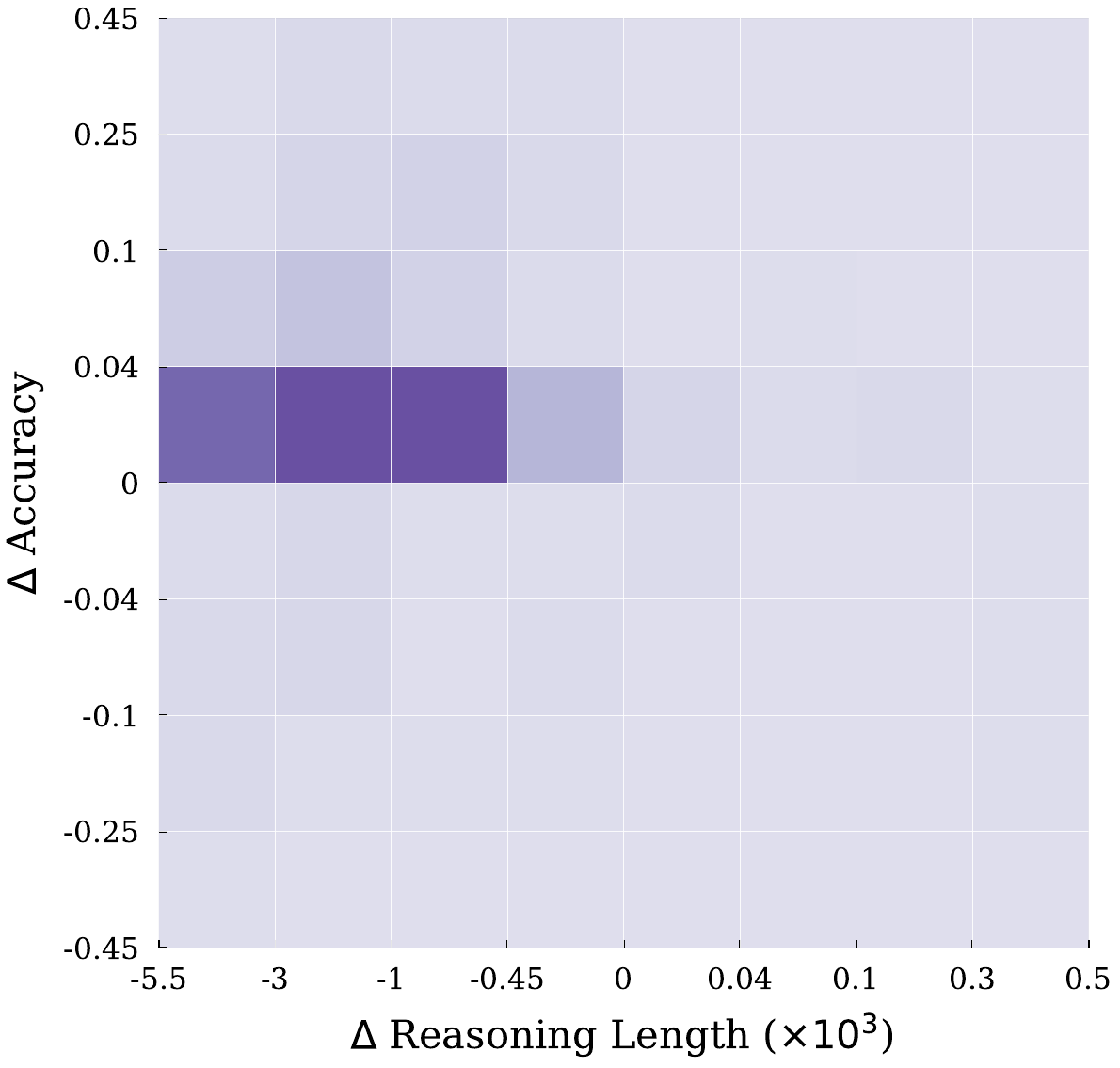}
        \caption{Difficulty Subset [0.0, 0.25]}
        \label{fig:heatmap_4b_short_1}
    \end{subfigure}
    \hfill
    \begin{subfigure}[t]{0.49\linewidth}
        \centering
        \includegraphics[width=\linewidth]{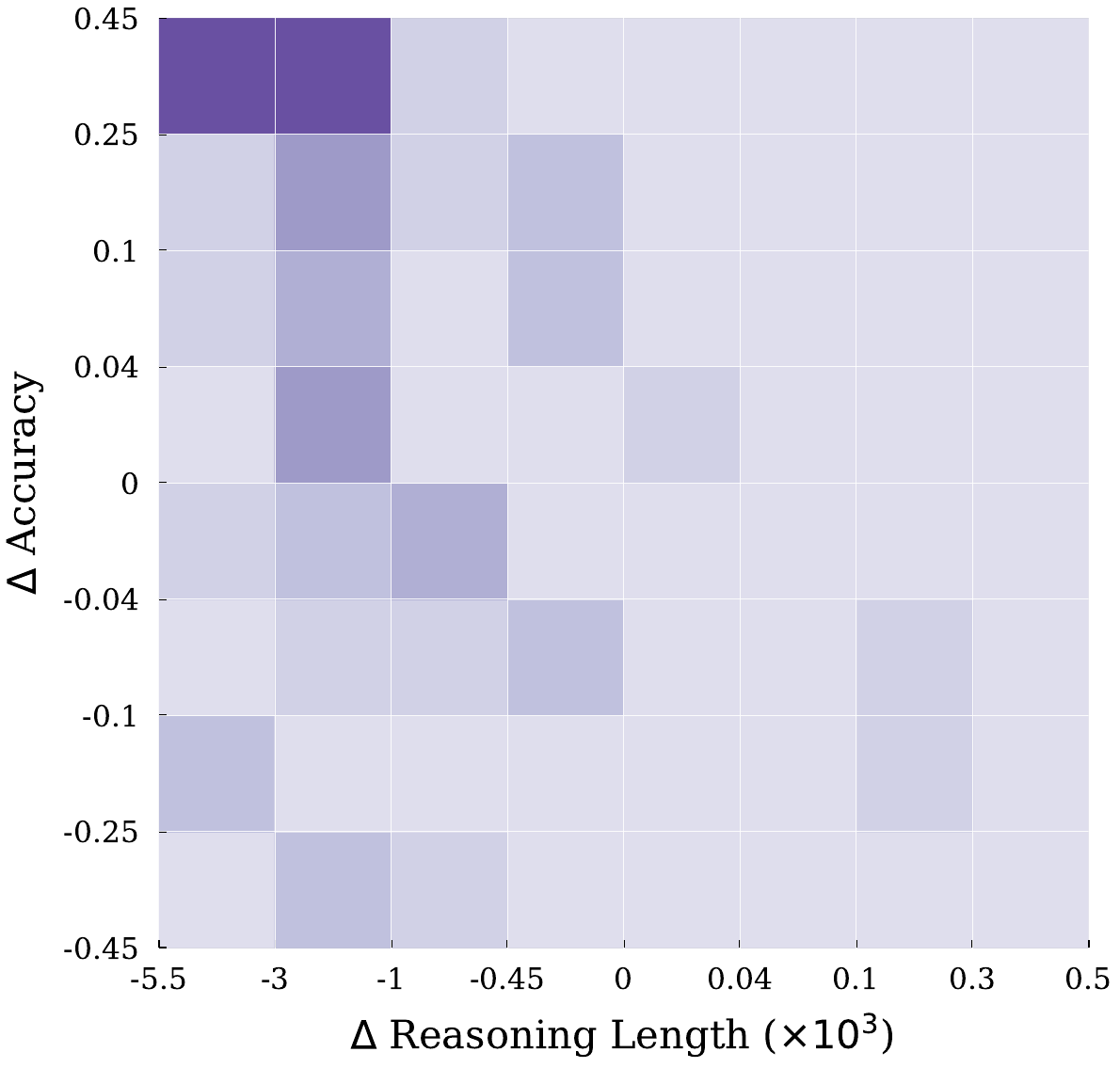}
        \caption{Difficulty Subset (0.25, 0.5]}
        \label{fig:heatmap_4b_short_2}
    \end{subfigure}
 
    \vspace{1em}
 
    \begin{subfigure}[t]{0.49\linewidth}
        \centering
        \includegraphics[width=\linewidth]{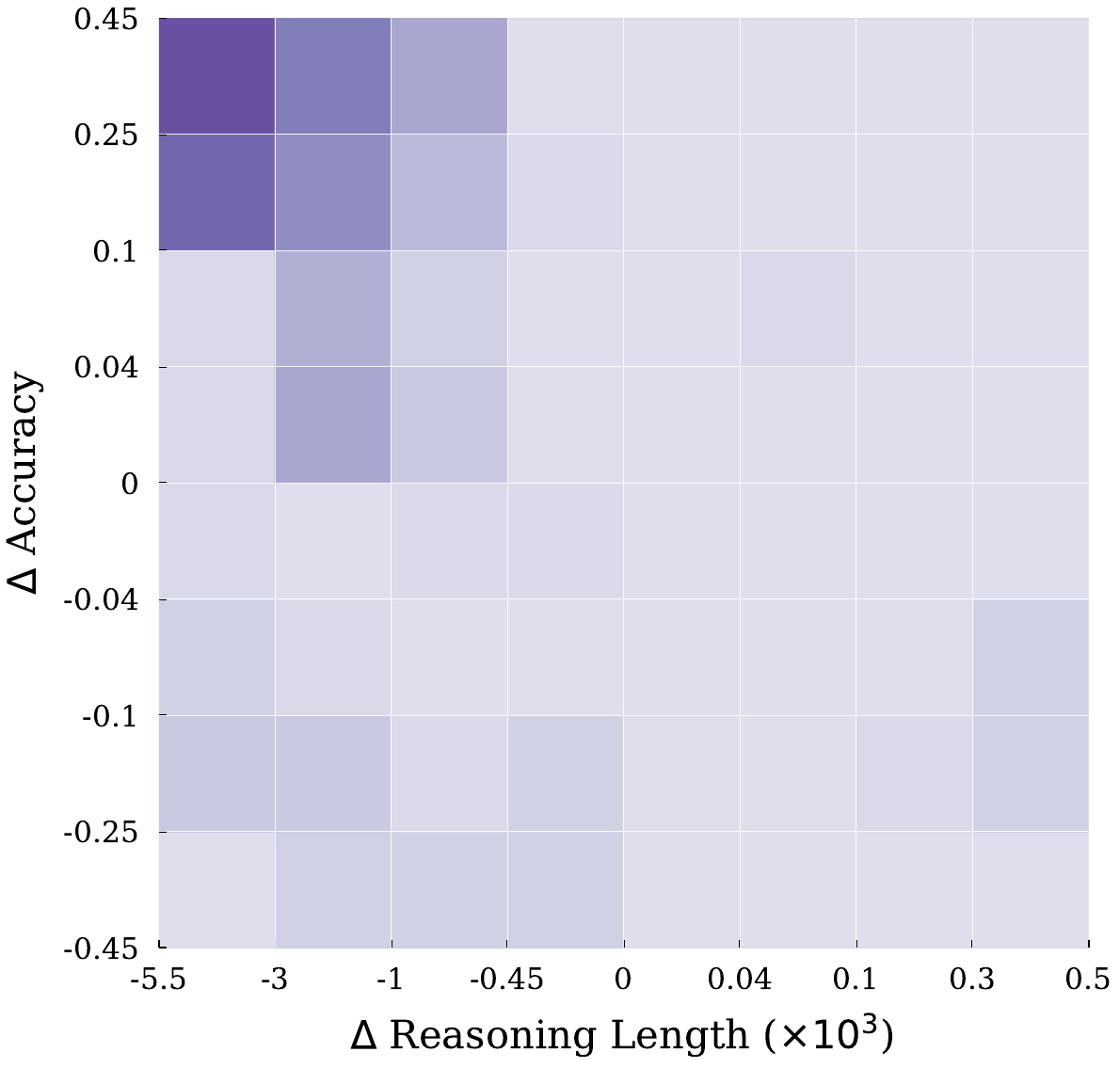}
        \caption{Difficulty Subset (0.5, 0.75]}
        \label{fig:heatmap_4b_short_3}
    \end{subfigure}
    \hfill
    \begin{subfigure}[t]{0.49\linewidth}
        \centering
        \includegraphics[width=\linewidth]{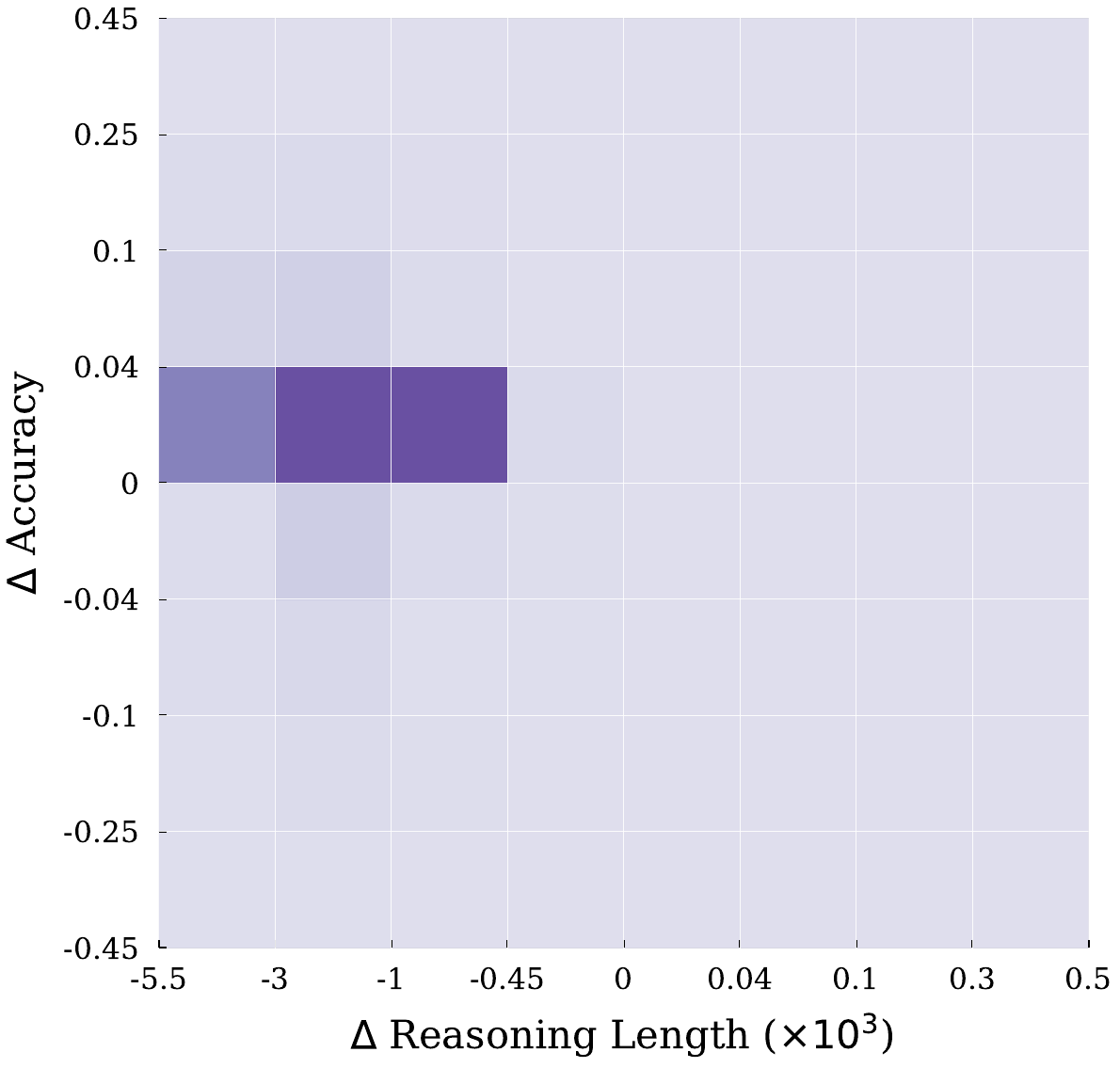}
        \caption{Difficulty Subset (0.75, 1.0]}
        \label{fig:heatmap_4b_short_4}
    \end{subfigure}
    \caption{Density heatmaps of \(\Delta\)length vs.\ \(\Delta\)accuracy for \textbf{Qwen3-4B} under Short-Reward.}
    \label{fig:heatmap_4b_short}
\end{figure}

\paragraph{Short-Reward.}
The short-reward heatmaps for Qwen3-4B are shown in Figure~\ref{fig:heatmap_4b_short}. Consistent with the 1.7B results, length compression is evident across all difficulty subsets, as indicated by the leftward-concentrated density. Most of the density is concentrated on negative \(\Delta\)Length regions, with only limited mass extending into the positive region. The most striking observation appears in the \((0.25, 0.5]\) subset, where a large fraction of questions concentrate in the upper-left region, with the strongest positive \(\Delta\)Accuracy occurring at relatively large negative \(\Delta\)Length regions, indicating that substantial length reductions could coincide with notable accuracy improvements for partially solvable questions. The \((0.5, 0.75]\) subset displays a similar but relatively weaker trend. These two partially solvable subsets also exhibit a broader spread along the \(\Delta\)Accuracy axis than the hardest and easiest subsets. For both the hardest \([0, 0.25]\) and easiest \((0.75, 1.0]\) subsets, \(\Delta\)Accuracy remains clustered around zero despite substantial length reductions, indicating that reasoning length has limited influence on questions at these two difficulty subsets.

\begin{figure}[H]
    \centering
    \begin{subfigure}[t]{0.49\linewidth}
        \centering
        \includegraphics[width=\linewidth]{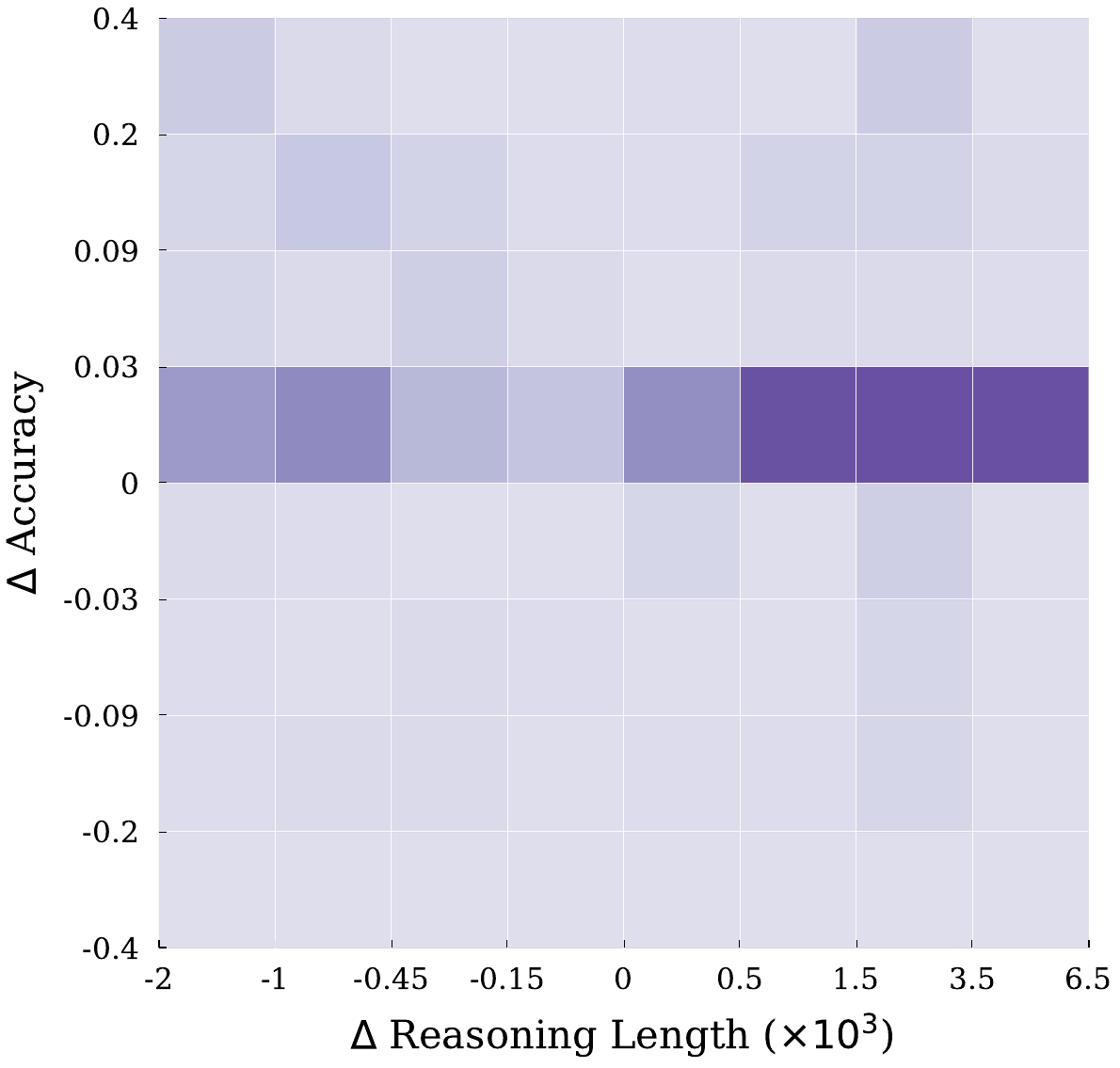}
        \caption{Difficulty Subset [0.0, 0.25]}
        \label{fig:heatmap_4b_long_1}
    \end{subfigure}
    \hfill
    \begin{subfigure}[t]{0.49\linewidth}
        \centering
        \includegraphics[width=\linewidth]{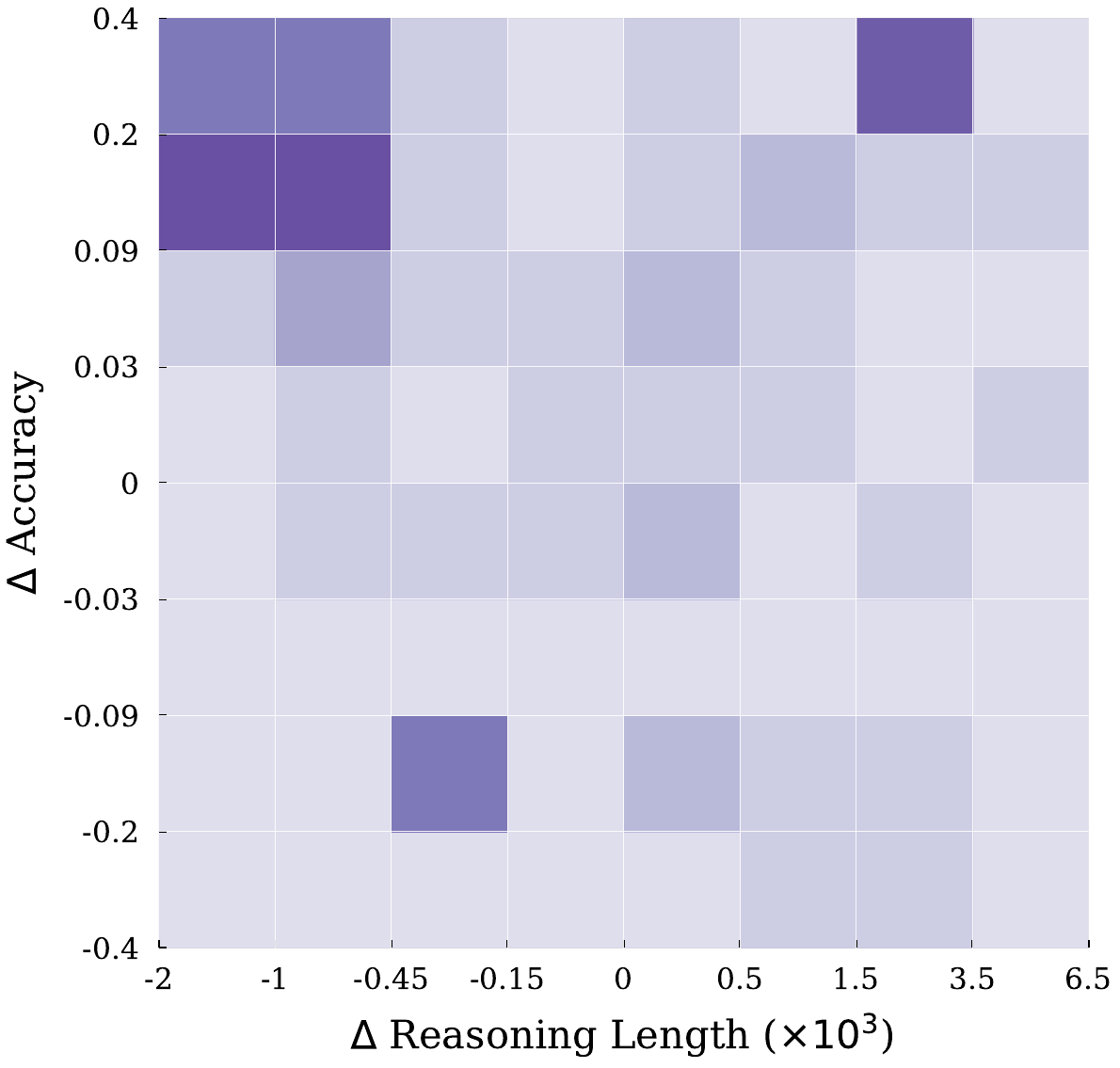}
        \caption{Difficulty Subset (0.25, 0.5]}
        \label{fig:heatmap_4b_long_2}
    \end{subfigure}

    \vspace{1em}
 
    \begin{subfigure}[t]{0.49\linewidth}
        \centering
        \includegraphics[width=\linewidth]{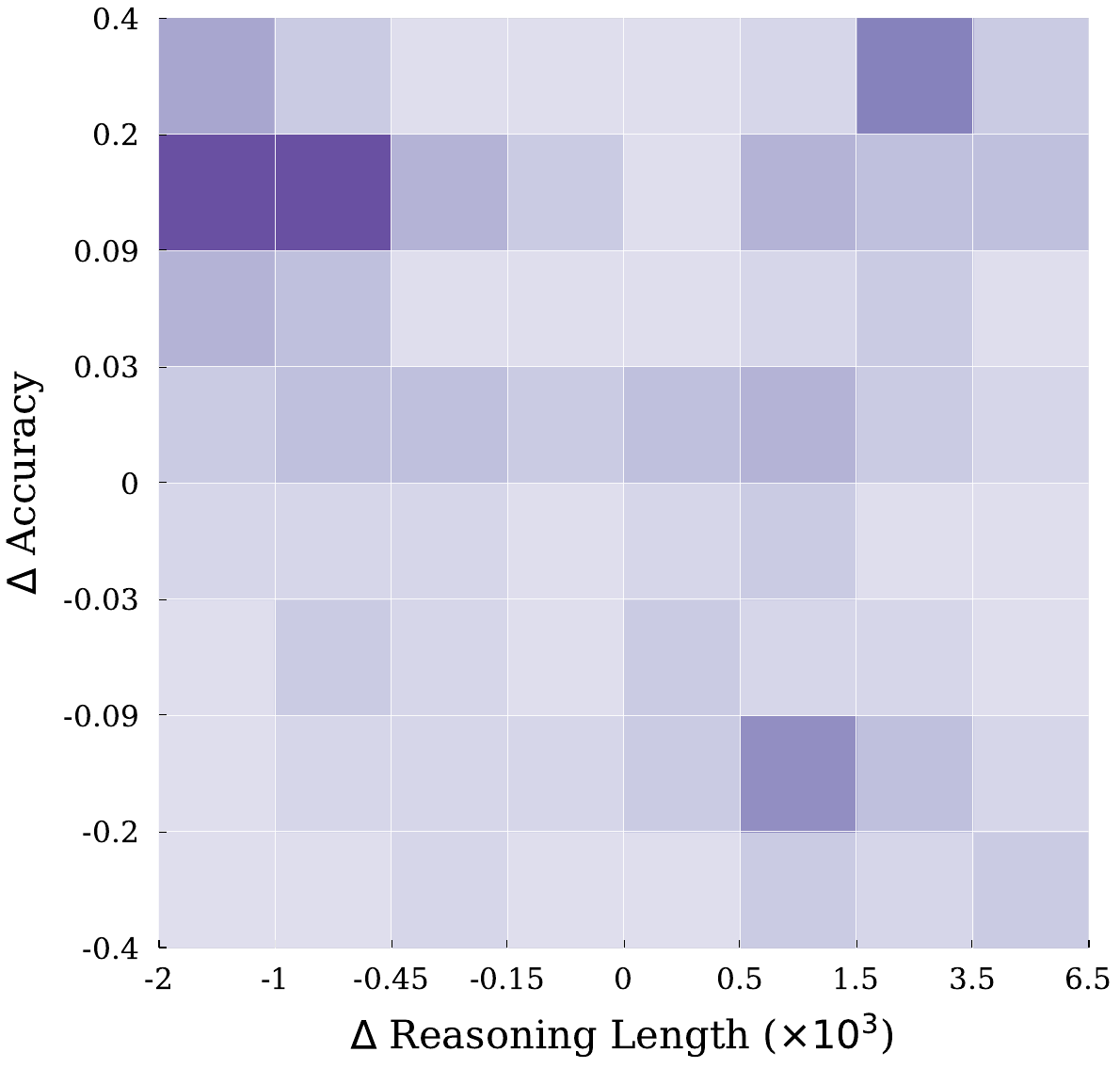}
        \caption{Difficulty Subset (0.5, 0.75]}
        \label{fig:heatmap_4b_long_3}
    \end{subfigure}
    \hfill
    \begin{subfigure}[t]{0.49\linewidth}
        \centering
        \includegraphics[width=\linewidth]{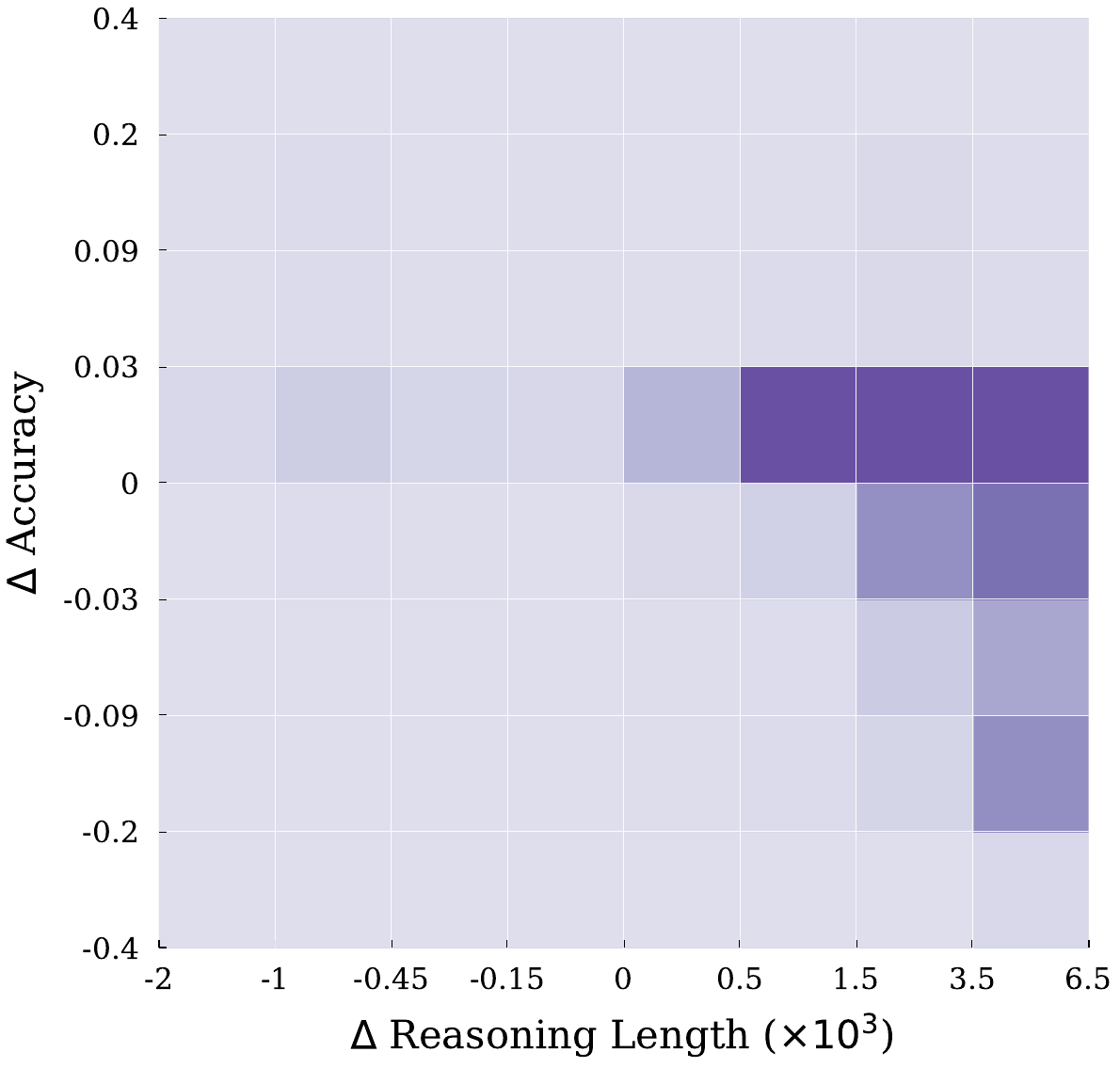}
        \caption{Difficulty Subset (0.75, 1.0]}
        \label{fig:heatmap_4b_long_4}
    \end{subfigure}
    \caption{Density heatmaps of \(\Delta\)length vs.\ \(\Delta\)accuracy for \textbf{Qwen3-4B} under Long-Reward.}
    \label{fig:heatmap_4b_long}
\end{figure}

\paragraph{Long-Reward.}
The long-reward heatmaps for Qwen3-4B are shown in Figure~\ref{fig:heatmap_4b_long}. All four subsets exhibit a rightward shift extending up to \(6{,}500\) tokens. The density also spans a substantially broader range of positive \(\Delta\)Length regions across the four subsets. This rightward spread is visible across the entire difficulty subset. In particular, the \((0.25, 0.5]\) subset exhibits no consistent accuracy trend under length extension. One cluster achieves accuracy improvement in the upper-right region at large length increases, while another shows accuracy decline, indicating that the long reward indiscriminately extends reasoning length yet produces divergent accuracy effects across different questions. Moreover, the \((0.5, 0.75]\) subset displays a similar trend to that of the \((0.25, 0.5]\) subset, with density distributed across multiple \(\Delta\)Accuracy regions. For both the hardest \([0, 0.25]\) and easiest \((0.75, 1.0]\) subsets, \(\Delta\)Accuracy remains clustered near zero despite reasoning length increases, though the easiest subset \((0.75, 1.0]\) is more prone to accuracy decline as reasoning length grows.

\clearpage

\paragraph{Key Observations for Qwen3-4B.}
\begin{itemize}[nosep,leftmargin=1.5em]
    \item Compared to 1.7B, the 4B model exhibits a narrower positive \(\Delta\)Length range under short-reward, indicating more length compression with almost no questions showing length increases. Despite a wider \(\Delta\)Accuracy improvement, the hardest \([0, 0.25]\) and easiest subsets \((0.75, 1.0]\) still concentrate near zero, further suggesting that these questions are insensitive to length changes.
    \item For the partially solvable subsets \((0.25, 0.5]\) and \((0.5, 0.75]\), the narrower positive length range leads to a sharper upper-left concentration under short-reward, indicating that Qwen3-4B achieves effective length compression on a broader range of questions within these subsets. Under long-reward, accuracy improvements remain inconsistent despite length extension.
\end{itemize}
\subsection{Qwen3-8B}
\begin{figure}[H]
    \centering
    \begin{subfigure}[t]{0.49\linewidth}
        \centering
        \includegraphics[width=\linewidth]{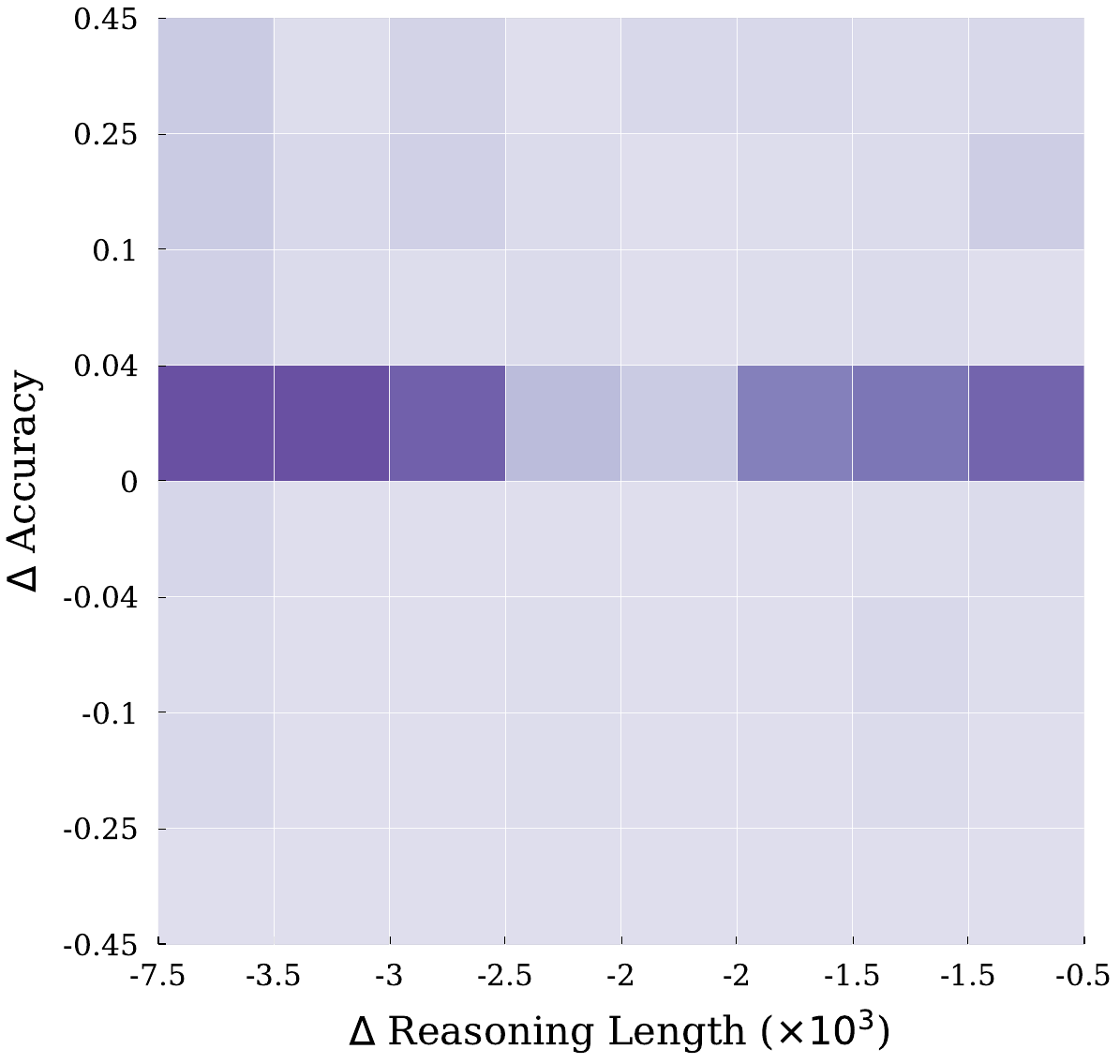}
        \caption{Difficulty Subset [0.0, 0.25]}
        \label{fig:heatmap_8b_short_1}
    \end{subfigure}
    \hfill
    \begin{subfigure}[t]{0.49\linewidth}
        \centering
        \includegraphics[width=\linewidth]{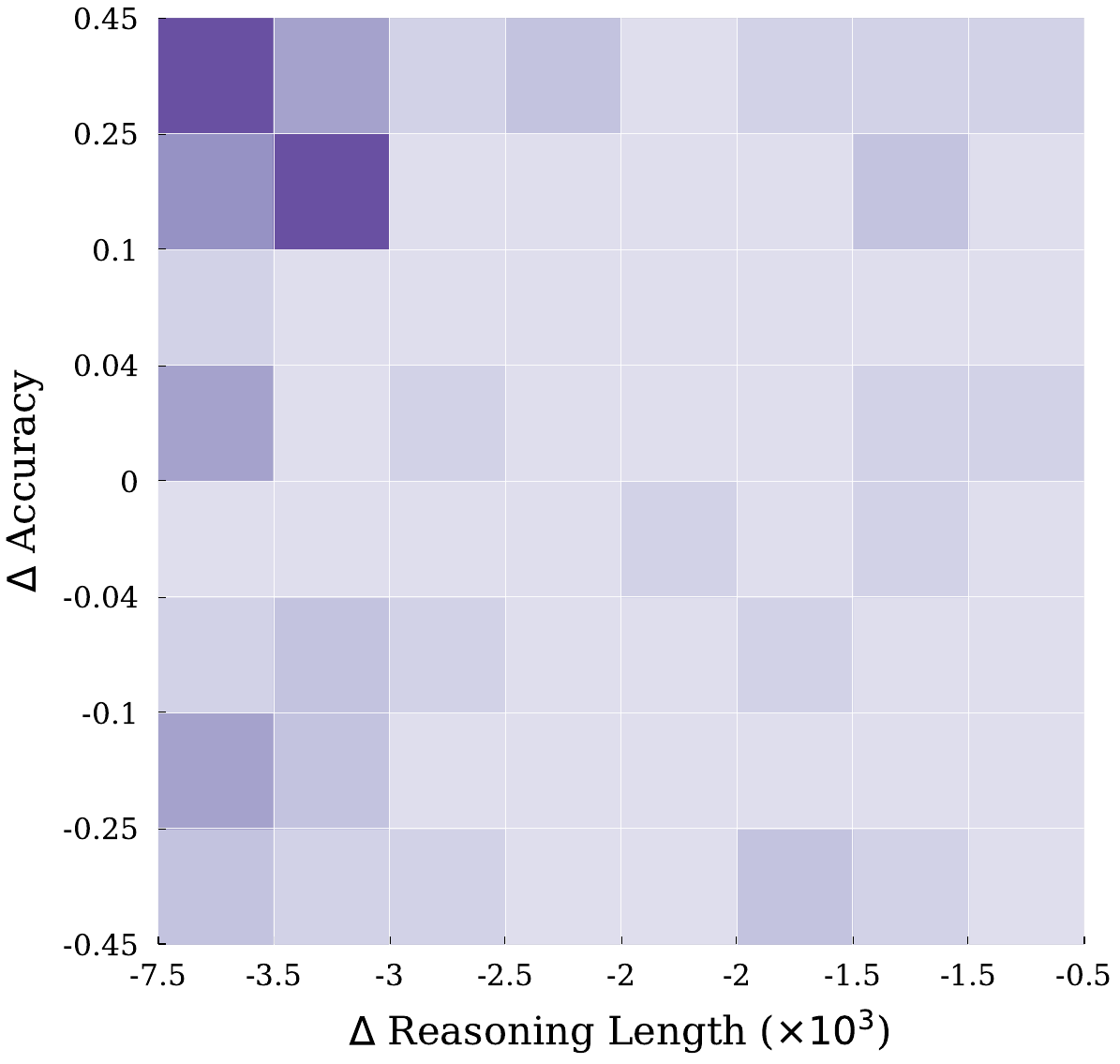}
        \caption{Difficulty Subset (0.25, 0.5]}
        \label{fig:heatmap_8b_short_2}
    \end{subfigure}

    \vspace{1em}
 
    \begin{subfigure}[t]{0.49\linewidth}
        \centering
        \includegraphics[width=\linewidth]{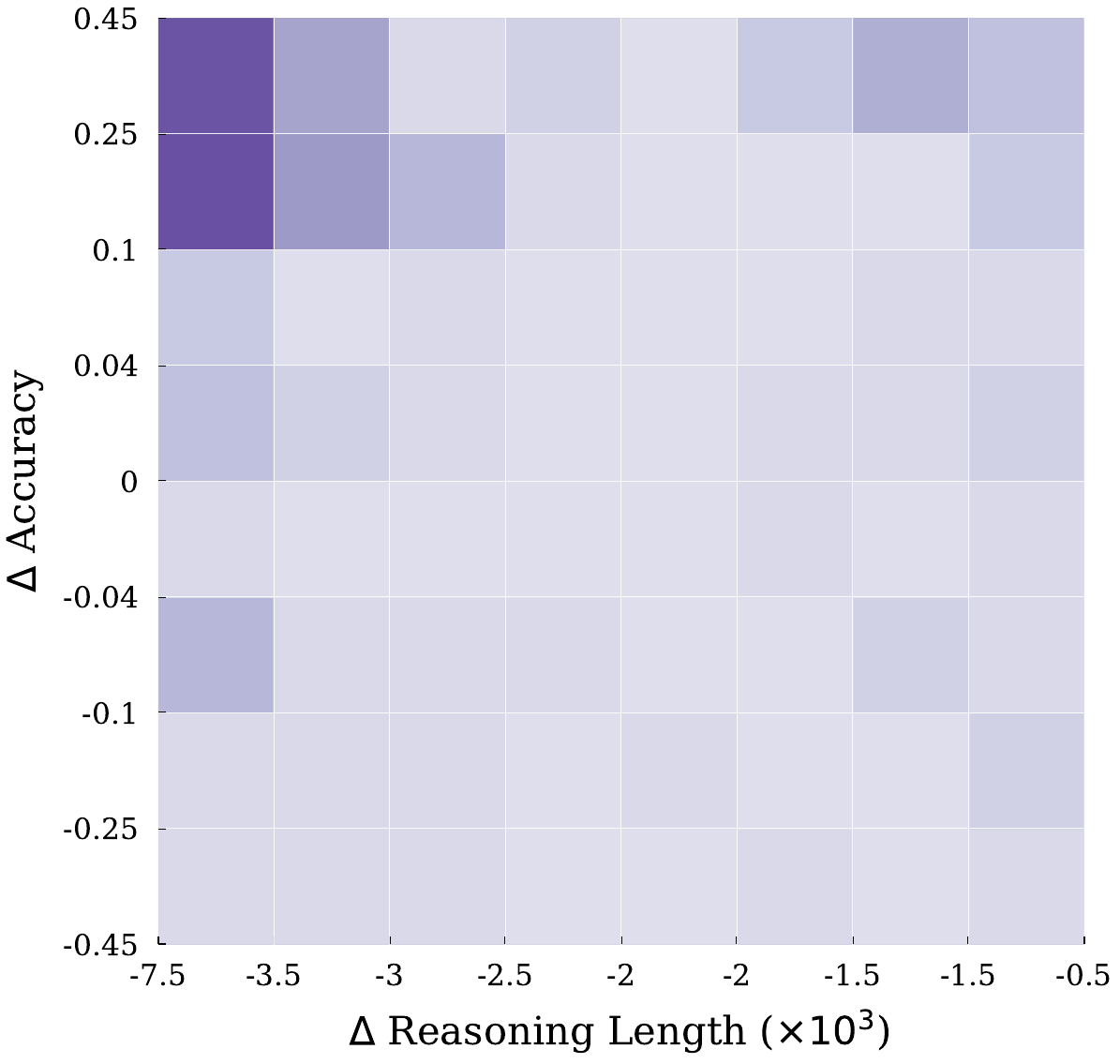}
        \caption{Difficulty Subset (0.5, 0.75]}
        \label{fig:heatmap_8b_short_3}
    \end{subfigure}
    \hfill
    \begin{subfigure}[t]{0.49\linewidth}
        \centering
        \includegraphics[width=\linewidth]{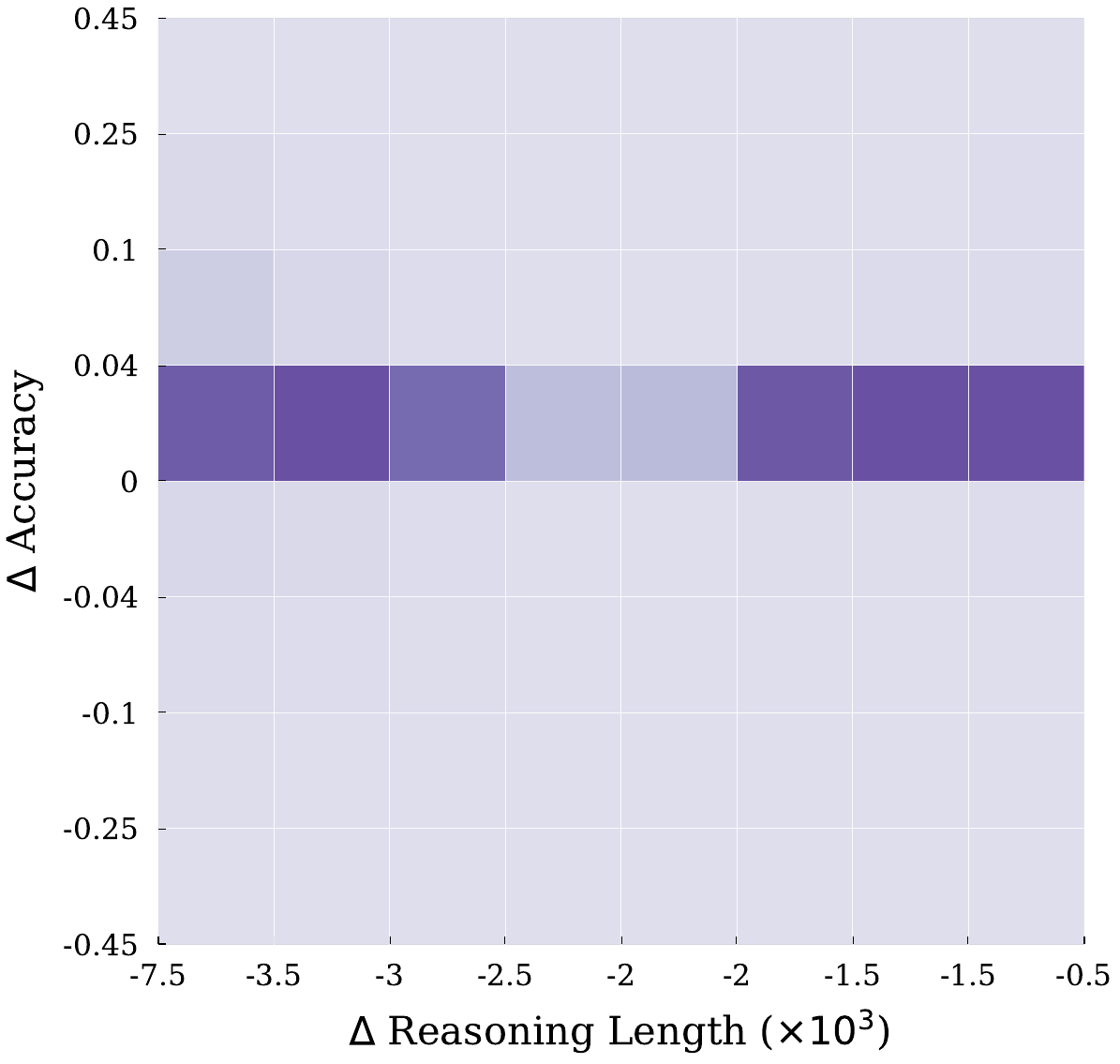}
        \caption{Difficulty Subset (0.75, 1.0]}
        \label{fig:heatmap_8b_short_4}
    \end{subfigure}
    \caption{Density heatmaps of \(\Delta\)length vs.\ \(\Delta\)accuracy for \textbf{Qwen3-8B} under Short-Reward.}
    \label{fig:heatmap_8b_short}
\end{figure}

\paragraph{Short-Reward.}
Figure~\ref{fig:heatmap_8b_short} illustrates the short-reward results for Qwen3-8B. Under short-reward training, density remains entirely in the negative \(\Delta\)Length region in four difficulty subsets, showing strong compression relative to the base model. Notably, these large reductions in reasoning length do not always lead to accuracy degradation. Accuracy changes are concentrated in the two partially solvable subsets. In \((0.25, 0.5]\), density spans both positive and negative \(\Delta\)Accuracy, but several high-density regions lie above zero under large length reductions. The \((0.5, 0.75]\) subset shows a similar but clearer tendency, with positive accuracy changes concentrated toward the strongest compression, indicating that shortening reasoning can still coincide with improved performance. In contrast, the hardest and easiest subsets remain centered near zero \(\Delta\)Accuracy throughout a broad range of negative \(\Delta\)Length, suggesting limited sensitivity to such reductions.

\begin{figure}[H]
    \centering
    \begin{subfigure}[t]{0.49\linewidth}
        \centering
        \includegraphics[width=\linewidth]{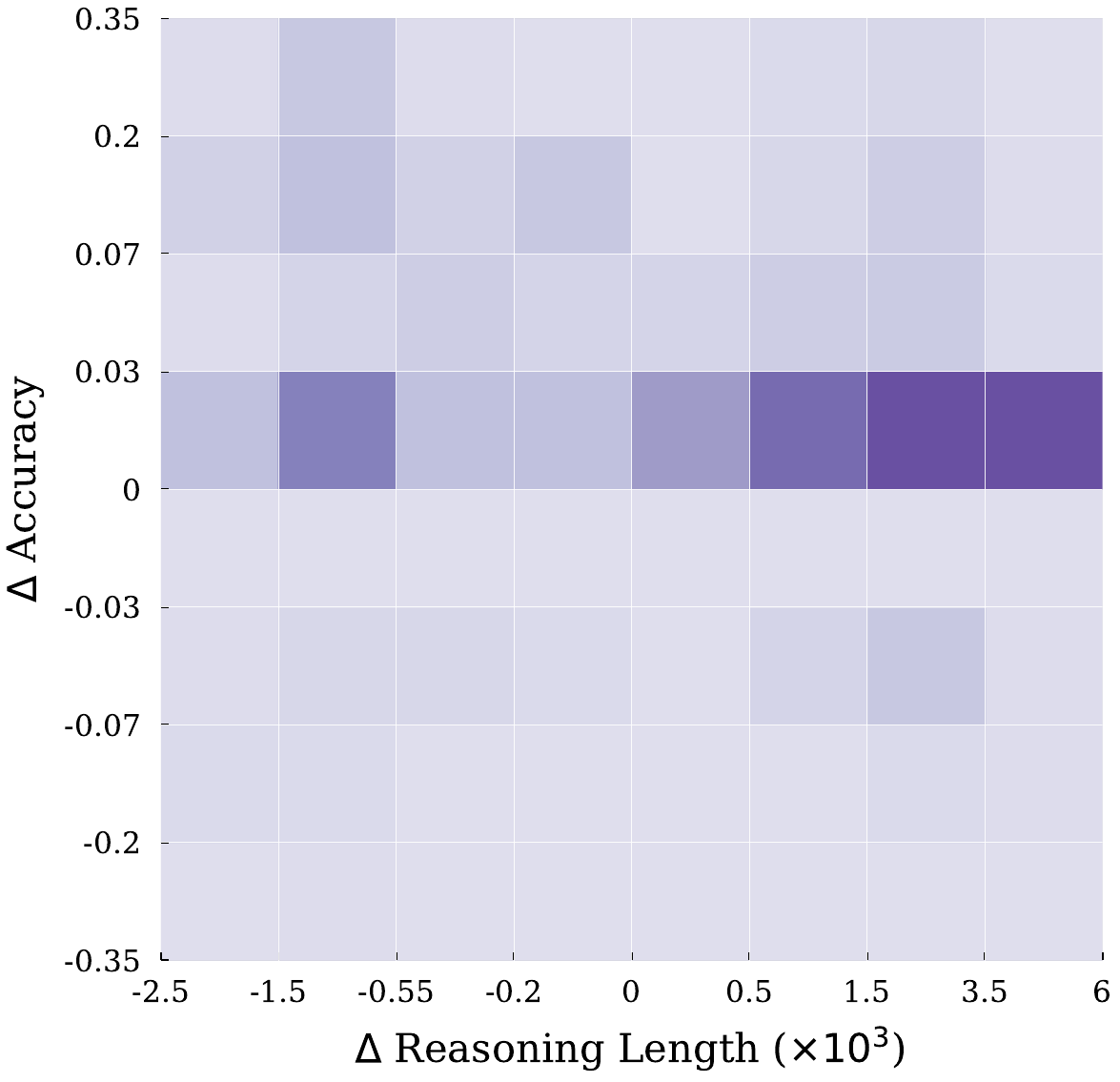}
        \caption{Difficulty Subset [0.0, 0.25]}
        \label{fig:heatmap_8b_long_1}
    \end{subfigure}
    \hfill
    \begin{subfigure}[t]{0.49\linewidth}
        \centering
        \includegraphics[width=\linewidth]{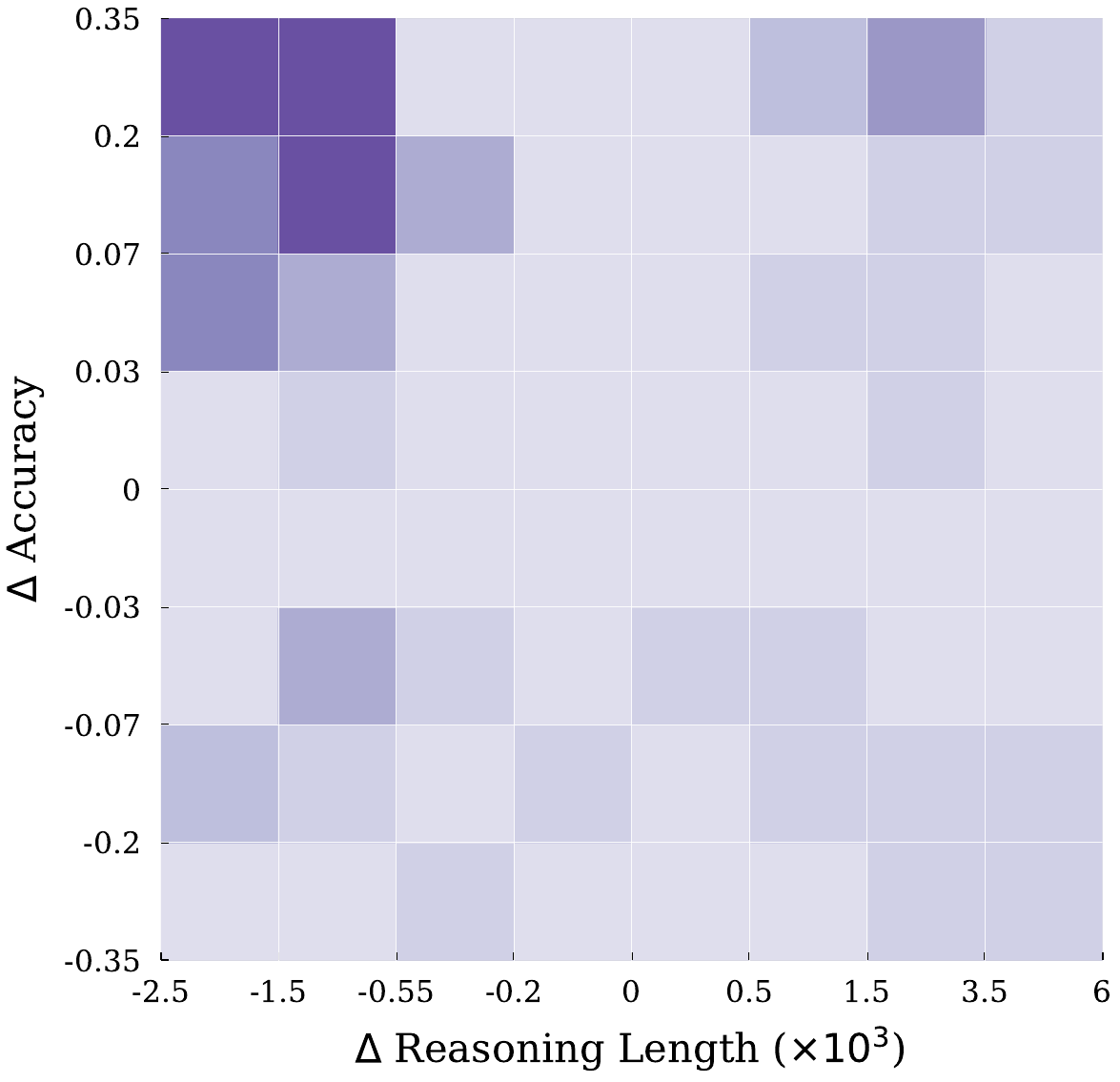}
        \caption{Difficulty Subset (0.25, 0.5]}
        \label{fig:heatmap_8b_long_2}
    \end{subfigure}

    \vspace{1em}
 
    \begin{subfigure}[t]{0.49\linewidth}
        \centering
        \includegraphics[width=\linewidth]{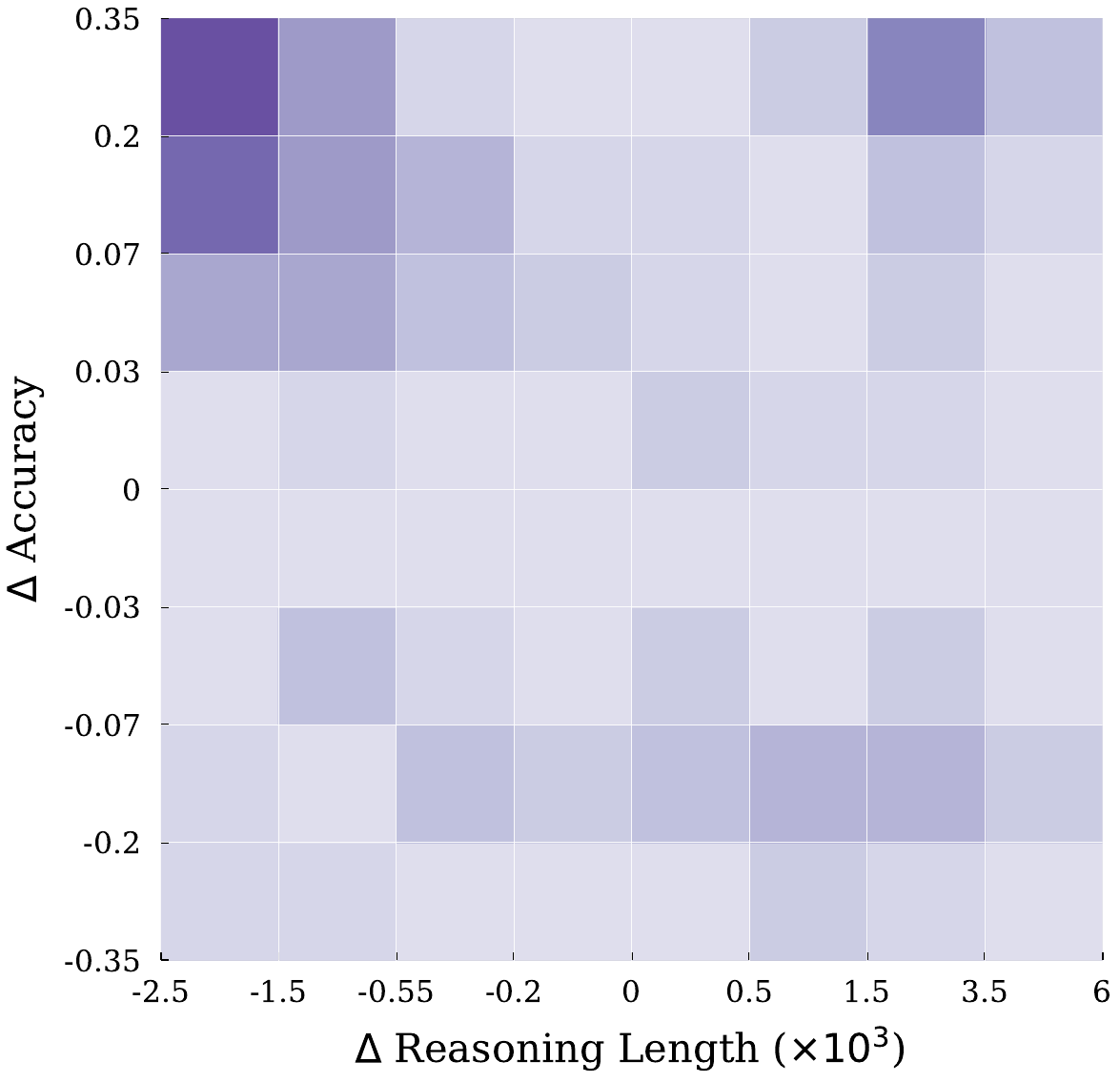}
        \caption{Difficulty Subset (0.5, 0.75]}
        \label{fig:heatmap_8b_long_3}
    \end{subfigure}
    \hfill
    \begin{subfigure}[t]{0.49\linewidth}
        \centering
        \includegraphics[width=\linewidth]{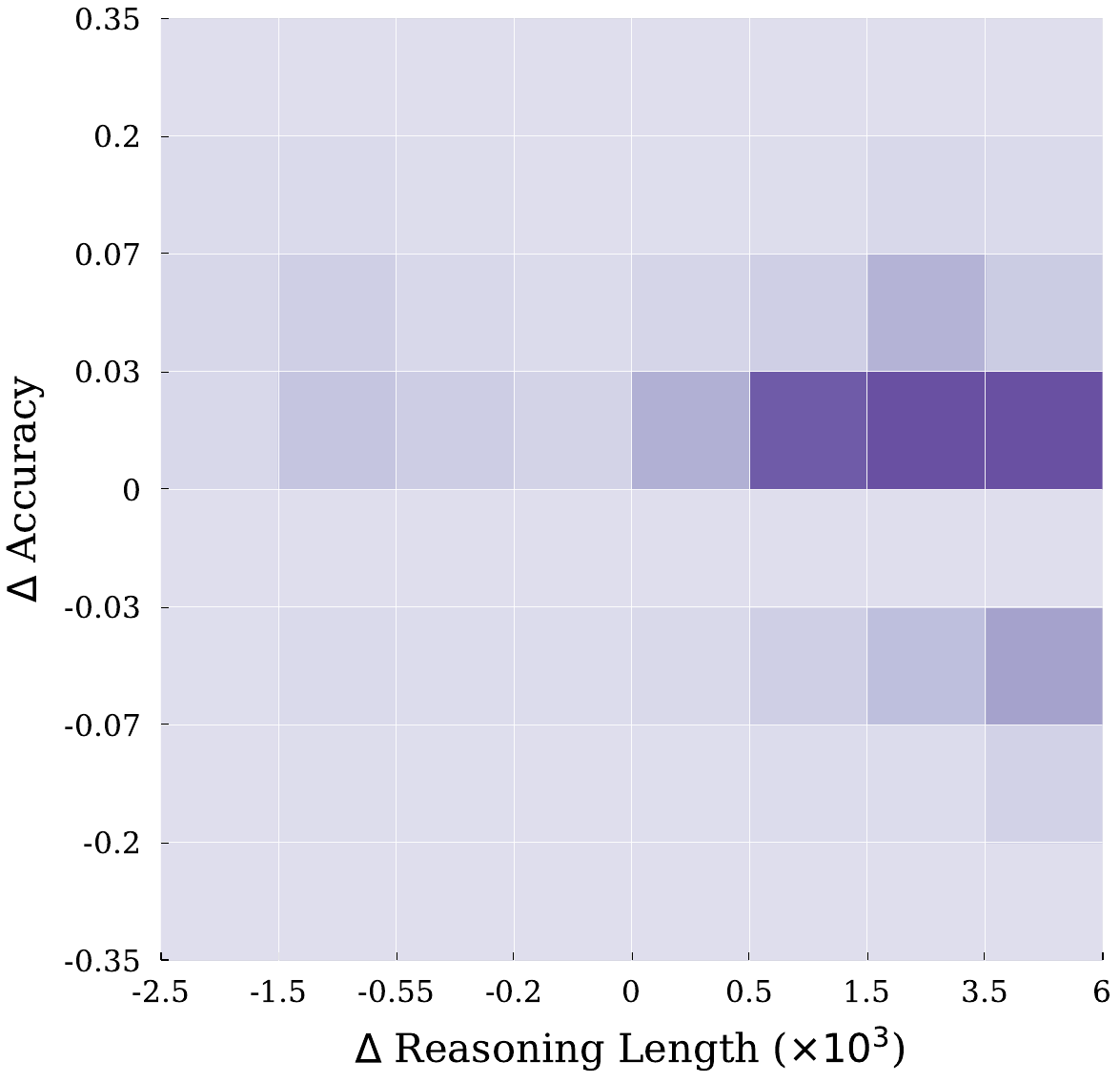}
        \caption{Difficulty Subset (0.75, 1.0]}
        \label{fig:heatmap_8b_long_4}
    \end{subfigure}
    \caption{Density heatmaps of \(\Delta\)length vs.\ \(\Delta\)accuracy for \textbf{Qwen3-8B} under Long-Reward.}
    \label{fig:heatmap_8b_long}
\end{figure}

\paragraph{Long-Reward.}
Figure~\ref{fig:heatmap_8b_long} presents the long-reward training results for Qwen3-8B. Similar to the smaller models, reasoning length varies substantially across all four difficulty subsets, while the corresponding accuracy changes remain highly inconsistent, particularly within the partially solvable subsets. In the \((0.25, 0.5]\) subset, notable accuracy improvements appear under both substantial length reductions and large length increases, indicating that longer or shorter reasoning does not consistently lead to better performance. The \((0.5, 0.75]\) subset exhibits a similar pattern, with positive and negative \(\Delta\)Accuracy distributed across a broad range of positive and negative \(\Delta\)Length regions. For the hardest \([0, 0.25]\) subset, \(\Delta\)Accuracy remains concentrated near zero despite considerable variation in reasoning length. The easiest \((0.75, 1.0]\) subset shows a stronger shift toward increased reasoning length, yet most questions still remain close to zero \(\Delta\)Accuracy, with a small fraction exhibiting accuracy degradation at larger length increases.

\paragraph{Key Observations for Qwen3-8B.}
\begin{itemize}[nosep,leftmargin=1.5em]
    \item Under short-reward training, four difficulty subsets exhibit reductions in reasoning length, yet the hardest and easiest subsets remain tightly concentrated around zero \(\Delta\)Accuracy. This further indicates that large changes in reasoning length have limited influence on these questions.
    
    \item The partially solvable subsets respond more strongly to length changes. Short-reward produces positive \(\Delta\)Accuracy primarily under length reductions, whereas long-reward distributes both accuracy improvements and declines over a much broader range of \(\Delta\)Length.
\end{itemize}

\subsection{Summary}
Across model sizes and reward types, the heatmaps reveal a consistent pattern. The hardest questions \([0, 0.25]\) cluster near zero \(\Delta\)Accuracy regardless of length changes, confirming that these questions are largely insensitive to reasoning length. The partially solvable subsets \((0.25, 0.5]\) and \((0.5, 0.75]\) exhibit  notable accuracy shifts. Under short-reward training, length reduction can coincide with positive \(\Delta\)Accuracy, whereas under long-reward training, additional token overhead does not effectively convert into consistent accuracy improvement. The easiest questions \((0.75, 1.0]\) show comparatively small accuracy variation, as the model already solves them reliably.

\clearpage

\section{Unexpected Behaviors}
\label{app:unexpected_behaviors_results}

In this part, we report tracked training dynamics and trajectories alongside the cyclic reasoning definition and its comparison between stable and unstable intervals across different training configurations.

\subsection{Tracked Training Dynamics}
\label{app:tracked_training_dynamics}

To characterize model's behavioral dynamics under both short-reward and long-reward length incentives, we mainly track the following five metrics across training:
\begin{itemize}[itemsep=1pt, leftmargin=1.5em]
    \item \textbf{Mean Reasoning Length}: average number of reasoning tokens across all \(4{,}096\) sampled responses at each training step.
    \item \textbf{Mean Correct / Incorrect Reasoning Length}: average 
          number of reasoning tokens conditioned on whether the response is  correct or incorrect, computed over the \(4{,}096\) samples 
          at each step.
    \item \textbf{Effective Answer Rate}: the fraction of \(4{,}096\) responses that contain a parsable final answer.
    \item \textbf{Correct Rate Among Effective Answers}: accuracy 
          computed only over responses with a parsable final answer, 
          excluding responses that fail to produce a valid output.
\end{itemize}

\subsection{Training Trajectories}
\label{app:training_trajectory}

\begin{figure}[H]
    \centering
    \begin{subfigure}[t]{0.48\linewidth}
        \centering
        \includegraphics[width=\linewidth]{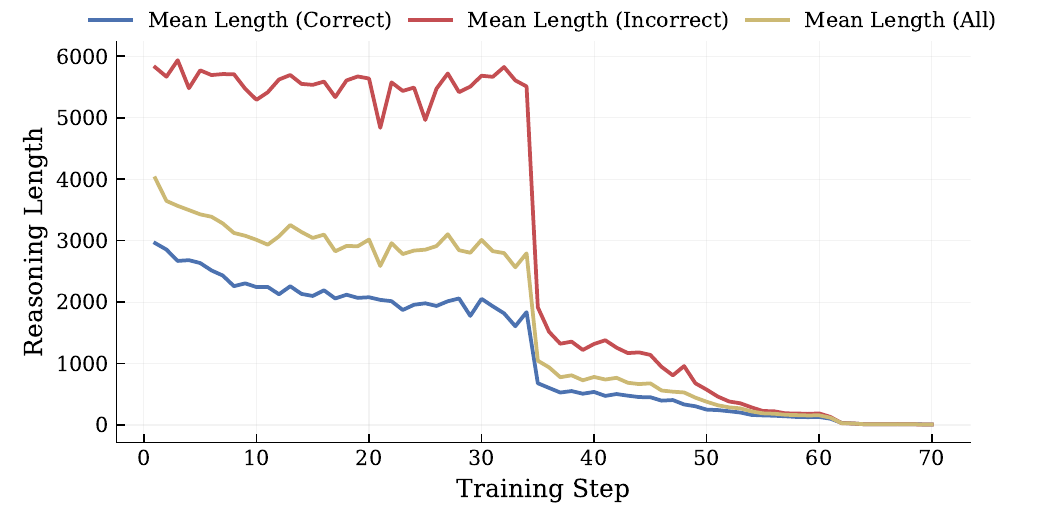}
        \caption{Reasoning length by correctness}
        \label{fig:traj_1.7b_short_1}
    \end{subfigure}
    \hfill
    \begin{subfigure}[t]{0.48\linewidth}
        \centering
        \includegraphics[width=\linewidth]{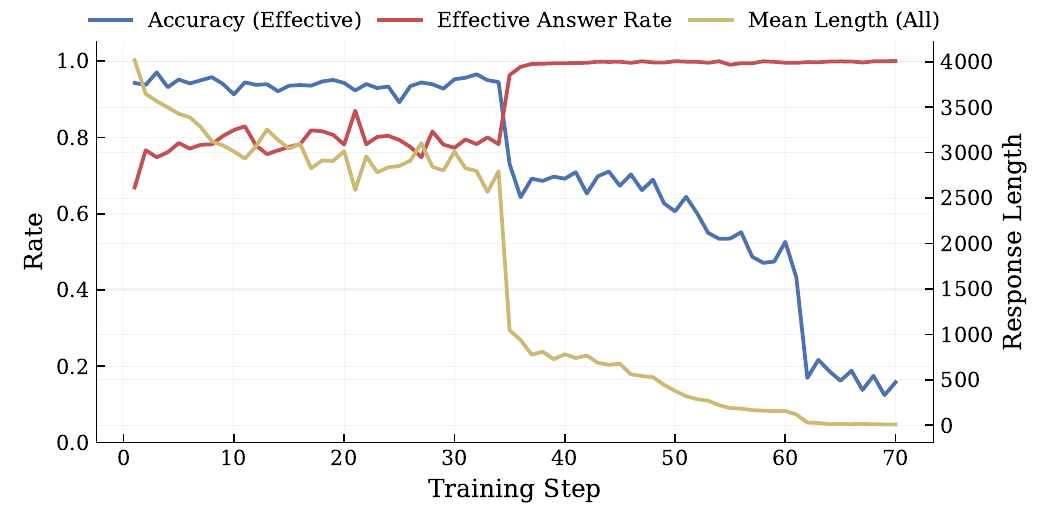}
        \caption{Effective answer rate and accuracy}
        \label{fig:traj_1.7b_short_2}
    \end{subfigure}
    \caption{Training trajectories under Short-Reward for \textbf{Qwen3-1.7B}.}
    \label{fig:traj_1.7b_short}
\end{figure}
\vspace{-2em} 
\begin{figure}[H]
    \centering
    \begin{subfigure}[t]{0.48\linewidth}
        \centering
        \includegraphics[width=\linewidth]{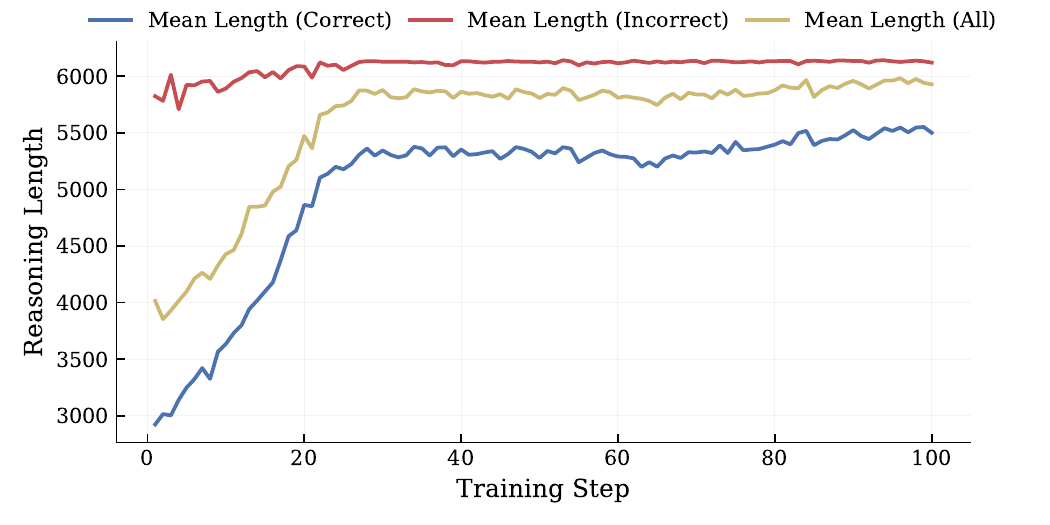}
        \caption{Reasoning length by correctness}
        \label{fig:traj_1.7b_long_1}
    \end{subfigure}
    \hfill
    \begin{subfigure}[t]{0.48\linewidth}
        \centering
        \includegraphics[width=\linewidth]{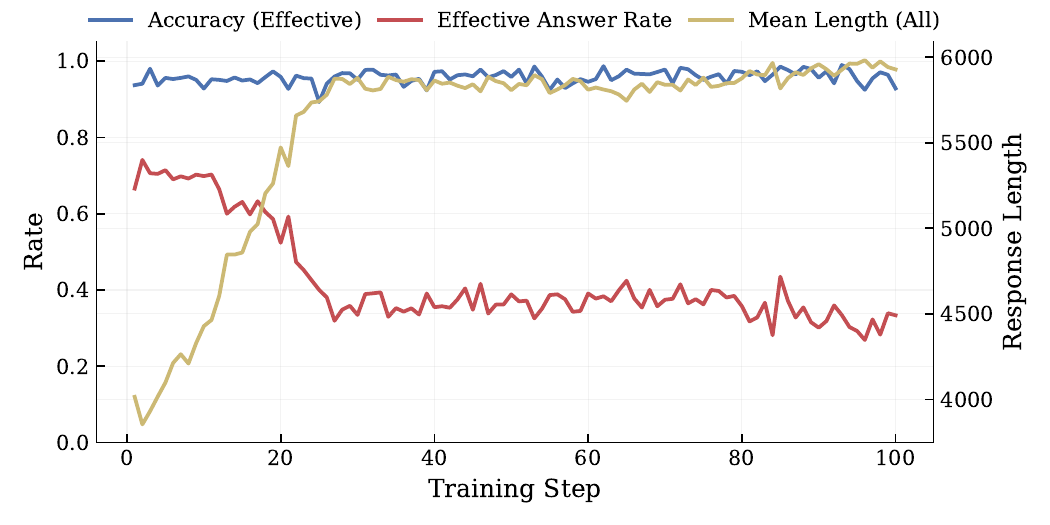}
        \caption{Effective answer rate and accuracy}
        \label{fig:traj_1.7b_long_2}
    \end{subfigure}
    \caption{Training trajectories under Long-Reward for \textbf{Qwen3-1.7B}.}
    \label{fig:traj_1.7b_long}
\end{figure}
\vspace{-2em}
\begin{figure}[H]
    \centering
    \begin{subfigure}[t]{0.48\linewidth}
        \centering
        \includegraphics[width=\linewidth]{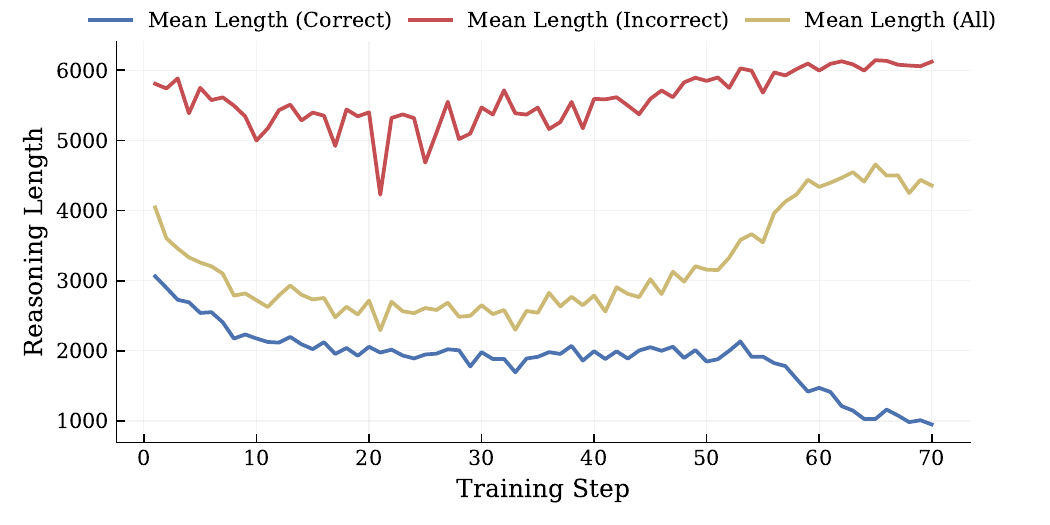}
        \caption{Reasoning length by correctness}
        \label{fig:traj_4b_short_1}
    \end{subfigure}
    \hfill
    \begin{subfigure}[t]{0.48\linewidth}
        \centering
        \includegraphics[width=\linewidth]{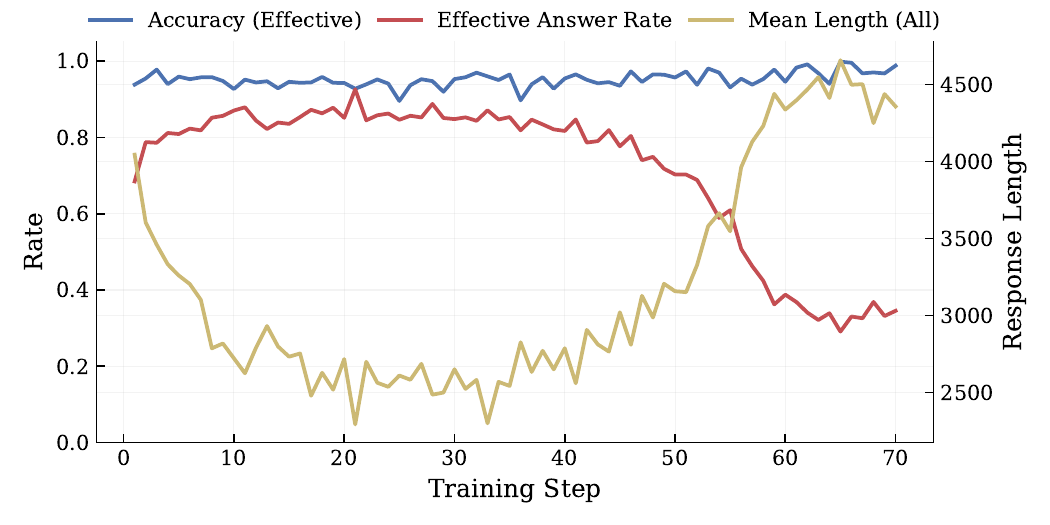}
        \caption{Effective answer rate and accuracy}
        \label{fig:traj_4b_short_2}
    \end{subfigure}
    \caption{Training trajectories under Short-Reward for \textbf{Qwen3-4B}.}
    \label{fig:traj_4b_short}
\end{figure}
\vspace{-2em}
\begin{figure}[H]
    \centering
    \begin{subfigure}[t]{0.48\linewidth}
        \centering
        \includegraphics[width=\linewidth]{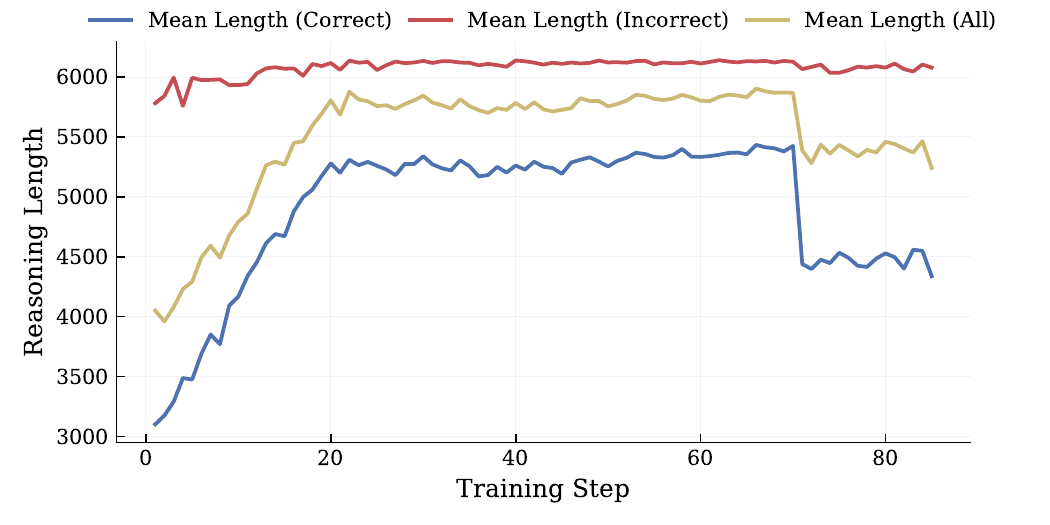}
        \caption{Reasoning length by correctness}
        \label{fig:traj_4b_long_1}
    \end{subfigure}
    \hfill
    \begin{subfigure}[t]{0.48\linewidth}
        \centering
        \includegraphics[width=\linewidth]{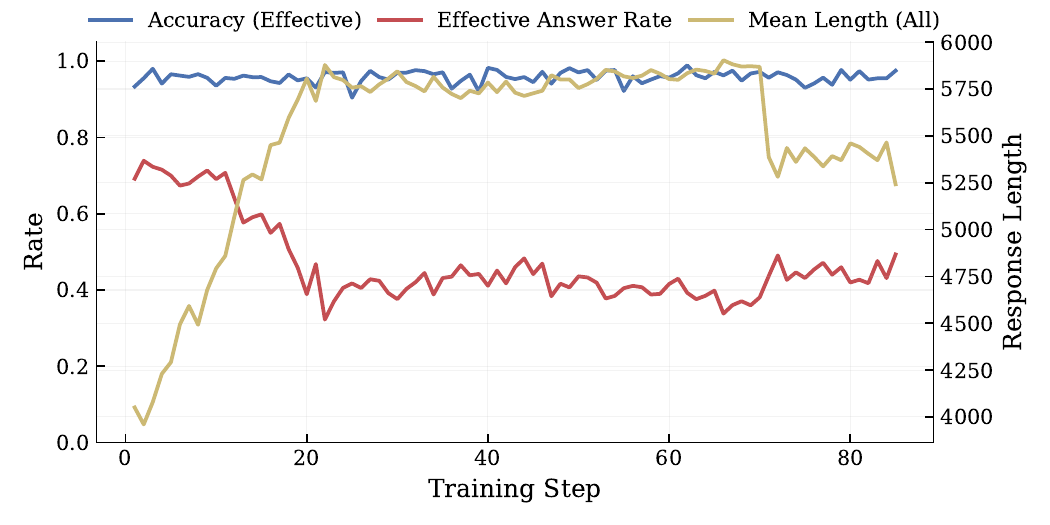}
        \caption{Effective answer rate and accuracy}
        \label{fig:traj_4b_long_2}
    \end{subfigure}
    \caption{Training trajectories under Long-Reward for \textbf{Qwen3-4B}.}
    \label{fig:traj_4b_long}
\end{figure}
\vspace{-2em}
\begin{figure}[H]
    \centering
    \begin{subfigure}[t]{0.48\linewidth}
        \centering
        \includegraphics[width=\linewidth]{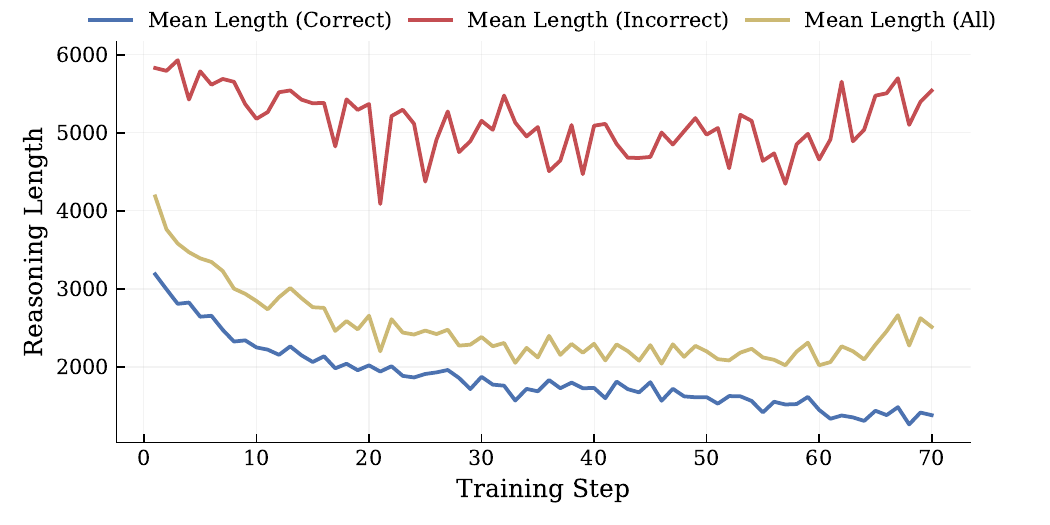}
        \caption{Reasoning length by correctness}
        \label{fig:traj_8b_short_1}
    \end{subfigure}
    \hfill
    \begin{subfigure}[t]{0.48\linewidth}
        \centering
        \includegraphics[width=\linewidth]{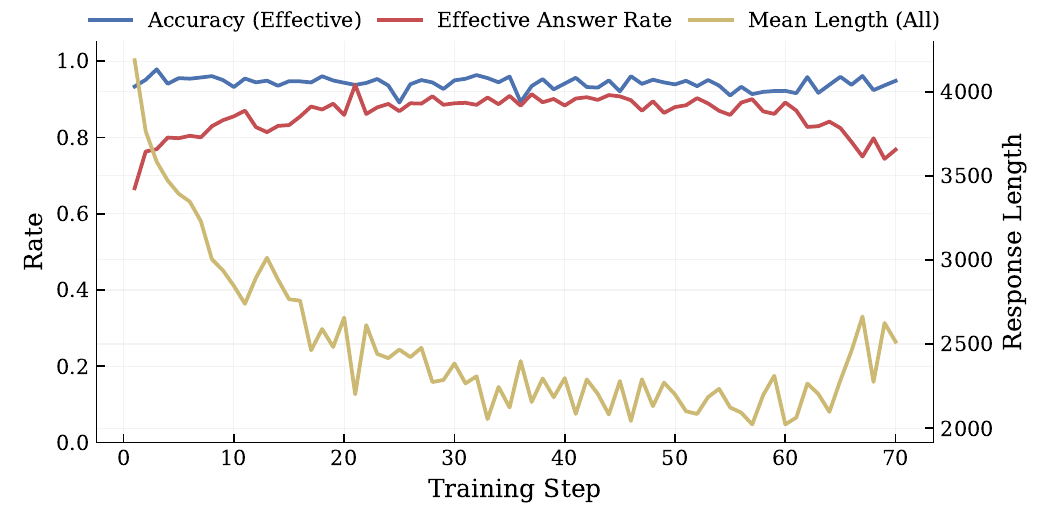}
        \caption{Effective answer rate and accuracy}
        \label{fig:traj_8b_short_2}
    \end{subfigure}
    \caption{Training trajectories under Short-Reward for \textbf{Qwen3-8B}.}
    \label{fig:traj_8b_short}
\end{figure}
\vspace{-2em}
\begin{figure}[H]
    \centering
    \begin{subfigure}[t]{0.48\linewidth}
        \centering
        \includegraphics[width=\linewidth]{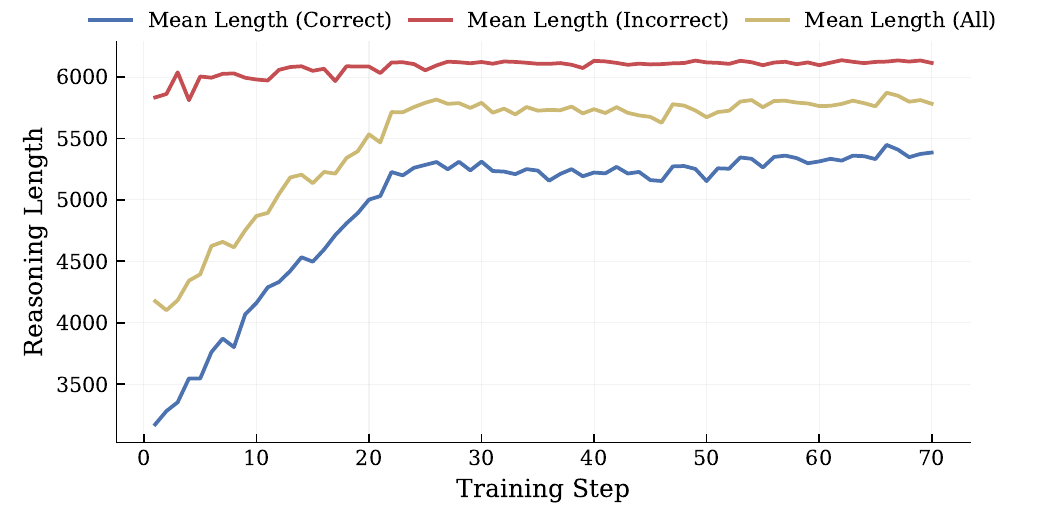}
        \caption{Reasoning length by correctness}
        \label{fig:traj_8b_long_1}
    \end{subfigure}
    \hfill
    \begin{subfigure}[t]{0.48\linewidth}
        \centering
        \includegraphics[width=\linewidth]{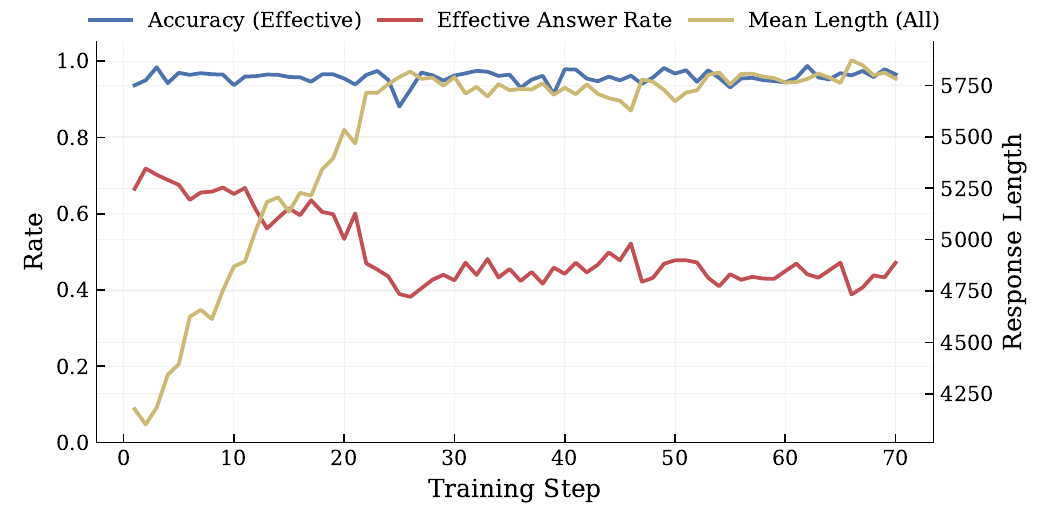}
        \caption{Effective answer rate and accuracy}
        \label{fig:traj_8b_long_2}
    \end{subfigure}
    \caption{Training trajectories under Long-Reward for \textbf{Qwen3-8B}.}
    \label{fig:traj_8b_long}
\end{figure}

Figures~\ref{fig:traj_1.7b_short}--\ref{fig:traj_8b_long} present the training trajectories for Qwen3-1.7B, Qwen3-4B, and Qwen3-8B under short-reward and long-reward training. For Qwen3-1.7B with short-reward training (Figure~\ref{fig:traj_1.7b_short}), mean reasoning length gradually decreases before collapsing abruptly after approximately step \(35\). Meanwhile, accuracy among effective answers drops substantially in the later stage, indicating that aggressive length compression eventually degrades reasoning performance. Under long-reward training (Figure~\ref{fig:traj_1.7b_long}), mean reasoning length rapidly increases and then remains close to the maximum response length, while the effective answer rate declines to around \(0.35\) and accuracy among effective answers remains relatively stable. For Qwen3-4B under short-reward training (Figure~\ref{fig:traj_4b_short}), mean reasoning length initially decreases but later rebounds, primarily as incorrect responses remain substantially longer while correct responses continue to shorten. This rebound coincides with a marked decline in the effective answer rate, whereas accuracy among effective answers remains stable. Under long-reward training (Figure~\ref{fig:traj_4b_long}), reasoning length similarly increases rapidly and remains at a high level, accompanied by a lower effective answer rate but little change in effective-answer accuracy. Qwen3-8B exhibits a more stable trajectory under short-reward training (Figure~\ref{fig:traj_8b_short}). Mean reasoning length decreases steadily and remains low, with only a modest late-stage rebound compared with the smaller models, while accuracy among effective answers stays consistently high. In contrast, under long-reward training (Figure~\ref{fig:traj_8b_long}), reasoning length again increases rapidly and stabilizes near the upper range, while the effective answer rate decreases to around \(0.4\) despite nearly unchanged accuracy among effective answers. This suggests that the additional reasoning length does not contribute to significant improvement in reasoning capability but instead increases the proportion of responses that fail to reach an answer within the maximum length.

\subsection{Cyclic Reasoning: Definitions}
\label{app:cyclic_reasoning}
 
\paragraph{Definition of Cyclic Segment.}
To approximate the prevalence of unproductive repetitive behavior in model responses, we introduce
the notion of a \textit{Cyclic Segment}. Given a response, we first segment it into sentences
using punctuation boundaries. A Cyclic Segment is defined as the substring spanning from the
first occurrence of any cyclic keyword in a sentence to the end of that sentence, where the
cyclic keywords are defined as follows (matched case-insensitively):
\begin{center}
    \texttt{wait},\quad \texttt{but},\quad \texttt{if},\quad
    \texttt{check},\quad \texttt{again},\quad \texttt{alternatively}
\end{center}

\noindent Two supplementary rules apply:
\begin{enumerate}
    \item[(1)] If multiple cyclic keywords appear within the same sentence, they are counted
          as a single Cyclic Segment.
    \item[(2)] Cyclic keywords appearing in different sentences are counted as separate
          Cyclic Segments.
\end{enumerate}

\paragraph{Cyclic Reasoning Ratio (CRR).}
We define the \textit{Cyclic Reasoning Ratio} (CRR) to measure the proportion of sentences
attributed to cyclic behavior within a response. For a given question~\(q\), let
\(\mathcal{A}(q)\) denote the set of sampled responses. For each response
\(a \in \mathcal{A}(q)\), let \(N_{\mathrm{cyc}}(a)\) denote the number of sentences
containing at least one Cyclic Segment and \(N_{\mathrm{tot}}(a)\) denote the total number
of sentences. The CRR for question~\(q\) is defined as:
\begin{equation}
    \mathrm{CRR}(q) = \frac{1}{|\mathcal{A}(q)|}
    \sum_{a \in \mathcal{A}(q)} \frac{N_{\mathrm{cyc}}(a)}{N_{\mathrm{tot}}(a)}
    \label{eq:crr}
\end{equation}
A higher CRR indicates that a greater proportion of sentences in the 
model output consist of cyclic reasoning patterns rather than 
substantive inference, and serves as a proxy for reasoning inefficiency.
\begin{remarkbox}[Remark]
These keywords were identified through manual inspection of 
model responses from unstable training intervals, where recurring 
patterns of unproductive reasoning loops were observed, such as 
re-checking a verified computation or reproducing near-identical 
paragraphs across successive reasoning steps. Individual occurrences 
of these keywords may correspond to productive reasoning, e.g., 
legitimate backtracking or self-verification. Our analysis therefore 
focuses on the \emph{change} in their density \(\Delta\)CRR between 
stable and unstable intervals, capturing the growth of repetitive 
patterns rather than isolated normal usage. We provide representative cases in 
Appendix~\ref{app:cyclic_cases}, illustrating how keyword-flagged sentences correspond to semantic repetition during unstable training intervals.
\end{remarkbox}
\paragraph{Training Interval Classification.}
To compare model behavior across different stages of training, we identify two distinct
step intervals that reflect different patterns in the training dynamics.
 
\textbf{Stable Interval}: a training interval in which the training metrics remain in their normal range and no anomalous training dynamics are observed.
 
\textbf{Unstable Interval}: a training interval in which the training dynamics exhibit clear anomalies, indicating that the training process has deviated from its expected trajectory.
 
\subsection{Cyclic Reasoning: Results}
We exclude Qwen3-1.7B under short-reward training from this analysis, as this setting exhibits length collapse rather than cyclic reasoning. A detailed case study is provided in Appendix~\ref{app:spcl_case_1.7b_short}.
 
\begin{table}[H]
    \caption{Step intervals selected as the \textit{Stable Interval} and
             \textit{Unstable Interval} under short-reward and long-reward training
             for each model.}
    \label{tab:phase_intervals}
    \centering
    \begin{tabular}{lcccc}
        \toprule
        \multirow{2}{*}[-3.0pt]{Model}
            & \multicolumn{2}{c}{Short-reward}
            & \multicolumn{2}{c}{Long-reward} \\
        \cmidrule(r){2-3} \cmidrule(l){4-5}
            & Stable Interval & Unstable Interval
            & Stable Interval & Unstable Interval \\
        \midrule
        Qwen3-1.7B & N/A            & N/A            & Steps 1--15  & Steps 31--45 \\
        Qwen3-4B   & Steps 1--10    & Steps 60--70   & Steps 1--15   & Steps 31--45   \\
        Qwen3-8B   & Steps 1--10    & Steps 60--70   & Steps 1--15   & Steps 31--45   \\
        \bottomrule
    \end{tabular}
\end{table}

\begin{figure}[H]
    \centering
    \begin{subfigure}[t]{0.48\linewidth}
        \centering
        \includegraphics[width=\linewidth]{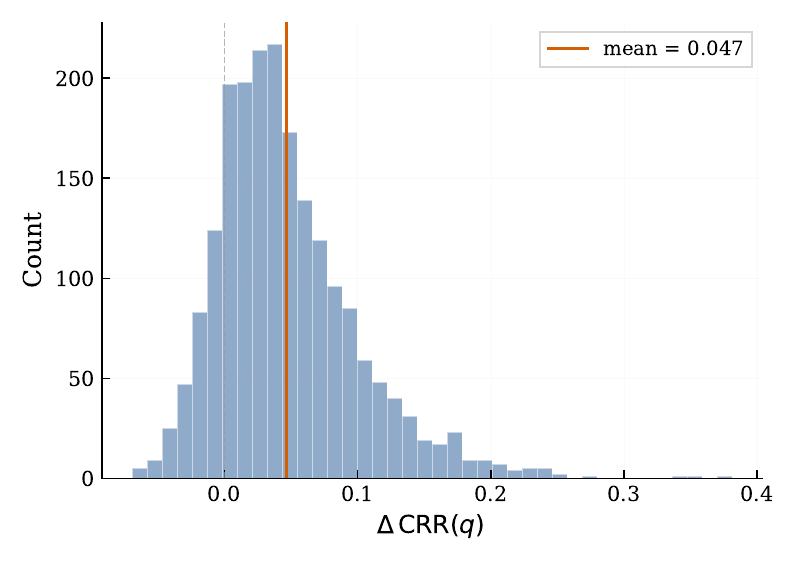}
        \caption{Distribution of \(\Delta\,\mathrm{CRR}\)}
        \label{fig:crr_1.7b_long_dist}
    \end{subfigure}
    \hfill
    \begin{subfigure}[t]{0.48\linewidth}
        \centering
        \includegraphics[width=\linewidth]{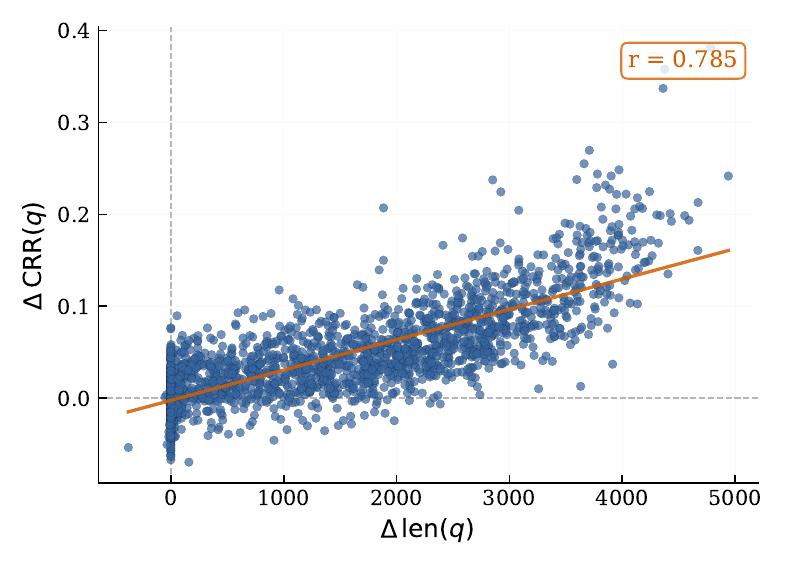}
        \caption{\(\Delta\,\mathrm{CRR}\) vs \(\Delta\,\mathrm{len}\)}
        \label{fig:crr_1.7b_long_scatter}
    \end{subfigure}
    \caption{CRR comparison for \textbf{Qwen3-1.7B} under Long-Reward training.}
    \label{fig:crr_1.7b_long}
\end{figure}

\begin{figure}[H]
    \centering
    \begin{subfigure}[t]{0.48\linewidth}
        \centering
        \includegraphics[width=\linewidth]{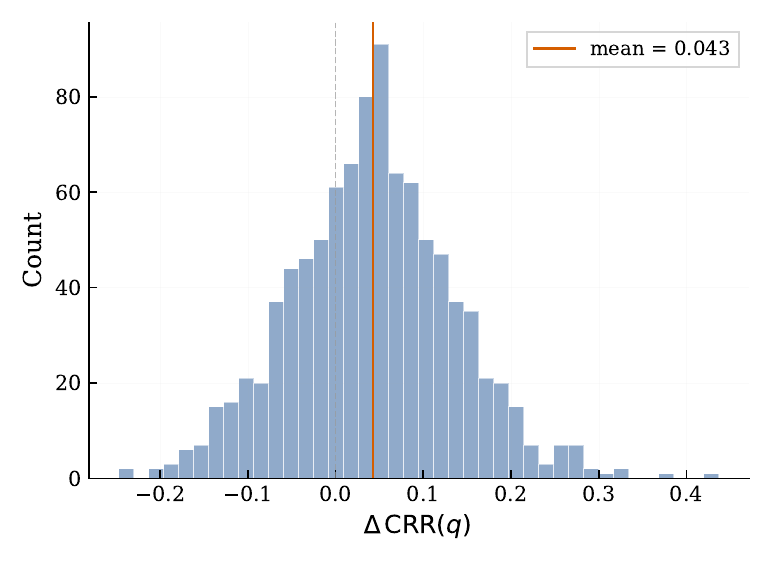}
        \caption{Distribution of \(\Delta\,\mathrm{CRR}\)}
        \label{fig:crr_4b_short_dist}
    \end{subfigure}
    \hfill
    \begin{subfigure}[t]{0.48\linewidth}
        \centering
       \includegraphics[width=\linewidth]{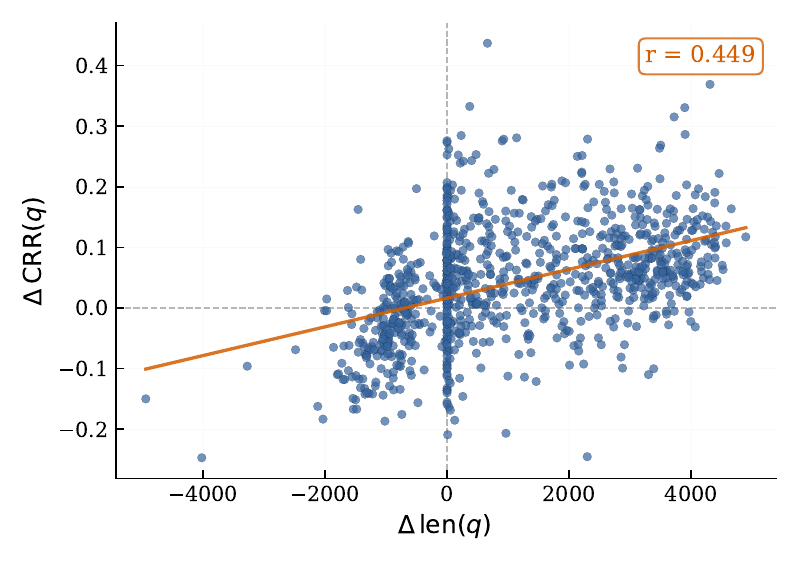}
        \caption{\(\Delta\,\mathrm{CRR}\) vs \(\Delta\,\mathrm{len}\)}
        \label{fig:crr_4b_short_scatter}
    \end{subfigure}
    \caption{CRR comparison for \textbf{Qwen3-4B} under Short-Reward training.}
    \label{fig:crr_4b_short}
\end{figure}

\begin{figure}[H]
    \centering
    \begin{subfigure}[t]{0.48\linewidth}
        \centering
        \includegraphics[width=\linewidth]{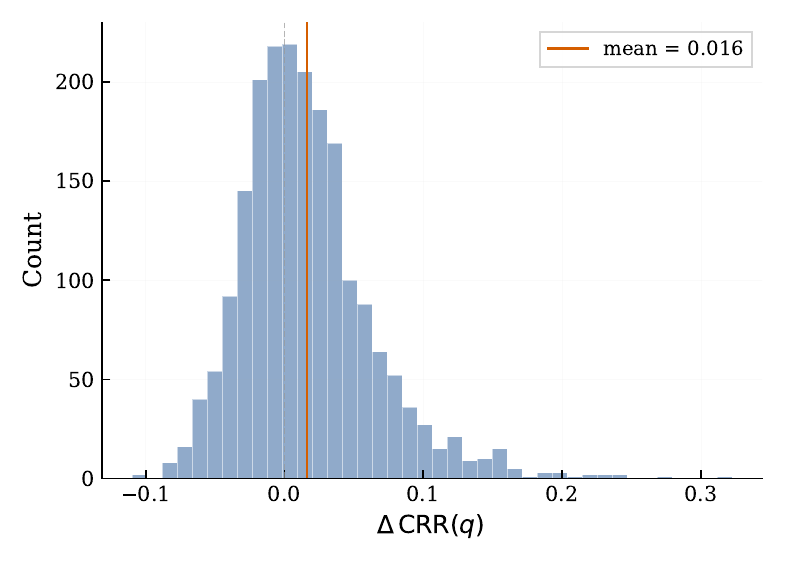}
        \caption{Distribution of \(\Delta\,\mathrm{CRR}\)}
        \label{fig:crr_4b_long_dist}
    \end{subfigure}
    \hfill
    \begin{subfigure}[t]{0.48\linewidth}
        \centering
        \includegraphics[width=\linewidth]{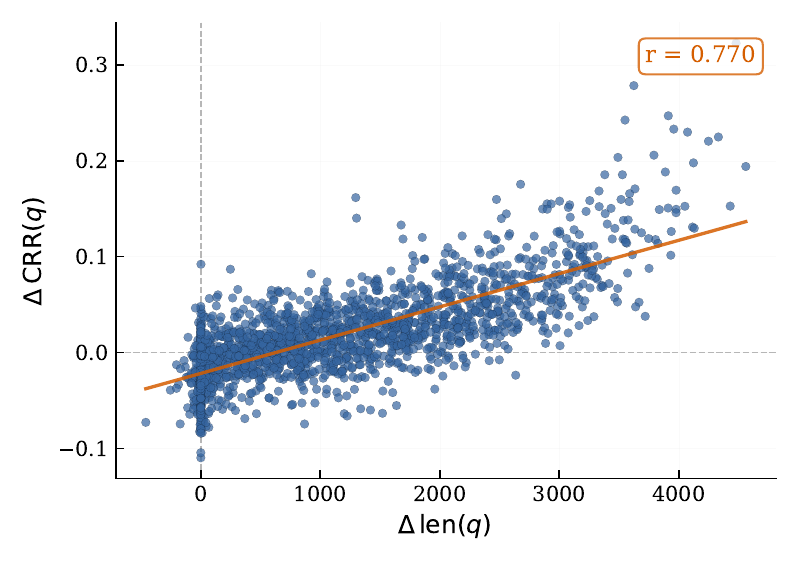}
        \caption{\(\Delta\,\mathrm{CRR}\) vs \(\Delta\,\mathrm{len}\)}
        \label{fig:crr_4b_long_scatter}
    \end{subfigure}
    \caption{CRR comparison for \textbf{Qwen3-4B} under Long-Reward training.}
    \label{fig:crr_4b_long}
\end{figure}

\begin{figure}[H]
    \centering
    \begin{subfigure}[t]{0.48\linewidth}
        \centering
        \includegraphics[width=\linewidth]{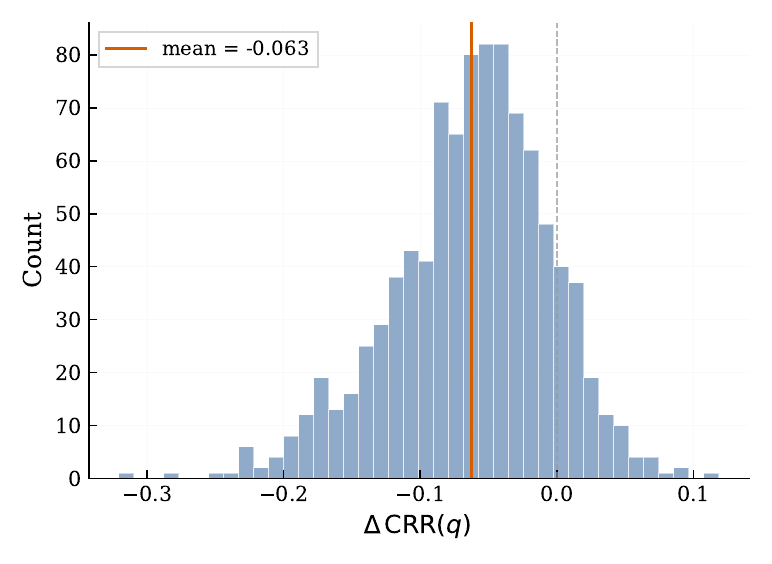}
        \caption{Distribution of \(\Delta\,\mathrm{CRR}\)}
        \label{fig:crr_8b_short_dist}
    \end{subfigure}
    \hfill
    \begin{subfigure}[t]{0.48\linewidth}
        \centering
        \includegraphics[width=\linewidth]{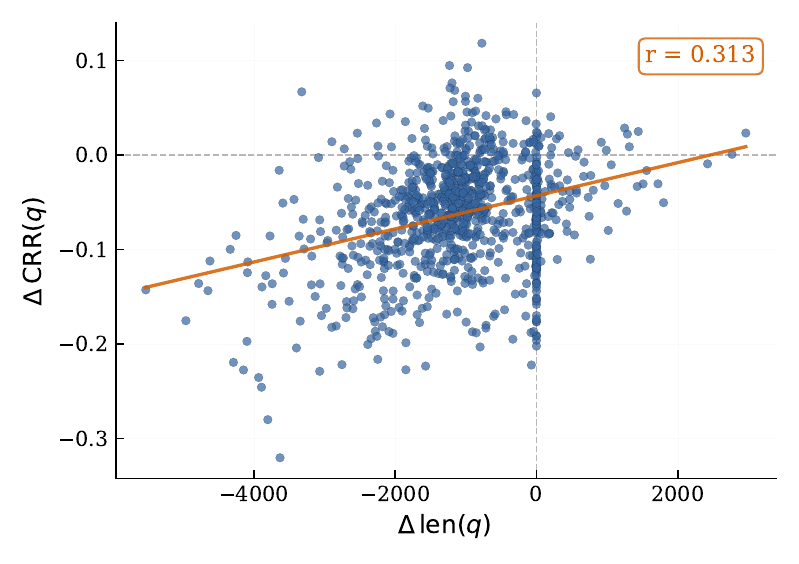}
        \caption{\(\Delta\,\mathrm{CRR}\) vs \(\Delta\,\mathrm{len}\)}
        \label{fig:crr_8b_short_scatter}
    \end{subfigure}
    \caption{CRR comparison for \textbf{Qwen3-8B} under Short-Reward training.}
    \label{fig:crr_8b_short}
\end{figure}

\begin{figure}[H]
    \centering
    \begin{subfigure}[t]{0.48\linewidth}
        \centering
        \includegraphics[width=\linewidth]{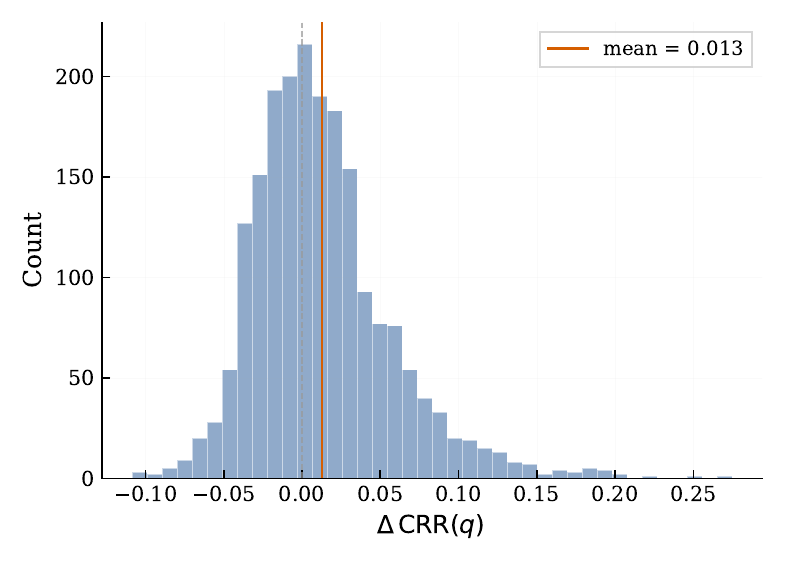}
        \caption{Distribution of \(\Delta\,\mathrm{CRR}\)}
        \label{fig:crr_8b_long_dist}
    \end{subfigure}
    \hfill
    \begin{subfigure}[t]{0.48\linewidth}
        \centering
        \includegraphics[width=\linewidth]{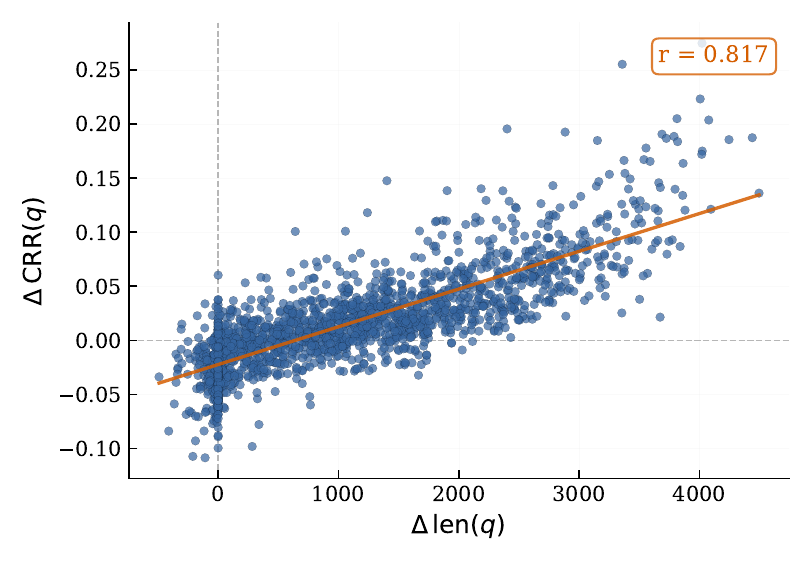}
        \caption{\(\Delta\,\mathrm{CRR}\) vs \(\Delta\,\mathrm{len}\)}
        \label{fig:crr_8b_long_scatter}
    \end{subfigure}
    \caption{CRR comparison for \textbf{Qwen3-8B} under Long-Reward training.}
    \label{fig:crr_8b_long}
\end{figure}

To examine whether the length increases observed during unstable training intervals are associated with cyclic reasoning, we compare the per-question CRR between the Stable and Unstable Intervals defined in Table~\ref{tab:phase_intervals}. For each matched question, we compute
\(\Delta\,\mathrm{CRR} = \mathrm{CRR}_{\text{unstable}} - \mathrm{CRR}_{\text{stable}}\) and \(\Delta\,\mathrm{len} = \mathrm{len}_{\text{unstable}} - \mathrm{len}_{\text{stable}}\).
A positive \(\Delta\,\mathrm{CRR}\) indicates that the proportion of cyclic segments increased during the unstable phase.
Figures~\ref{fig:crr_1.7b_long} --~\ref{fig:crr_8b_long} present the distribution of \(\Delta\,\mathrm{CRR}\) alongside its correlation with \(\Delta\,\mathrm{len}\). The results suggest that longer reasoning traces do not always lead to deeper or more productive thinking, and may instead introduce redundant cyclic patterns that degrade reasoning efficiency. In particular, the distribution of \(\Delta\,\mathrm{CRR}\) has a positive mean under long-reward settings, indicating that cyclic reasoning becomes more prevalent during unstable training phases. The positive correlation between \(\Delta\,\mathrm{len}\) and \(\Delta\,\mathrm{CRR}\) further confirms that the observed length increases are associated with a higher proportion of repetitive reasoning patterns rather than deeper exploration. 

\vfill

\vfill
\vfill

\vfill
\vfill
\section{Accuracy Gap Analysis}
\label{app:directional_signal}

In this part, we provide the full directional signal analysis between CARE and GRPO, across AIME~2025, AIME~2026, HMMT~2026, and MATH500 for Qwen3-1.7B, 4B and 8B.

\subsection{Metrics and Difficulty Partitioning}

For each question \(q\), the accuracy gap \(\Delta(q) = |\bar{y}_{\mathcal{S}}(q) - \bar{y}_{\mathcal{L}}(q)|\) measures the absolute accuracy difference between the shorter and longer response halves, capturing the degree of the length sensitivity of each question. We compute \(\Delta(q)\) for GRPO and CARE and aggregate it within each difficulty group across all benchmarks. For MATH500, questions are partitioned by the dataset's built-in difficulty labels (Levels~1--5). Since AIME~2025, AIME~2026, and HMMT~2026 lack predefined labels, difficulty is estimated from the base model's per-question Pass@1 and partitioned into three subsets: \([0,\,0.25]\), \((0.25,\,0.75]\), and \((0.75,\,1.0]\). The two middle subsets \((0.25,\,0.5]\) and \((0.5,\,0.75]\) are merged into \((0.25,\,0.75]\) to ensure sufficient sample sizes, as these benchmarks contain few questions and finer splitting would introduce high variance in \(\Delta(q)\) estimates.

\subsection{Experiment Results}
\begin{figure}[H]
  \centering
  \begin{subfigure}[t]{0.49\linewidth}
    \centering
    \includegraphics[width=\linewidth]{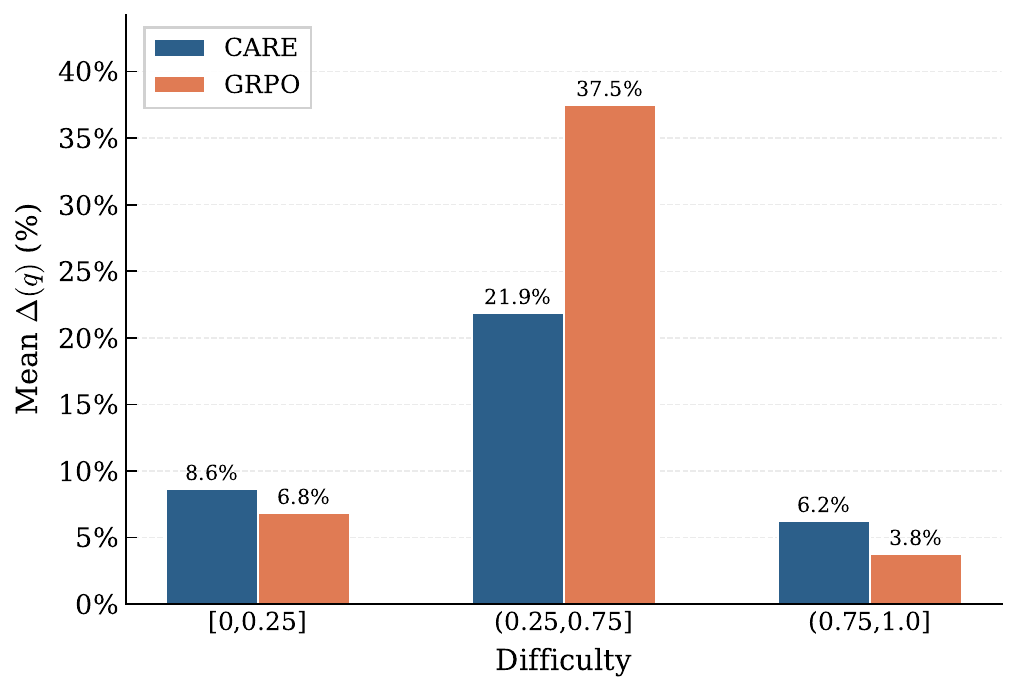}
    \caption{AIME 2025}
  \end{subfigure}\hfill
  \begin{subfigure}[t]{0.49\linewidth}
    \centering
    \includegraphics[width=\linewidth]{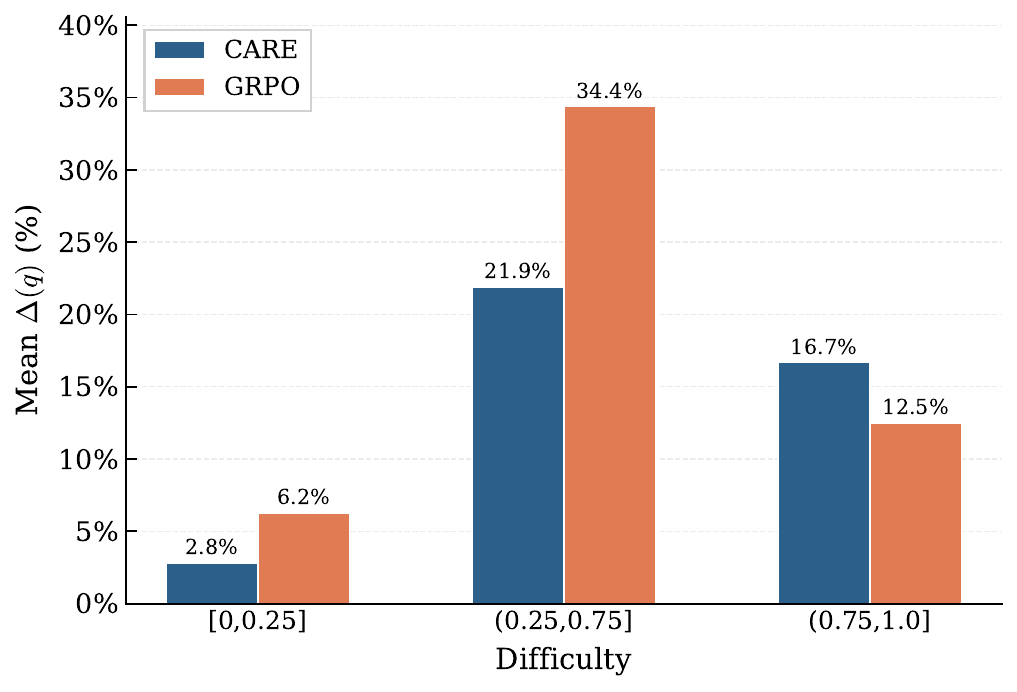}
    \caption{AIME 2026}
  \end{subfigure}
 
  \begin{subfigure}[t]{0.49\linewidth}
    \centering
    \includegraphics[width=\linewidth]{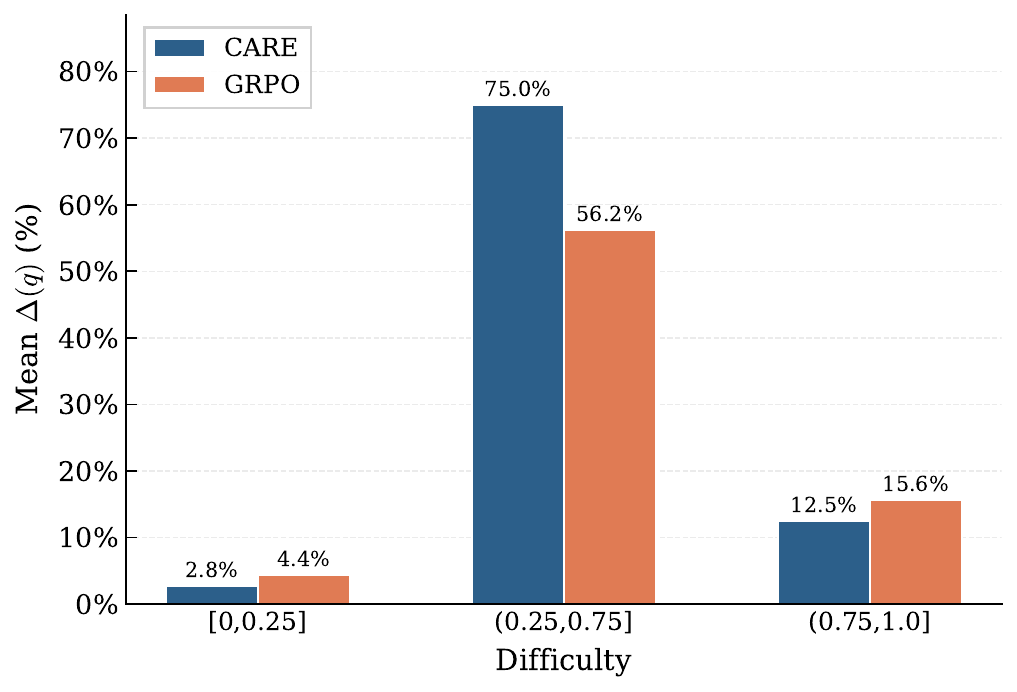}
    \caption{HMMT 2026}
  \end{subfigure}\hfill
  \begin{subfigure}[t]{0.49\linewidth}
    \centering
    \includegraphics[width=\linewidth]{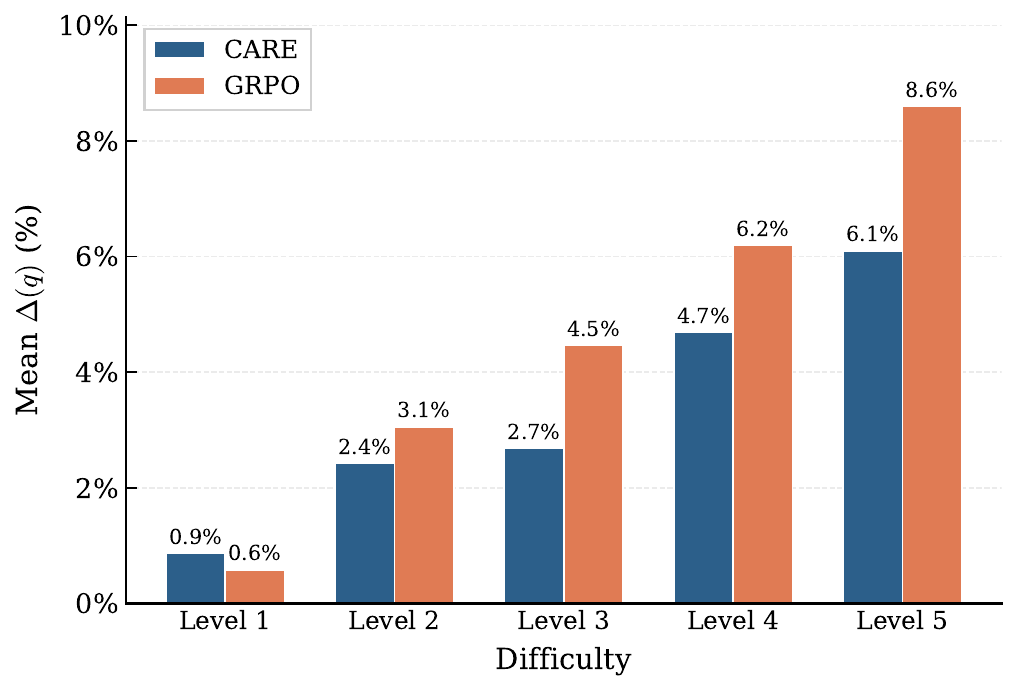}
    \caption{MATH500}
  \end{subfigure}
  \caption{Accuracy gap \(\Delta(q)\)  at each difficulty group for \textbf{Qwen3-1.7B}.}
  \label{fig:app_directional_1.7b}
\end{figure}

\begin{figure}[H]
  \centering
  \begin{subfigure}[t]{0.49\linewidth}
    \centering
    \includegraphics[width=\linewidth]{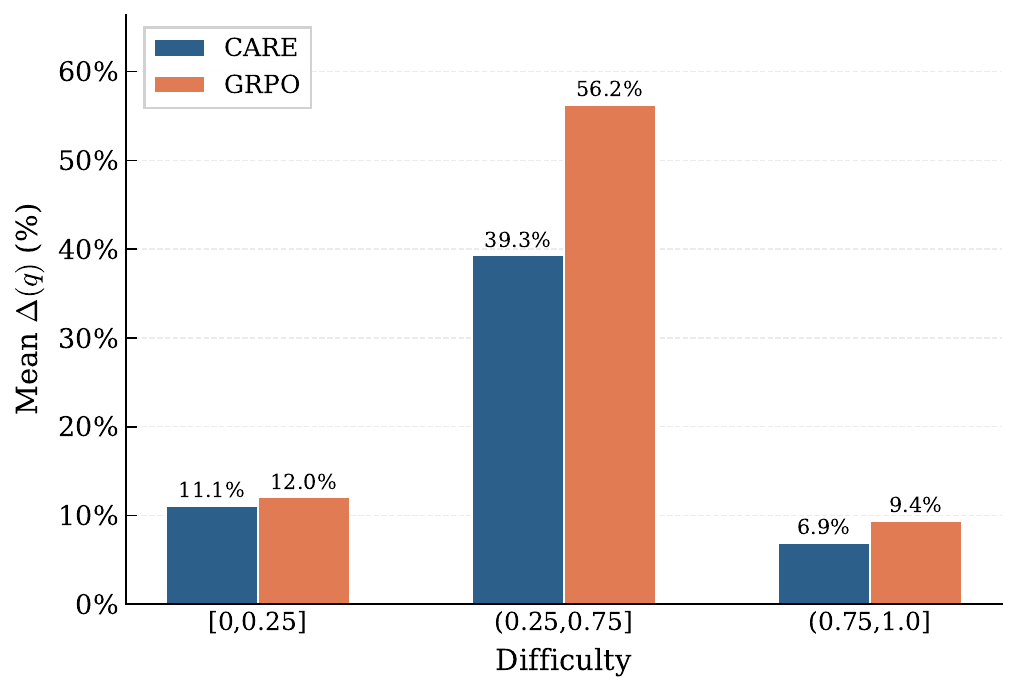}
    \caption{AIME 2025}
  \end{subfigure}\hfill
  \begin{subfigure}[t]{0.49\linewidth}
    \centering
    \includegraphics[width=\linewidth]{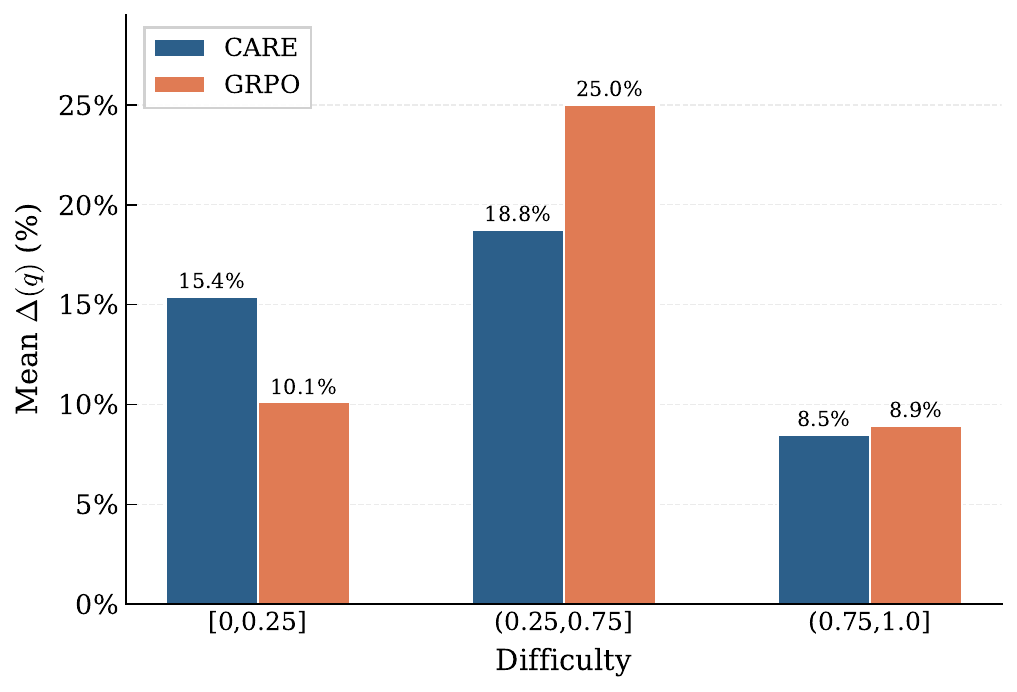}
    \caption{AIME 2026}
  \end{subfigure}
 
  \begin{subfigure}[t]{0.49\linewidth}
    \centering
    \includegraphics[width=\linewidth]{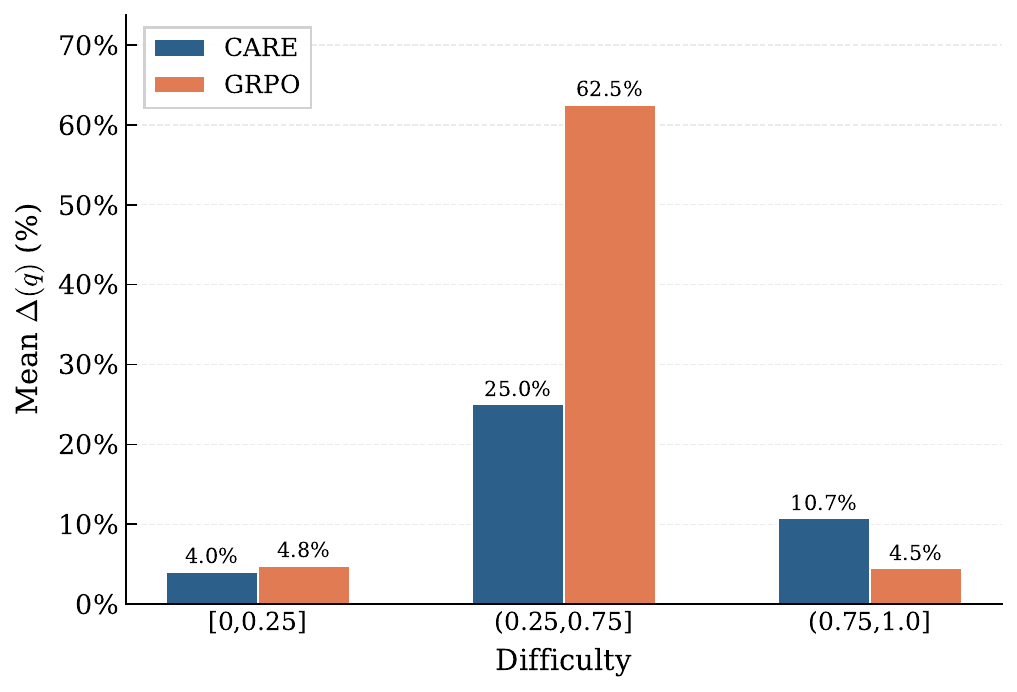}
    \caption{HMMT 2026}
  \end{subfigure}\hfill
  \begin{subfigure}[t]{0.49\linewidth}
    \centering
    \includegraphics[width=\linewidth]{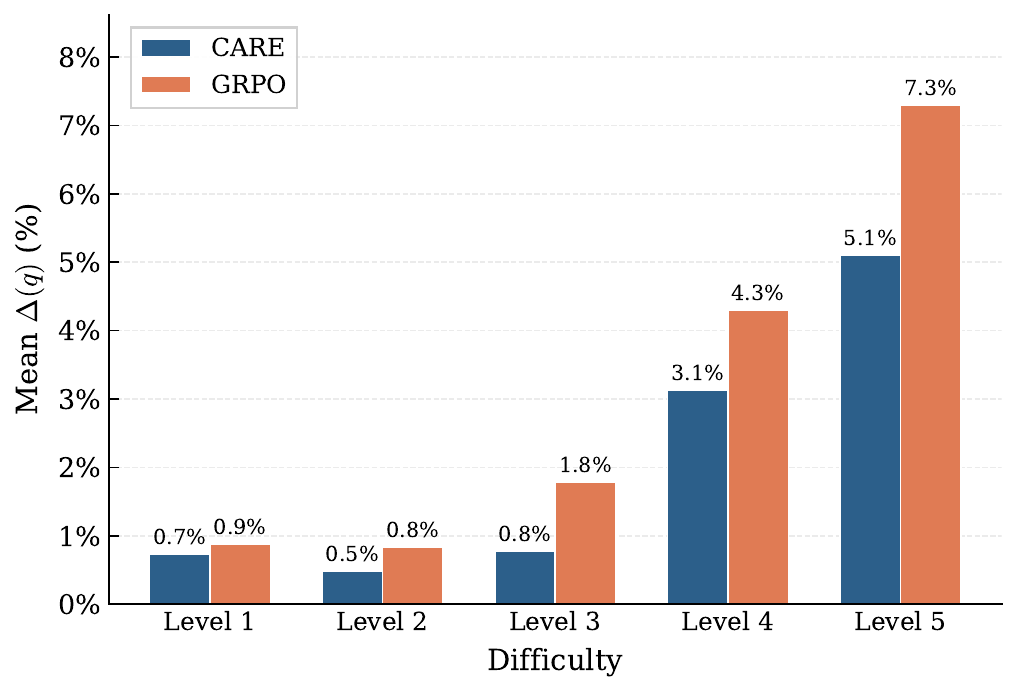}
    \caption{MATH500}
  \end{subfigure}
  \caption{Accuracy gap \(\Delta(q)\) at each difficulty group for \textbf{Qwen3-4B}.}
  \label{fig:app_directional_4b}
\end{figure}

\begin{figure}[H]
  \centering
  \begin{subfigure}[t]{0.49\linewidth}
    \centering
    \includegraphics[width=\linewidth]{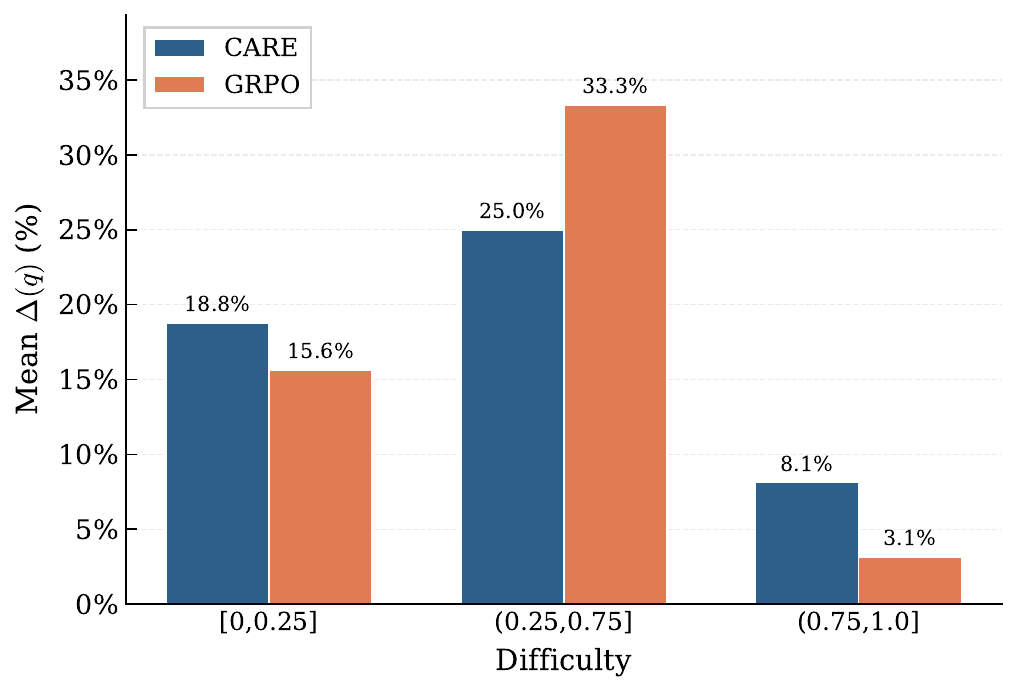}
    \caption{AIME 2025}
  \end{subfigure}\hfill
  \begin{subfigure}[t]{0.49\linewidth}
    \centering
    \includegraphics[width=\linewidth]{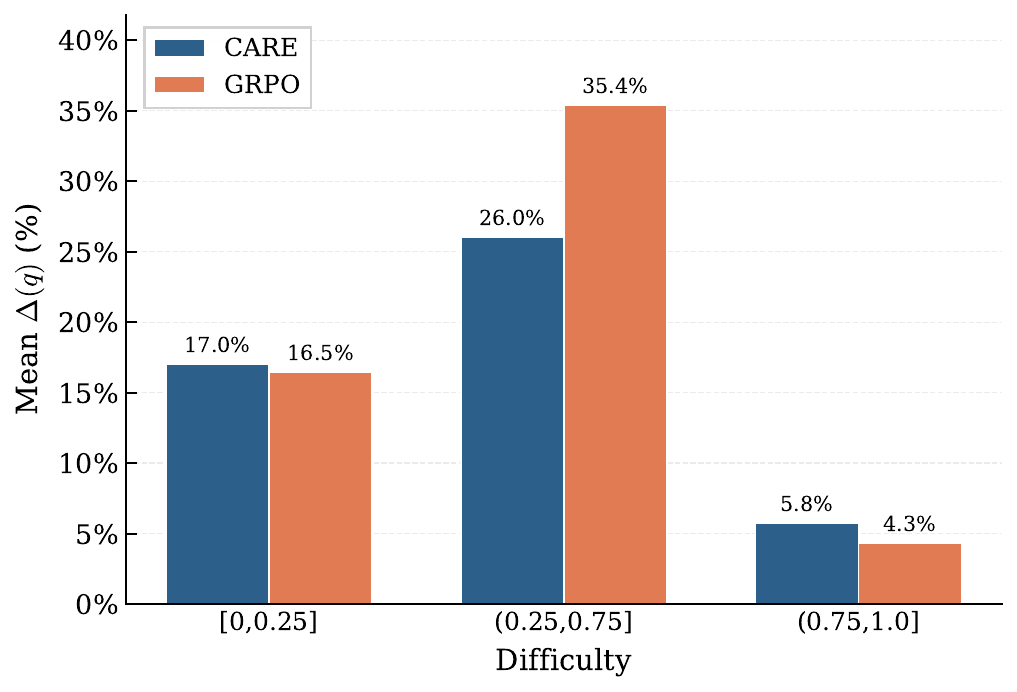}
    \caption{AIME 2026}
  \end{subfigure}
 
  \begin{subfigure}[t]{0.49\linewidth}
    \centering
    \includegraphics[width=\linewidth]{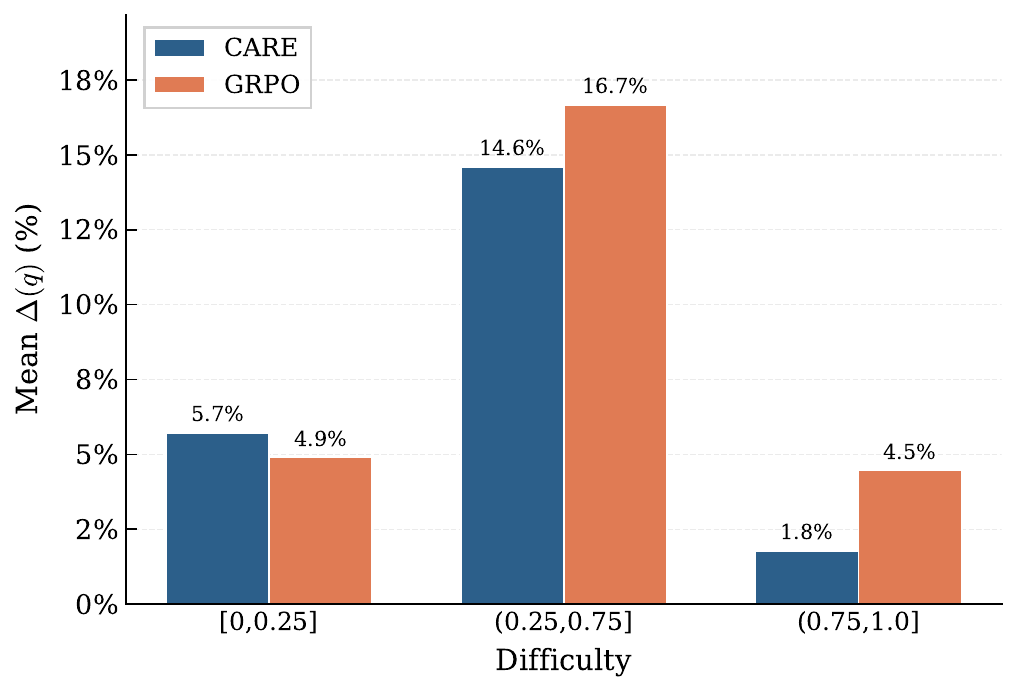}
    \caption{HMMT 2026}
  \end{subfigure}\hfill
  \begin{subfigure}[t]{0.49\linewidth}
    \centering
    \includegraphics[width=\linewidth]{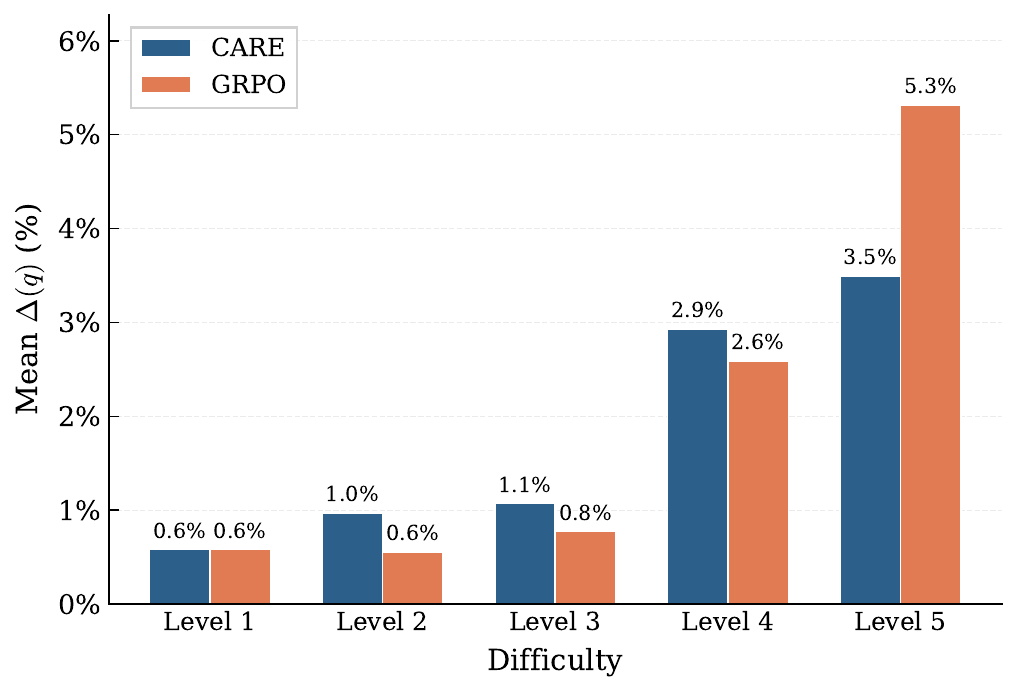}
    \caption{MATH500}
  \end{subfigure}
  \caption{Accuracy gap \(\Delta(q)\) at each difficulty group for \textbf{Qwen3-8B}.}
  \label{fig:app_directional_8b}
\end{figure}

Figures~\ref{fig:app_directional_1.7b} --~\ref{fig:app_directional_8b} report the mean accuracy gap \(\Delta(q)\) at each difficulty group across different benchmarks. Over the majority of benchmarks, CARE achieves a lower mean \(\Delta(q)\) than GRPO, indicating that the directional signal successfully reduces the accuracy gap between shorter and longer response halves. This reduction is most consistent on the partially solvable subset \((0.25,\,0.75]\) and on harder MATH500 levels, with the clearest consistent reduction appearing at Level~5, where such length sensitivity is more prevalent. On the hardest subsets \([0,0.25]\), both methods exhibit comparably lower \(\Delta(q)\) than the partially solvable subsets, confirming that hard questions for models are insensitive to length variation.
\vfill
\vfill
\clearpage
\vspace{-1.0em}

\clearpage
\subsubsection{Benchmark-Specific Analysis}
On AIME~2025 and AIME~2026, the reduction in \(\Delta(q)\) generally concentrates on the partially solvable subset, with this pattern particularly clear for Qwen3-1.7B and Qwen3-4B. For Qwen3-4B, CARE reduces the partially solvable gap from \(56.2\%\) to \(39.3\%\) on AIME~2025 and from \(25.0\%\) to \(18.8\%\) on AIME~2026, while the hardest and easiest subsets show smaller or reversed differences. On MATH500, the reduction is more evident at harder difficulty levels for Qwen3-1.7B and Qwen3-4B, whereas for Qwen3-8B the clearest decrease appears at Level~5, from \(5.3\%\) to \(3.5\%\).

A different pattern emerges on HMMT~2026. CARE achieves lower \(\Delta(q)\) on the hardest and partially solvable subsets for Qwen3-4B, on the hardest and easiest subsets for Qwen3-1.7B, and on the partially solvable and easiest subsets for Qwen3-8B. Since all three base models attain relatively low accuracy on HMMT~2026, questions are concentrated disproportionately in the hardest subset, leaving the remaining subsets with limited sample sizes. These smaller subsets make \(\Delta(q)\) estimates more susceptible to per-question variation, although CARE still achieves higher overall Pass@1 than GRPO on HMMT~2026, as shown in Table~\ref{tab:main}.

\section{Length Consistency and Accuracy}
\label{app:length_consistency}
In this section, we compute the dispersion \(D(q)\) for CARE and GRPO on Qwen3-1.7B, Qwen3-4B and Qwen3-8B across AIME 2025, AIME 2026, HMMT 2026, and MATH500, measuring how consistently the model converges to a stable reasoning length when correctly solving a question \(q\).

\subsection{Metrics and Difficulty Partitioning}
For each question \(q\), the dispersion \(D(q)\) is defined as the mean absolute deviation of correct reasoning lengths normalized by their mean, quantifying how tightly correct responses cluster around a common reasoning length. Formally, let \(\{L_i\}_{i \in C(q)}\) denote the lengths of correct responses for question \(q\) and \(\bar{L} = \frac{1}{|C(q)|}\sum_{i \in C(q)} L_i\) their mean; then \(D(q) = \frac{1}{|C(q)|}\sum_{i \in C(q)} |L_i - \bar{L}| \,/\, \bar{L}\). A lower \(D(q)\) indicates that the model converges to a more consistent reasoning depth when correctly solving that question. Questions are partitioned into four difficulty subsets based on the base model's per-question Pass@1: \([0,\,0.25]\), \((0.25,\,0.5]\), \((0.5,\,0.75]\), and \((0.75,\,1.0]\). A dash (--) in the tables indicates that no question falls within that difficulty subset under the base model's Pass@1 distribution.

\subsection{Experiment Results}
Tables~\ref{tab:app_dispersion_1.7b} --~\ref{tab:app_dispersion_8b} present the full Pass@1 and dispersion \(D(q)\) results. In these three models, CARE attains lower \(D(q)\) than GRPO in most subsets while maintaining comparable or higher Pass@1, although several
exceptions appear on AIME~2025, AIME~2026, and MATH500.
\vspace{-1.0em}
\begin{table}[H]
  \centering
  \caption{Pass@1\,(\%) and \(D(q)\)\,(\%) for \textbf{Qwen3-1.7B} across four benchmarks.}
  \label{tab:app_dispersion_1.7b}
  \small
  \renewcommand{\arraystretch}{0.95}
  \begin{tabular*}{\textwidth}{@{\extracolsep{\fill}} c S[table-format=2.2] S[table-format=2.2] S[table-format=2.2] S[table-format=2.2] S[table-format=2.2] S[table-format=2.2] S[table-format=2.2] S[table-format=2.2]}
    \toprule
    \multirow{3}{*}[-6pt]{Difficulty Subset}
      & \multicolumn{4}{c}{AIME 2025}
      & \multicolumn{4}{c}{AIME 2026} \\
    \cmidrule(lr){2-5} \cmidrule(lr){6-9}
      & \multicolumn{2}{c}{GRPO} & \multicolumn{2}{c}{CARE}
      & \multicolumn{2}{c}{GRPO} & \multicolumn{2}{c}{CARE} \\
    \cmidrule(lr){2-3} \cmidrule(lr){4-5} \cmidrule(lr){6-7} \cmidrule(lr){8-9}
      & {Pass@1} & {\(D(q)\)} & {Pass@1} & {\(D(q)\)}
      & {Pass@1} & {\(D(q)\)} & {Pass@1} & {\(D(q)\)} \\
    \midrule
    \([0,\,0.25]\)    &  8.04 & 14.48 & 11.61 & 12.27
                      &  2.78 &  4.85 &  4.86 & 9.81 \\
    \((0.25,\,0.5]\)  & 37.50 &  8.95 & 37.50 & 10.81
                      & 40.62 & 15.63 & 37.50 &  4.65 \\
    \((0.5,\,0.75]\)  & 77.08 & 14.49 & 91.67 & 16.30
                      & 76.56 & 29.15 & 73.44 & 29.05 \\
    \((0.75,\,1.0]\)  & 95.00 & 13.10 & 92.50 & 12.48
                      & 80.21 & 17.48 & 85.42 & 15.21 \\
    \midrule
    \multirow{3}{*}[-6pt]{Difficulty Subset}
      & \multicolumn{4}{c}{HMMT 2026}
      & \multicolumn{4}{c}{MATH500} \\
    \cmidrule(lr){2-5} \cmidrule(lr){6-9}
      & \multicolumn{2}{c}{GRPO} & \multicolumn{2}{c}{CARE}
      & \multicolumn{2}{c}{GRPO} & \multicolumn{2}{c}{CARE} \\
    \cmidrule(lr){2-3} \cmidrule(lr){4-5} \cmidrule(lr){6-7} \cmidrule(lr){8-9}
      & {Pass@1} & {\(D(q)\)} & {Pass@1} & {\(D(q)\)}
      & {Pass@1} & {\(D(q)\)} & {Pass@1} & {\(D(q)\)} \\
    \midrule
    \([0,\,0.25]\)    &  5.32 & 22.12 &  8.33 & 17.15
                      & 16.55 & 20.78 & 13.85 & 29.85 \\
    \((0.25,\,0.5]\)  & \multicolumn{1}{c}{--} & \multicolumn{1}{c}{--}
                      & \multicolumn{1}{c}{--} & \multicolumn{1}{c}{--}
                      & 44.85 & 16.29 & 50.37 & 16.15 \\
    \((0.5,\,0.75]\)  & 34.38 & 30.40 & 62.50 & 24.79
                      & 70.22 & 20.95 & 70.22 & 22.95 \\
    \((0.75,\,1.0]\)  & 92.19 & 21.80 & 92.19 & 17.35
                      & 97.35 & 14.45 & 96.77 & 13.83 \\
    \bottomrule
  \end{tabular*}
\end{table}
\vspace{-1.5em}
\begin{table}[H]
  \centering
  \caption{Pass@1\,(\%) and \(D(q)\)\,(\%) for \textbf{Qwen3-4B} across four benchmarks.}
  \label{tab:app_dispersion_4b}
  \small
  \renewcommand{\arraystretch}{0.95}
  \begin{tabular*}{\textwidth}{@{\extracolsep{\fill}} c S[table-format=2.2] S[table-format=2.2] S[table-format=2.2] S[table-format=2.2] S[table-format=2.2] S[table-format=2.2] S[table-format=2.2] S[table-format=2.2]}
    \toprule
    \multirow{3}{*}[-6pt]{Difficulty Subset}
      & \multicolumn{4}{c}{AIME 2025}
      & \multicolumn{4}{c}{AIME 2026} \\
    \cmidrule(lr){2-5} \cmidrule(lr){6-9}
      & \multicolumn{2}{c}{GRPO} & \multicolumn{2}{c}{CARE}
      & \multicolumn{2}{c}{GRPO} & \multicolumn{2}{c}{CARE} \\
    \cmidrule(lr){2-3} \cmidrule(lr){4-5} \cmidrule(lr){6-7} \cmidrule(lr){8-9}
      & {Pass@1} & {\(D(q)\)} & {Pass@1} & {\(D(q)\)}
      & {Pass@1} & {\(D(q)\)} & {Pass@1} & {\(D(q)\)} \\
    \midrule
    \([0,\,0.25]\)    & 12.02 & 12.53 & 14.90 &  9.83
                      & 16.35 & 12.58 & 18.27 & 13.17 \\
    \((0.25,\,0.5]\)  & 46.88 & 12.35 & 53.12 & 10.96
                      & 47.92 & 24.12 & 56.25 & 15.63 \\
    \((0.5,\,0.75]\)  & 65.00 & 31.16 & 67.50 & 21.36
                      & \multicolumn{1}{c}{--} & \multicolumn{1}{c}{--}
                      & \multicolumn{1}{c}{--} & \multicolumn{1}{c}{--} \\
    \((0.75,\,1.0]\)  & 90.62 & 11.25 & 90.62 & 10.06
                      & 87.50 & 12.47 & 88.84 & 9.84 \\
    \midrule
    \multirow{3}{*}[-6pt]{Difficulty Subset}
      & \multicolumn{4}{c}{HMMT 2026}
      & \multicolumn{4}{c}{MATH500} \\
    \cmidrule(lr){2-5} \cmidrule(lr){6-9}
      & \multicolumn{2}{c}{GRPO} & \multicolumn{2}{c}{CARE}
      & \multicolumn{2}{c}{GRPO} & \multicolumn{2}{c}{CARE} \\
    \cmidrule(lr){2-3} \cmidrule(lr){4-5} \cmidrule(lr){6-7} \cmidrule(lr){8-9}
      & {Pass@1} & {\(D(q)\)} & {Pass@1} & {\(D(q)\)}
      & {Pass@1} & {\(D(q)\)} & {Pass@1} & {\(D(q)\)} \\
    \midrule
    \([0,\,0.25]\)    &  7.25 & 14.99 &  7.75 & 14.48
                      & 22.95 & 15.67 & 24.29 & 14.81 \\
    \((0.25,\,0.5]\)  & \multicolumn{1}{c}{--} & \multicolumn{1}{c}{--}
                      & \multicolumn{1}{c}{--} & \multicolumn{1}{c}{--}
                      & 62.89 & 21.09 & 62.89 & 19.41 \\
    \((0.5,\,0.75]\)  & 56.25 & 22.46 & 81.25 & 21.56
                      & 77.26 & 19.73 & 83.85 & 18.83 \\
    \((0.75,\,1.0]\)  & 91.07 & 27.47 & 90.18 & 18.60
                      & 98.69 & 12.05 & 98.60 & 11.54 \\
    \bottomrule
  \end{tabular*}
\end{table}

\begin{table}[H]
  \centering
  \caption{Pass@1\,(\%) and \(D(q)\)\,(\%) for \textbf{Qwen3-8B} across four benchmarks.}
  \label{tab:app_dispersion_8b}
  \small
  \renewcommand{\arraystretch}{0.95}
  \begin{tabular*}{\textwidth}{@{\extracolsep{\fill}} c S[table-format=2.2] S[table-format=2.2] S[table-format=2.2] S[table-format=2.2] S[table-format=2.2] S[table-format=2.2] S[table-format=2.2] S[table-format=2.2]}
    \toprule
    \multirow{3}{*}[-6pt]{Difficulty Subset}
      & \multicolumn{4}{c}{AIME 2025}
      & \multicolumn{4}{c}{AIME 2026} \\
    \cmidrule(lr){2-5} \cmidrule(lr){6-9}
      & \multicolumn{2}{c}{GRPO} & \multicolumn{2}{c}{CARE}
      & \multicolumn{2}{c}{GRPO} & \multicolumn{2}{c}{CARE} \\
    \cmidrule(lr){2-3} \cmidrule(lr){4-5} \cmidrule(lr){6-7} \cmidrule(lr){8-9}
      & {Pass@1} & {\(D(q)\)} & {Pass@1} & {\(D(q)\)}
      & {Pass@1} & {\(D(q)\)} & {Pass@1} & {\(D(q)\)} \\
    \midrule
    \([0,\,0.25]\)    & 16.74 & 14.57 & 14.73 & 12.33
                      & 14.49 & 18.03 & 15.91 & 13.21 \\
    \((0.25,\,0.5]\)  & 62.50 & 20.27 & 65.62 & 16.31
                      & 60.94 & 20.92 & 63.54 & 17.84 \\
    \((0.5,\,0.75]\)  & 70.62 & 12.80 & 72.50 & 12.72
                      & \multicolumn{1}{c}{--} & \multicolumn{1}{c}{--}
                      & \multicolumn{1}{c}{--} & \multicolumn{1}{c}{--} \\
    \((0.75,\,1.0]\)  & 96.56 & 13.41 & 92.81 & 13.39
                      & 95.43 & 13.60 & 92.79 & 13.10 \\
    \midrule
    \multirow{3}{*}[-6pt]{Difficulty Subset}
      & \multicolumn{4}{c}{HMMT 2026}
      & \multicolumn{4}{c}{MATH500} \\
    \cmidrule(lr){2-5} \cmidrule(lr){6-9}
      & \multicolumn{2}{c}{GRPO} & \multicolumn{2}{c}{CARE}
      & \multicolumn{2}{c}{GRPO} & \multicolumn{2}{c}{CARE} \\
    \cmidrule(lr){2-3} \cmidrule(lr){4-5} \cmidrule(lr){6-7} \cmidrule(lr){8-9}
      & {Pass@1} & {\(D(q)\)} & {Pass@1} & {\(D(q)\)}
      & {Pass@1} & {\(D(q)\)} & {Pass@1} & {\(D(q)\)} \\
    \midrule
    \([0,\,0.25]\)    &  7.07 & 14.90 &  8.83 & 12.74 
                      & 23.40 & 14.44 & 23.26 & 19.57 \\
    \((0.25,\,0.5]\)  & 18.75 & 24.35 & 21.88 & 22.71
                      & 69.06 & 20.74 & 71.88 & 21.40 \\
    \((0.5,\,0.75]\)  & 92.19 & 24.67 & 95.31 & 23.20
                      & 84.01 & 15.89 & 84.38 & 14.42 \\
    \((0.75,\,1.0]\)  & \multicolumn{1}{c}{--} & \multicolumn{1}{c}{--}
                      & \multicolumn{1}{c}{--} & \multicolumn{1}{c}{--}
                      & 98.80 & 11.62 & 98.98 & 11.90 \\
    \bottomrule
  \end{tabular*}
\end{table}

\subsubsection{Benchmark-Specific Analysis}

For Qwen3-1.7B, the reduction in \(D(q)\) varies across benchmarks. The clearest improvement appears on HMMT~2026, where CARE lowers dispersion in all three available subsets, including a reduction from \(30.40\%\) to \(24.79\%\) on \((0.5,\,0.75]\) together with a Pass@1 improvement. AIME~2026 also shows a decrease on \((0.25,\,0.5]\), from \(15.63\%\) to \(4.65\%\). However, on MATH500, CARE slightly reduces \(D(q)\) on \((0.25,\,0.5]\) and the easiest subset, while dispersion increases on the other two subsets.

A stronger and more consistent pattern emerges for Qwen3-4B. On AIME~2025, \(D(q)\) decreases across all four difficulty subsets, while Pass@1 is maintained or improved in most cases. The \((0.25,\,0.5]\) subset of AIME~2026 is particularly notable, with dispersion dropping from \(24.12\%\) to \(15.63\%\) as Pass@1 rises from \(47.92\%\) to \(56.25\%\). CARE likewise achieves lower \(D(q)\) throughout MATH500. For HMMT~2026, the largest accuracy improvement occurs on \((0.5,\,0.75]\), where Pass@1 increases from \(56.25\%\) to \(81.25\%\) while \(D(q)\) decreases slightly.

The Qwen3-8B results further show that this behavior persists at a larger model size. CARE lowers \(D(q)\) across all available subsets on both AIME~2025 and AIME~2026, with clear reductions in \(D(q)\) on the partially solvable subsets. HMMT~2026 exhibits a similarly consistent trend, where lower dispersion is accompanied by higher Pass@1 in all three available subsets. In contrast, MATH500 exhibits a different pattern, with \(D(q)\) decreasing on \((0.5,\,0.75]\) but increasing  in the other subsets despite largely comparable or improved Pass@1.

\clearpage
\section{Accuracy Under Varying Budget Constraints}
\label{app:budget_constraints}

This section evaluates the robustness of CARE under varying inference budgets and reports Pass@1 on AIME~2025, AIME~2026, HMMT~2026, and MATH500 for Qwen3-1.7B, Qwen3-4B and Qwen3-8B.

\subsection{Evaluation Protocol}
We evaluate both CARE and GRPO on Qwen3-1.7B, Qwen3-4B and Qwen3-8B by sweeping the maximum reasoning length across five token budgets: \(\{2048,\, 4096,\, 8192,\, 12288,\, 16384\}\).
\subsection{Experiment Results}
\begin{figure}[H]
  \centering
  \begin{subfigure}[t]{0.48\linewidth}
    \centering
    \includegraphics[width=\linewidth]{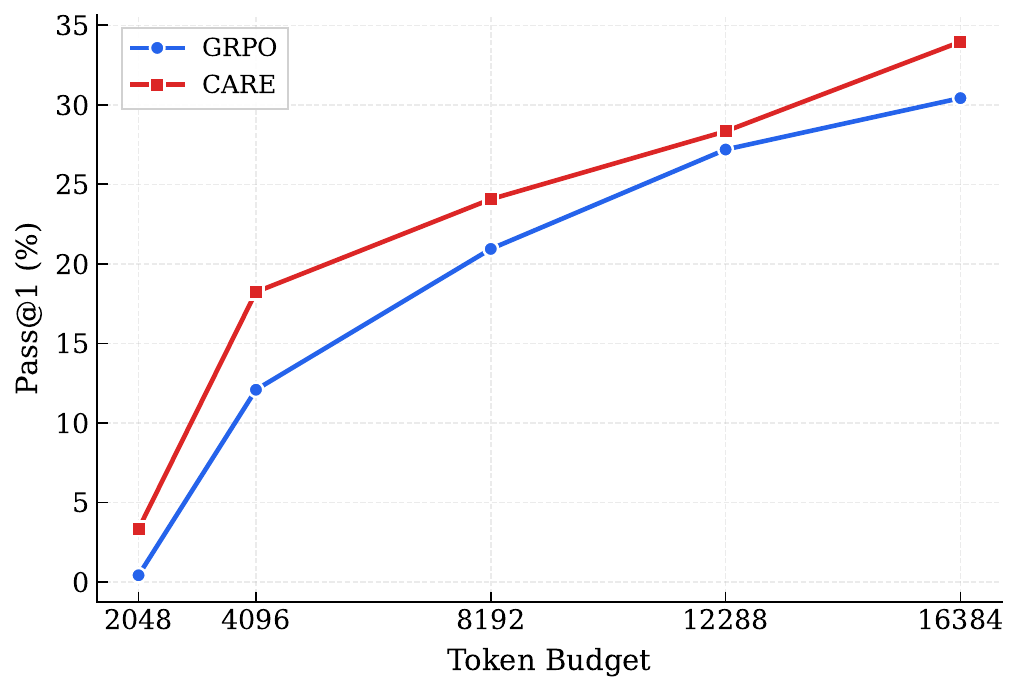}
    \caption{AIME 2025}
  \end{subfigure}\hfill
  \begin{subfigure}[t]{0.48\linewidth}
    \centering
    \includegraphics[width=\linewidth]{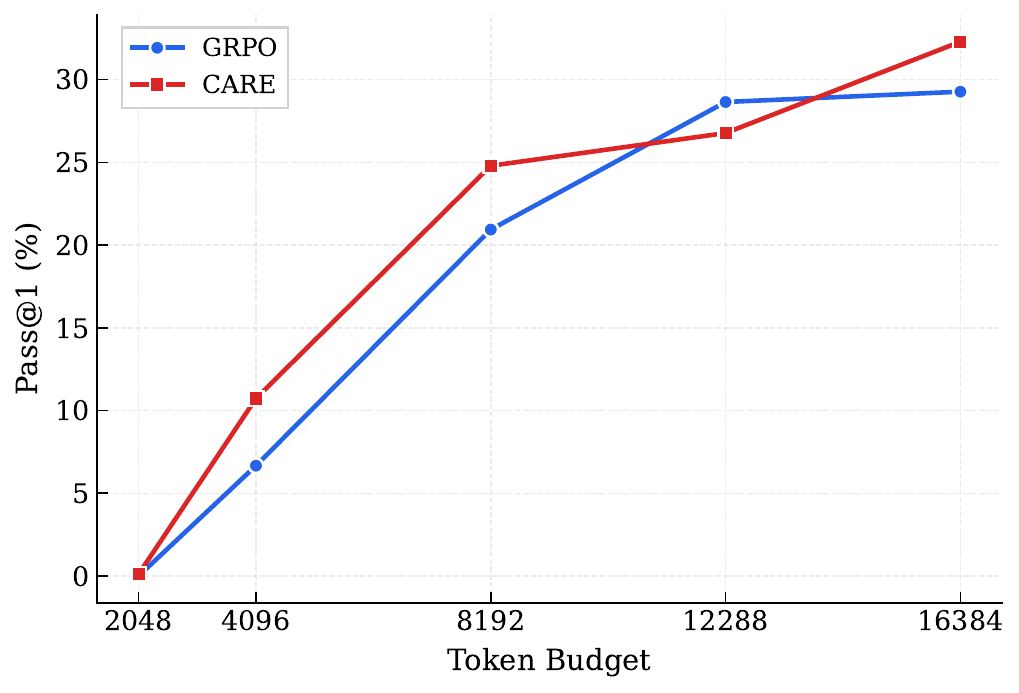}
    \caption{AIME 2026}
  \end{subfigure}
 
  \begin{subfigure}[t]{0.48\linewidth}
    \centering
    \includegraphics[width=\linewidth]{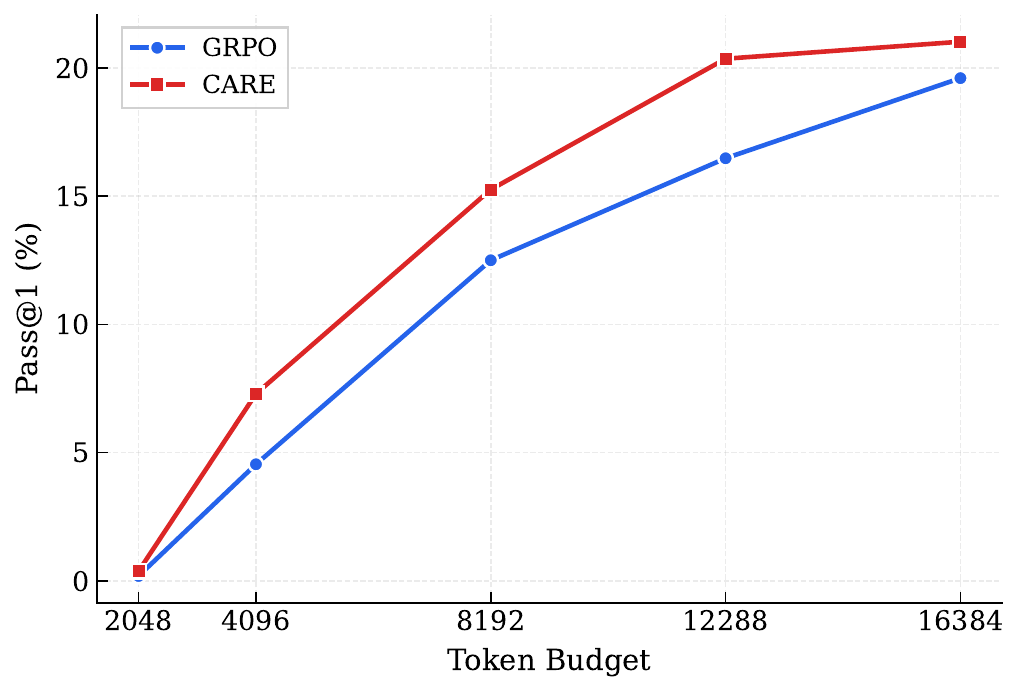}
    \caption{HMMT 2026}
  \end{subfigure}\hfill
  \begin{subfigure}[t]{0.48\linewidth}
    \centering
    \includegraphics[width=\linewidth]{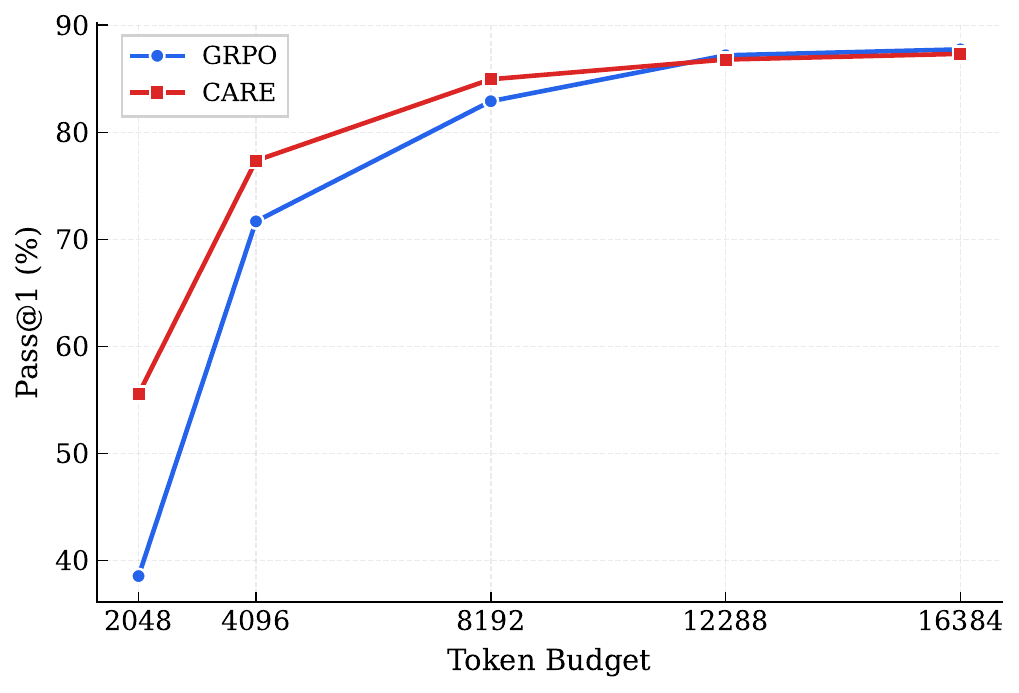}
    \caption{MATH500}
  \end{subfigure}
  \caption{Pass@1 under varying token budgets for \textbf{Qwen3-1.7B}.}
  \label{fig:app_budget_1.7b}
\end{figure}

\begin{figure}[H]
  \centering
  \begin{subfigure}[t]{0.48\linewidth}
    \centering
    \includegraphics[width=\linewidth]{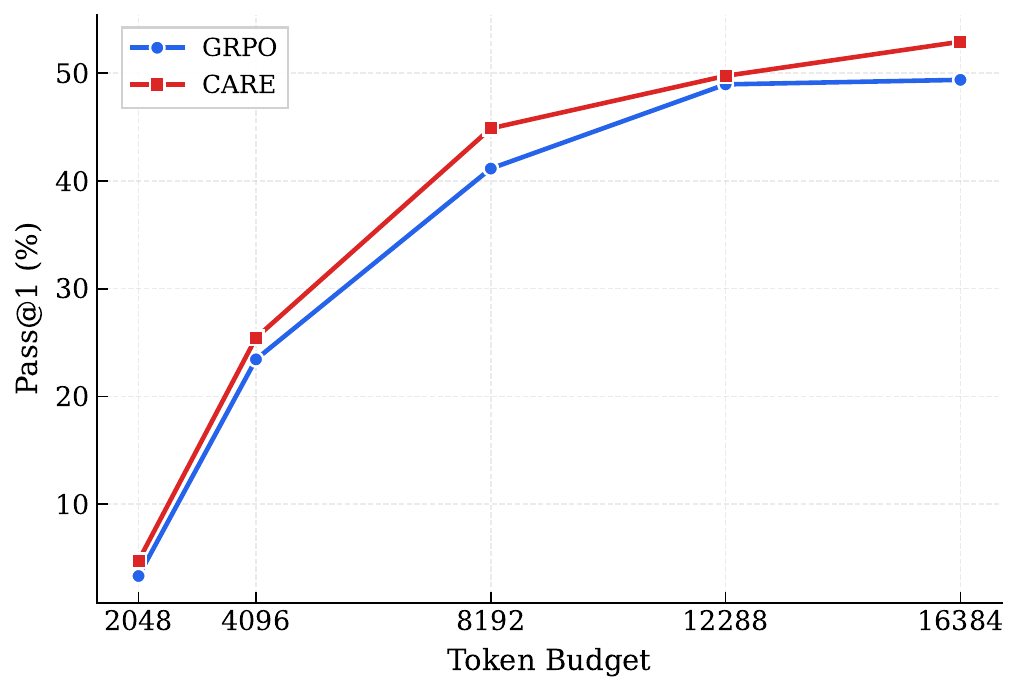}
    \caption{AIME 2025}
  \end{subfigure}\hfill
  \begin{subfigure}[t]{0.48\linewidth}
    \centering
    \includegraphics[width=\linewidth]{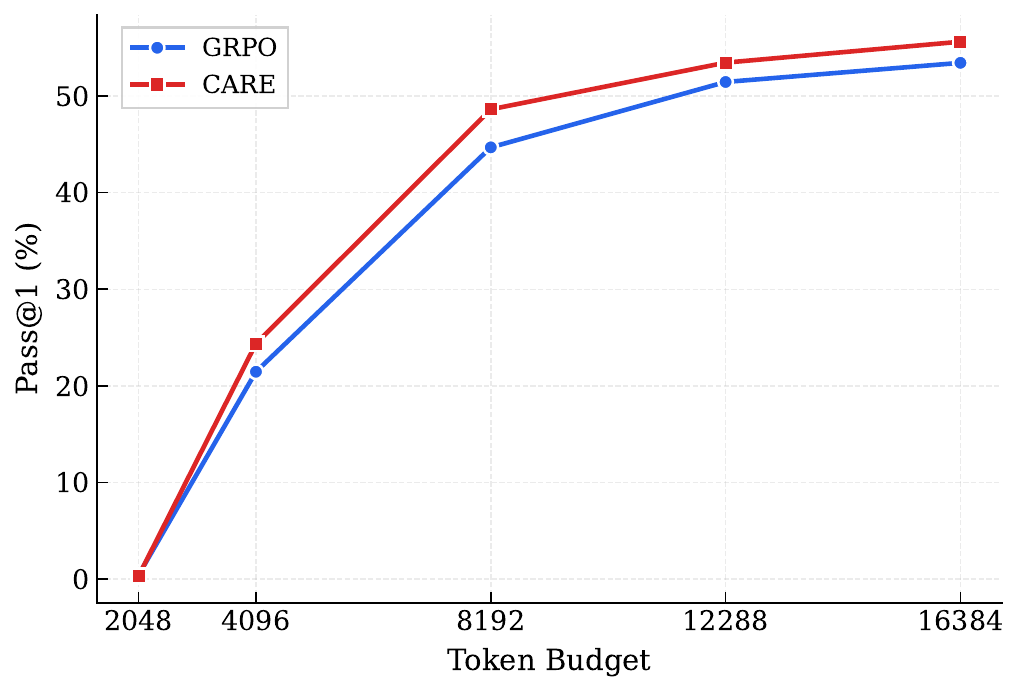}
    \caption{AIME 2026}
  \end{subfigure}
 
  \begin{subfigure}[t]{0.48\linewidth}
    \centering
    \includegraphics[width=\linewidth]{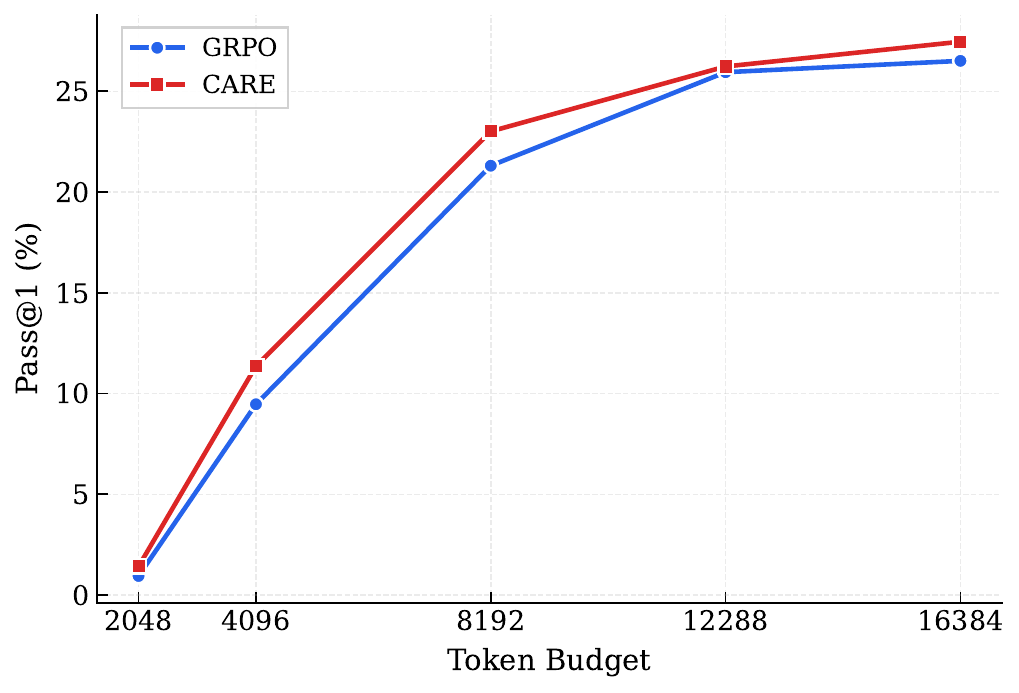}
    \caption{HMMT 2026}
  \end{subfigure}\hfill
  \begin{subfigure}[t]{0.48\linewidth}
    \centering
    \includegraphics[width=\linewidth]{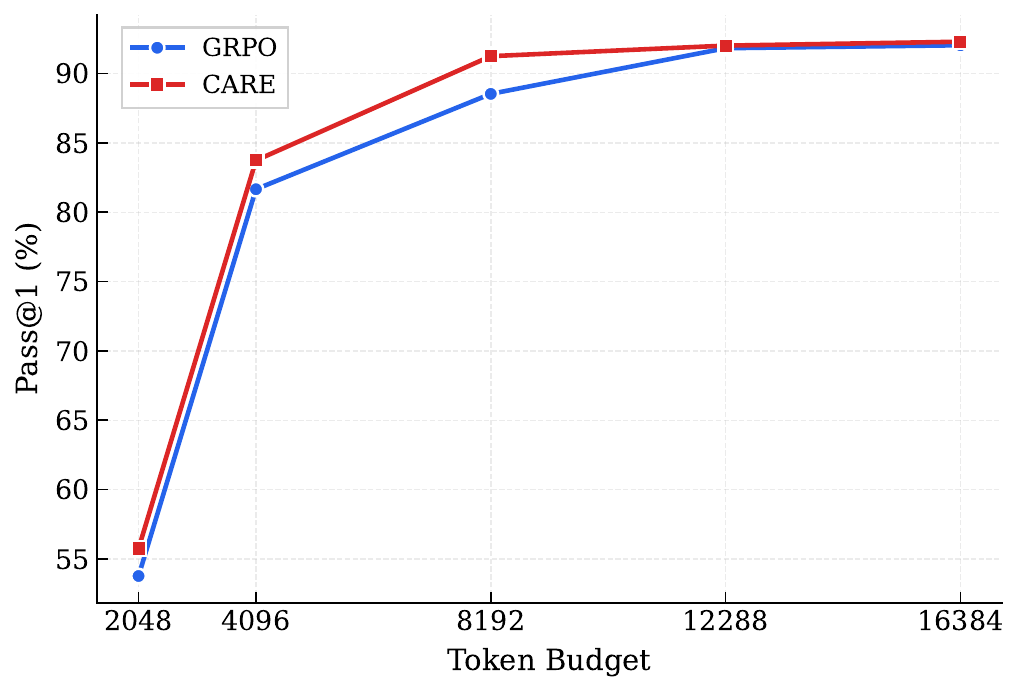}
    \caption{MATH500}
  \end{subfigure}
  \caption{Pass@1 under varying token budgets for \textbf{Qwen3-4B}.}
  \label{fig:app_budget_4b}
\end{figure}

\begin{figure}[H]
  \centering
  \begin{subfigure}[t]{0.48\linewidth}
    \centering
    \includegraphics[width=\linewidth]{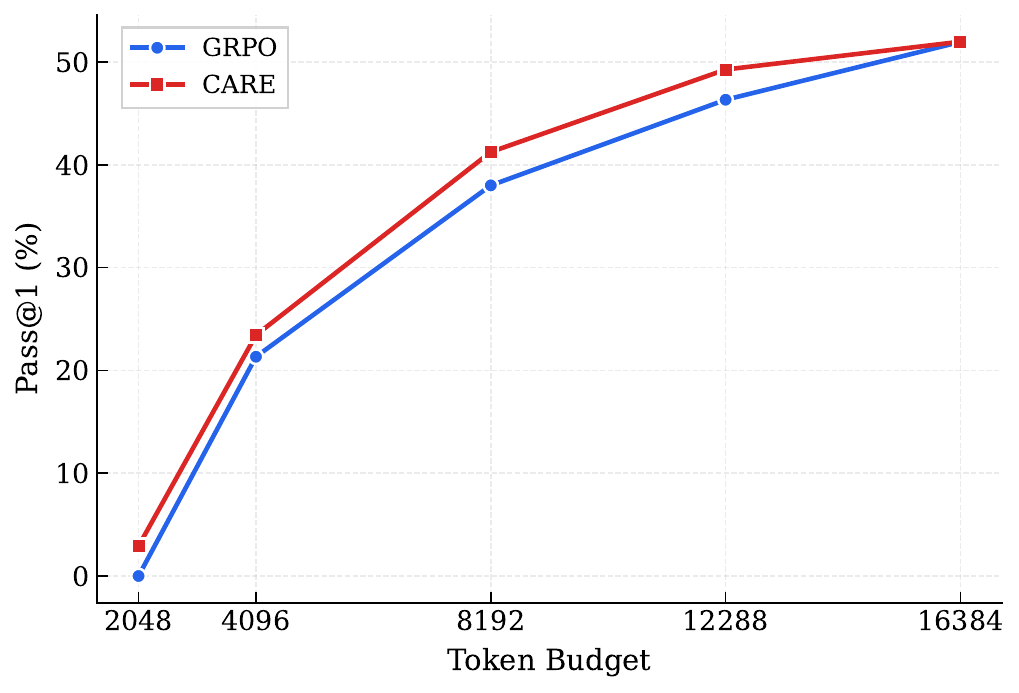}
    \caption{AIME 2025}
  \end{subfigure}\hfill
  \begin{subfigure}[t]{0.48\linewidth}
    \centering
    \includegraphics[width=\linewidth]{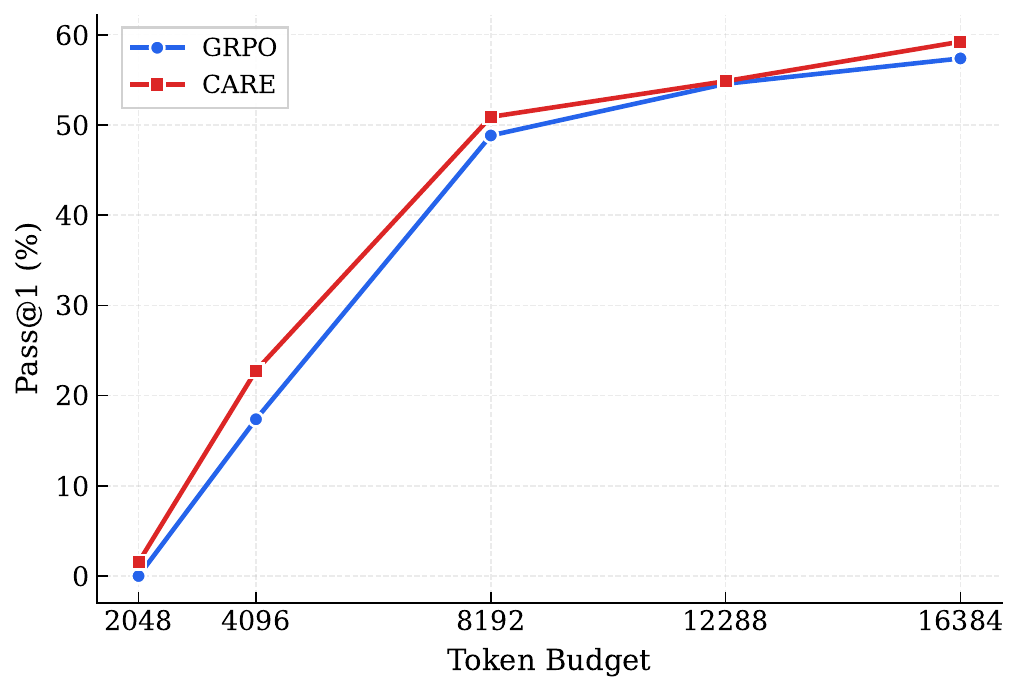}
    \caption{AIME 2026}
  \end{subfigure}
 
  \begin{subfigure}[t]{0.48\linewidth}
    \centering
    \includegraphics[width=\linewidth]{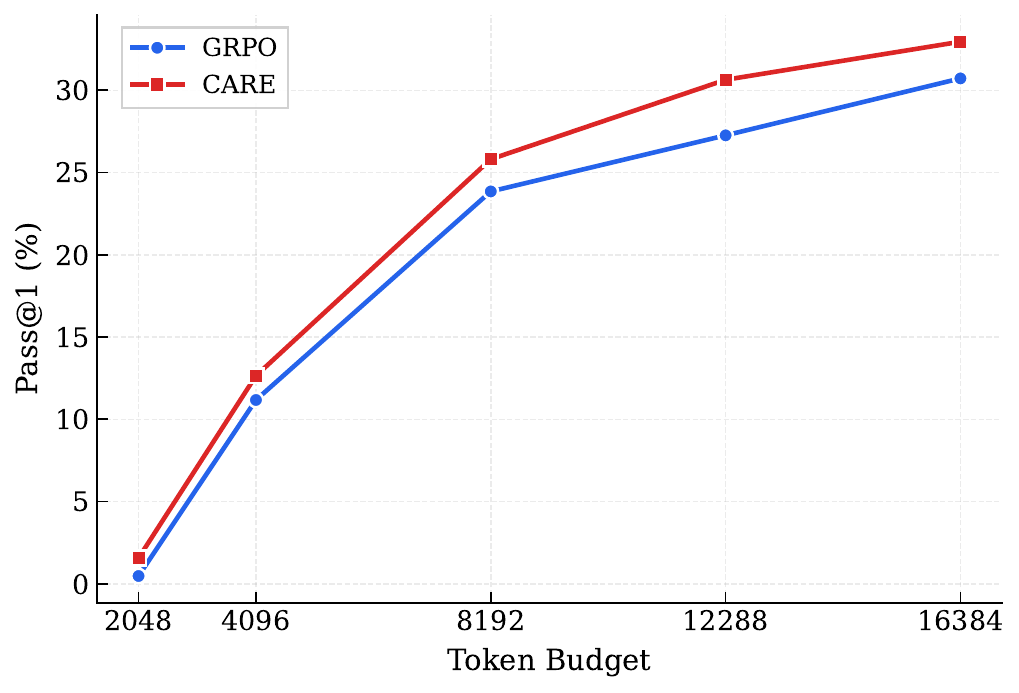}
    \caption{HMMT 2026}
  \end{subfigure}\hfill
  \begin{subfigure}[t]{0.48\linewidth}
    \centering
    \includegraphics[width=\linewidth]{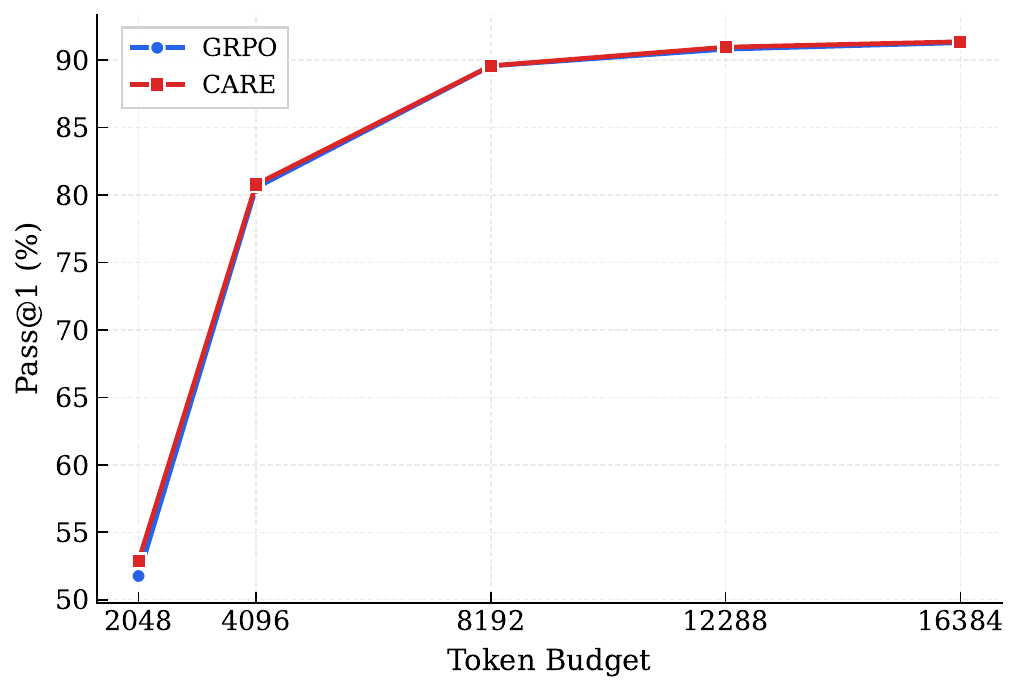}
    \caption{MATH500}
  \end{subfigure}
  \caption{Pass@1 under varying token budgets for \textbf{Qwen3-8B}.}
  \label{fig:app_budget_8b}
\end{figure}

As shown in Figures~\ref{fig:app_budget_1.7b}--~\ref{fig:app_budget_8b}, CARE mostly achieves comparable or higher Pass@1 than GRPO under restricted token budget, while the gap tends to diminish as the budget increases, although the pattern varies in different models and
benchmarks. This suggests that CARE prioritizes token-efficient reasoning paths, an advantage more apparent under limited computation budgets.

\subsubsection{Benchmark-Specific Analysis}

On these benchmarks, these models exhibit a shared pattern but differ in Pass@1. For Qwen3-1.7B, CARE begins to outperform GRPO starting from \(2{,}048\) tokens on both AIME~2025 and AIME~2026, and this gap remains relatively stable through \(16{,}384\). Qwen3-4B follows a similar trajectory over the same budget range. On AIME~2026, CARE leads GRPO by approximately \(4\%\) at \(8{,}192\) tokens, while on AIME~2025 the gap is slightly smaller but equally persistent from \(4{,}096\) onward. For Qwen3-8B, the improvement is also more visible under restricted budgets. CARE achieves higher Pass@1 on both AIME benchmarks at \(4{,}096\) and \(8{,}192\) tokens, while the difference becomes small at \(12{,}288\) and \(16{,}384\) tokens. On HMMT~2026, CARE consistently maintains an advantage under these five token budgets and all three model sizes, whereas the curves on MATH500 become increasingly close as the token budget grows, particularly for Qwen3-8B.

\section{Statistical Significance Analysis}
\label{app:statistical_significance}
This section analyzes the main results across random seeds. We additionally include ALP~\citep{xiang2025alp} and DAST~\citep{shen2025dast}, two efficient-reasoning baselines discussed in Section~\ref{sec:related_work}, and compare them with Base, GRPO, and CARE across Qwen3-1.7B, Qwen3-4B, and Qwen3-8B.

\subsection{Evaluation Protocol}

For each model--method combination, we conduct three independent evaluation runs using different random seeds on AIME~2025, AIME~2026, HMMT~2026, and MATH500 under the same evaluation settings as Section~\ref{sec:experimental_setup}. Each entry reports the mean Pass@1, sample standard deviation, and the corresponding 95\% Student-\(t\) confidence interval.

\subsection{Experiment Results}

Table~\ref{tab:multiseed_results} presents the complete multi-seed evaluation results. CARE achieves the highest mean Pass@1 in nine of the twelve model--benchmark combinations and remains competitive in the remaining evaluated settings. In particular, CARE consistently obtains the best mean performance on all four benchmarks for Qwen3-4B, and on AIME~2026, HMMT~2026, and MATH500 for Qwen3-8B. For Qwen3-1.7B, CARE performs best on AIME~2026 and HMMT~2026, while remaining close to the strongest baseline on AIME~2025. These results show that the performance of CARE remains robust across different random seeds and model sizes.

\begin{table}[H]
\centering
\caption{Pass@1 (\%) across three random seeds for Qwen3 family. Each entry reports mean \(\pm\) sample SD [95\% Student-\(t\) CI]. The best mean result is shown in bold.}
\label{tab:multiseed_results}
\small
\setlength{\tabcolsep}{3.5pt}
\renewcommand{\arraystretch}{1.08}
\resizebox{\textwidth}{!}{
\begin{tabular}{llcccc}
\toprule
Model Size & Method & AIME 2025 & AIME 2026 & HMMT 2026 & MATH500 \\
\midrule
\multirow{5}{*}{Qwen3-1.7B}
& Base
& \(28.68 \pm 1.65\,[24.59, 32.77]\)
& \(31.63 \pm 1.29\,[28.42, 34.85]\)
& \(16.41 \pm 0.88\,[14.23, 18.60]\)
& \(85.42 \pm 2.02\,[80.41, 90.44]\) \\
& GRPO
& \(29.48 \pm 0.81\,[27.46, 31.50]\)
& \(30.21 \pm 0.81\,[28.19, 32.23]\)
& \(19.16 \pm 1.38\,[15.73, 22.59]\)
& \(\mathbf{87.15 \pm 1.03\,[84.60, 89.70]}\) \\
& ALP
& \(\mathbf{32.29 \pm 0.72\,[30.50, 34.08]}\)
& \(30.21 \pm 1.83\,[25.66, 34.76]\)
& \(20.14 \pm 0.67\,[18.47, 21.81]\)
& \(85.90 \pm 0.16\,[85.51, 86.29]\) \\
& DAST
& \(26.22 \pm 0.22\,[25.68, 26.75]\)
& \(28.96 \pm 1.09\,[26.26, 31.66]\)
& \(16.22 \pm 0.36\,[15.33, 17.12]\)
& \(83.14 \pm 0.13\,[82.82, 83.46]\) \\
& CARE
& \(32.26 \pm 1.50\,[28.54, 35.97]\)
& \(\mathbf{32.64 \pm 0.87\,[30.47, 34.81]}\)
& \(\mathbf{21.28 \pm 0.44\,[20.19, 22.36]}\)
& \(86.66 \pm 1.09\,[83.96, 89.37]\) \\
\midrule
\multirow{5}{*}{Qwen3-4B}
& Base
& \(46.88 \pm 0.63\,[45.30, 48.45]\)
& \(51.49 \pm 0.59\,[50.02, 52.96]\)
& \(24.27 \pm 0.85\,[22.17, 26.38]\)
& \(89.57 \pm 1.11\,[86.80, 92.33]\) \\
& GRPO
& \(51.70 \pm 2.03\,[46.65, 56.75]\)
& \(53.44 \pm 0.31\,[52.66, 54.21]\)
& \(27.71 \pm 1.04\,[25.12, 30.31]\)
& \(92.16 \pm 0.18\,[91.71, 92.61]\) \\
& ALP
& \(51.56 \pm 1.17\,[48.65, 54.48]\)
& \(51.28 \pm 1.33\,[47.99, 54.58]\)
& \(28.03 \pm 0.16\,[27.62, 28.44]\)
& \(90.67 \pm 0.06\,[90.53, 90.82]\) \\
& DAST
& \(34.48 \pm 1.93\,[29.69, 39.27]\)
& \(37.36 \pm 0.51\,[36.08, 38.64]\)
& \(20.68 \pm 0.39\,[19.70, 21.65]\)
& \(86.79 \pm 0.08\,[86.59, 86.99]\) \\
& CARE
& \(\mathbf{52.22 \pm 0.60\,[50.73, 53.72]}\)
& \(\mathbf{54.90 \pm 0.79\,[52.94, 56.85]}\)
& \(\mathbf{28.38 \pm 0.20\,[27.89, 28.87]}\)
& \(\mathbf{92.27 \pm 0.13\,[91.95, 92.60]}\) \\
\midrule
\multirow{5}{*}{Qwen3-8B}
& Base
& \(46.42 \pm 0.64\,[44.84, 48.00]\)
& \(52.26 \pm 0.12\,[51.96, 52.56]\)
& \(26.23 \pm 0.57\,[24.82, 27.64]\)
& \(88.16 \pm 0.10\,[87.92, 88.40]\) \\
& GRPO
& \(\mathbf{56.04 \pm 0.91\,[53.79, 58.30]}\)
& \(57.50 \pm 0.95\,[55.13, 59.87]\)
& \(32.67 \pm 0.49\,[31.45, 33.89]\)
& \(90.65 \pm 0.18\,[90.20, 91.09]\) \\
& ALP
& \(53.54 \pm 0.45\,[52.41, 54.67]\)
& \(57.15 \pm 0.75\,[55.28, 59.02]\)
& \(32.20 \pm 0.43\,[31.12, 33.27]\)
& \(90.65 \pm 0.13\,[90.34, 90.97]\) \\
& DAST
& \(52.08 \pm 0.48\,[50.90, 53.27]\)
& \(57.26 \pm 0.80\,[55.28, 59.23]\)
& \(29.55 \pm 0.09\,[29.31, 29.78]\)
& \(90.33 \pm 0.04\,[90.23, 90.43]\) \\
& CARE
& \(54.93 \pm 0.63\,[53.37, 56.49]\)
& \(\mathbf{58.30 \pm 0.66\,[56.66, 59.94]}\)
& \(\mathbf{33.18 \pm 0.67\,[31.52, 34.83]}\)
& \(\mathbf{92.72 \pm 0.08\,[92.52, 92.93]}\) \\
\bottomrule
\end{tabular}
}
\end{table}

\section{Other Experimental Details}
\label{app:other_details}
This section provides additional implementation details that
supplement the configurations reported in
Section~\ref{sec:experiment} and Appendix~\ref{app:experimental_details}, including training prompts and answer extraction.

\subsection{Training Prompt}
\label{app:training_prompt}
\begin{promptbox}[Training Prompt]
Let's think step by step, and put your final answer within \textbackslash boxed\{\}.
\end{promptbox}
\subsection{Answer Extraction}
\label{app:answer_extraction}

We extract the final answer from each model response
by locating the last \texttt{\textbackslash boxed\{...\}} expression and extracting its content, following
the extraction protocol from ~\citet{hendrycks2021math}.

\section{Case Study}
\label{app:case_study}

This section presents representative examples to illustrate the behaviors discussed in the paper. 

\subsection{Length Collapse in Qwen3-1.7B under Short-Reward}
\label{app:spcl_case_1.7b_short}
\begin{tcolorbox}[colback=white, colframe=black!80, 
  title={\textbf{Example: Length Collapse} \hfill  \textit{AIME 2025, Problem\_idx 1}}, 
  fonttitle=\small\bfseries]
\small
\textbf{Question:} Find the sum of all integer bases \(b>9\) for which \(17_b\) is a divisor of \(97_b\).
\vspace{1pt}
\hfill \colorbox{violet!12}{\small\strut \textbf{Answer:} \(70\)}
\vspace{1pt}
\begin{tcolorbox}[colback=green!5, colframe=green!50!black, 
  title={\small \textbf{Qwen3-1.7B (Base)} \hfill 7{,}162 Tokens \quad \textcolor{white}{\ding{51}}}]
\small
\texttt{<think>} \\
Okay, so I need to find all integer bases \(b\) greater than 9 where \(17_b\) is a divisor of \(97_b\). \(\cdots\) \\
So, \(17_b = b + 7\) and \(97_b = 9b + 7\). We need \((9b + 7)/(b + 7)\) to be an integer. \(\cdots\) \\
Let me write \(9b + 7 = 9(b + 7) - 56\). Ah! So this is a key insight. \(\cdots\) \\
So \(b + 7\) must be a positive divisor of 56, and \(b + 7 > 16\). \(\cdots\) \\[-2pt]
\hspace*{1.0em}\(\cdots\) \\[-2pt]
Therefore \(b = 21\) or \(b = 49\). The answer is 70. \\
\texttt{</think>} \\[2pt]
Final Answer \\
\texttt{\textbackslash boxed\{70\}}
\end{tcolorbox}
\begin{tcolorbox}[colback=red!5, colframe=red!50!black,
  title={\small \textbf{Qwen3-1.7B (GRPO + Short Reward)} \hfill 8 Tokens \quad \textcolor{white}{\ding{55}}}]
\footnotesize
\texttt{</think>} \\
\texttt{\textbackslash boxed\{6\}}
\end{tcolorbox}
\end{tcolorbox}

\subsection{Comparison of GRPO and CARE Responses}
\label{app:case_grpo_care}

\begin{tcolorbox}[colback=white, colframe=black!80, boxrule=0.5pt,
  title={\small \textbf{Example: Reasoning Trace in GRPO and CARE} \hfill \textit{HMMT 2026, Problem\_idx 1}}]
\small
\textbf{Question:} A line intersects the graph of \( y = x^2 + \frac{2}{x} \) at three distinct points. Given that the \( x \)-coordinates of two of the points are 6 and 7, compute the \( x \)-coordinate of the third point.
\vspace{1pt}
\hfill \colorbox{violet!12}{\small\strut \textbf{Answer:} \(-1/21\)}
\vspace{1pt}
\begin{minipage}[t]{0.48\linewidth}
\begin{tcolorbox}[colback=orange!5, colframe=orange!60!black, 
  title={\scriptsize GRPO \hfill 13{,}569 Tokens \quad \textcolor{white}{\ding{55}}},
  height=6.5cm, valign=top]
\scriptsize
\texttt{<think>} \\
Okay, so I need to find the x-coordinate of the \(\cdots\) \\[-2pt]
\hspace*{1.0em}\(\cdots\) \\[-2pt]
Therefore, the third root is 5/21. Therefore, the \(\cdots\)\\[-2pt]
\hspace*{1.0em}\(\cdots\) \\[-2pt]
\textcolor{red}{\textbf{Wait}}, that seems straightforward. Let me check \(\cdots\)\\[-2pt]
\hspace*{1.0em}\(\cdots\) \\[-2pt]
\textcolor{red}{\textbf{Wait}}, that's not zero. So something is wrong here. That means my calculation is wrong. Hmm. So \(\cdots\)\\[-2pt]
\hspace*{1.0em}\(\cdots\) \\[-2pt]
\textcolor{red}{\textbf{But}} when I plug \(x\!=\!7\) into the cubic equation, it gives \(-42\), but according to Vieta's formula, the \(\cdots\)  \\[-2pt]
\hspace*{1.0em}\(\cdots\) \\[-2pt]
\textcolor{red}{\textbf{But}} how? This is very confusing. I must have \(\cdots\) \\[-2pt]
\hspace*{1.0em}\(\cdots\) \\[-2pt]
\textcolor{red}{\textbf{Alternatively}}, maybe \(x\!=\!6\) and \(x\!=\!7\) are not \(\cdots\) \\[-2pt]
\hspace*{1.0em}\(\cdots\) \\[-2pt]
I will trust Vieta's formula here. The x-coordinate of the third point is \(5/21\). \\
\texttt{</think>} \\[-2pt]
\hspace*{1.0em}\(\cdots\) \\[-2pt]
Final Answer \\
\texttt{\textbackslash boxed\{5/21\}}
\end{tcolorbox}
\end{minipage}
\hfill
\begin{minipage}[t]{0.48\linewidth}
\begin{tcolorbox}[colback=blue!5, colframe=blue!60!black,
  title={\scriptsize CARE \hfill 1{,}591 Tokens \quad \textcolor{white}{\ding{51}}},
  height=6.5cm, valign=top]
\scriptsize
\texttt{<think>} \\
Okay, so I need to find the x-coordinate of the third point where a line intersects the graph of \( y \) \(\cdots\) \\[-2pt]
\hspace*{1.0em}\(\cdots\) \\[-2pt]
Let's say the line has equation \(y = mx + b\). Then, setting this equal to the function \(nmx + b = x^2 \)\(\cdots\) \\[-2pt]
\hspace*{1.0em}\(\cdots\) \\[-2pt]
Then, by Vieta's formula, the sum of roots of the cubic equation is equal to the coefficient of \(x^2\) term\(\cdots\)\\[-2pt]
\hspace*{1.0em}\(\cdots\) \\[-2pt]
Multiply both sides by \(x\): \(x^3 - mx^2 - bx + 2 = 0\). \\
The product of the roots is \(-2\). Therefore, \(6 \cdot 7 \cdot r\)\(\cdots\) \\[-2pt]
\hspace*{1.0em}\(\cdots\) \\[-2pt]
Let me verify. If the product of the roots is \(-2\), \(\cdots\) \\[-2pt]
\hspace*{1.0em}\(\cdots\) \\[-2pt]
Therefore, the answer is \(-1/21\). \\
\texttt{</think>} \\[-2pt]
\hspace*{1.0em}\(\cdots\) \\[-2pt]
Final Answer \\
\texttt{\textbackslash boxed\{-1/21\}}
\end{tcolorbox}
\end{minipage}
\end{tcolorbox}

\subsection{Cyclic Reasoning: Representative Cases}
\label{app:cyclic_cases}
 
In this part, we present three representative cases from the unstable interval of Qwen3-1.7B and Qwen3-4B under long-reward training, where keyword-flagged sentences correspond to semantic repetition. Cyclic keywords are marked in \textcolor{red}{\textbf{red bold}}.
 
\begin{tcolorbox}[colback=white, colframe=black!80, boxrule=0.5pt,
  title={\small \textbf{Example\,1} \hfill
         \textit{Qwen3-1.7B \(\cdot\) Long-Reward}}]
\small
\textbf{Question:} Simplify \(90r - 44r\).
\hfill \colorbox{violet!12}{\small\strut \textbf{Answer:} \(46r\)}
 
\textbf{Selection rationale.}
The unstable response re-derives the \emph{identical} subtraction
\(90 - 44 = 46\) immediately after stating it,
triggered by ``\textcolor{red}{Alternatively}.''
 
\begin{tcolorbox}[colback=unstableinterval_one!5,
colframe=unstableinterval_one,
  valign=top,
  title={\scriptsize Unstable Interval \hfill Step 40
         \quad 4{,}139 Tokens}]
\scriptsize
\texttt{<think>} \\[-2pt]
\hspace*{1.0em}\(\cdots\) \\[-2pt]
Okay, so I need to simplify the expression 90r minus 44r. Let me start by thinking about what this means. \(\cdots\) \\[-2pt]
\hspace*{1.0em}\(\cdots\) \\[-2pt]
Hmm, 90 minus 40 is 50, and then minus 4 more would be 46. So, 90 minus 44 is 46. Therefore, 90r minus 44r is 46r.  \\[2pt]
\textcolor{red}{\textbf{Alternatively}}, maybe I can use algebraic manipulation. Let me write it as: \(90r - 44r = (90 - 44)r\). Because when you have the same variable, you can combine the coefficients. So, \(90 - 44\) is 46, so it's \(46r\). \(\cdots\) \\[-2pt]
\hspace*{1.0em}\(\cdots\) \\[-2pt]
\texttt{</think>} \\[-2pt]
\hspace*{1.0em}\(\cdots\) \\[-2pt]
Final Answer \\
The simplified form of \(90r - 44r\) is \texttt{\textbackslash boxed\{\(46r\)\}}.
\end{tcolorbox}
\end{tcolorbox}
\vspace{3pt}
\begin{tcolorbox}[colback=white, colframe=black!80, boxrule=0.5pt,
  title={\small \textbf{Example\,2} \hfill
         \textit{Qwen3-1.7B \(\cdot\) Long-Reward}}]
\small
\textbf{Question:} A room is 19 feet long and 11 feet wide.
Find the ratio of the length of the room to its perimeter.
Express your answer in the form \(a:b\).
\hfill \colorbox{violet!12}{\small\strut \textbf{Answer:} \(19\!:\!60\)}
 
\textbf{Selection rationale.}
The unstable response re-derives the \emph{identical} perimeter calculation
\(2\!\times\!19 + 2\!\times\!11 = 60\) immediately after stating it,
triggered by ``\textcolor{red}{Wait}.''
 
\begin{tcolorbox}[colback=unstableinterval_two!5, colframe=unstableinterval_two,
  valign=top,
  title={\scriptsize Unstable Interval \hfill Step 43
         \quad 6{,}144 Tokens}]
\scriptsize
\texttt{<think>} \\[-2pt]
Okay, so I need to find the ratio of the length of the room to its perimeter. The room is 19 feet long and 11 feet wide. \(\cdots\) \\[-2pt]
\hspace*{1.0em}\(\cdots\) \\[-2pt]
So the perimeter is 60 feet?  \\[2pt]
\textcolor{red}{\textbf{Wait}}, let me check that again. \(2\!\times\!19\) is 38, \(2\!\times\!11\) is 22. \(38 + 22\). Hmm, \(38 + 22\). Yeah, \(38 + 20\) is 58, plus 2 is 60. \(\cdots\) \\[-2pt]
\hspace*{1.0em}\(\cdots\) \\[-2pt]
For example, if the question was about the ratio of the width to the perimeter, then it would be \(11\!:\!60\), but the problem says length to perimeter. \textcolor{red}{\textbf{Alternatively}}, maybe I need to think if there's a mistake in the problem \(\cdots\)
\end{tcolorbox}
\end{tcolorbox}
\vspace{3pt}
\begin{tcolorbox}[colback=white, colframe=black!80, boxrule=0.5pt,
  title={\small \textbf{Example\,3} \hfill
         \textit{Qwen3-4B \(\cdot\) Long-Reward}}]
\small
\textbf{Question:} The operation \(\#\) is defined as \(a \# b = a + \frac{a}{b}\).
What is the value of \(6 \# 2\)?
\hfill \colorbox{violet!12}{\small\strut \textbf{Answer:} \(9\)}
 
\textbf{Selection rationale.}
The unstable response returns to the same substitution-and-calculation path
\(6 \# 2 = 6 + \frac{6}{2} = 6 + 3 = 9\) after considering an alternative
interpretation, triggered by ``\textcolor{red}{But}.''
 
\begin{tcolorbox}[colback=brown!5,
colframe=brown!45!black,
  valign=top,
  title={\scriptsize Unstable Interval \hfill Step 37
         \quad 5{,}193 Tokens}]
\scriptsize
\texttt{<think>} \\[-2pt]
Okay, so I need to figure out what \(6 \# 2\) is, given that the operation \(\#\) is defined as \(a \# b = a + (a/b)\). \(\cdots\) \\[-2pt]
\hspace*{1.0em}\(\cdots\) \\[-2pt]
So, it's definitely a plus \((a\) divided by \(b)\). So, in this case, with \(a\) being 6 and \(b\) being 2, it's 6 plus \(6\) divided by 2.\\[2pt]
\textcolor{red}{\textbf{But}} that would be different. Let me check the original problem again.\(\cdots\)  \textcolor{red}{\textbf{Alternatively}}, if it were written without parentheses, \(\cdots\)  So, going back to the original calculation: \(6 \# 2\) is \(6 + (6 / 2)\).\(\cdots\)\\[-2pt]
\hspace*{1.0em}\(\cdots\) \\[-2pt]
</think>\\[-2pt]
\hspace*{1.0em}\(\cdots\) \\[-2pt]
Final Answer \\[2pt]
\texttt{\textbackslash boxed\{\(9\)\}}
\end{tcolorbox}
\end{tcolorbox}
 
\section{Compute Resources}
\label{app:experiments_compute_resources}
Each training run (one model × one reward configuration) required approximately 120 hours on 8\(\times\)A100~80GB GPUs. All evaluations were 
performed on 8\(\times\)NVIDIA RTX 4080 Super GPUs.

\section{Limitations}
\label{app:Limitations}
Our evaluation is limited to mathematical reasoning tasks. The generalizability of CARE to other reasoning domains, such as code generation or commonsense reasoning, remains to be explored. In addition, the current direction signal \(d(q)\) is a discrete ternary value (\(-1, 0, +1\)). A continuous variant may provide finer-grained control but is left for future work.

\clearpage

\end{document}